\documentclass{article}

\usepackage[final, nonatbib]{neurips_2023}

\usepackage[utf8]{inputenc} 
\usepackage[T1]{fontenc}    
\usepackage{hyperref}       
\usepackage{url}            
\usepackage{physics}
\usepackage{cite}
\usepackage{booktabs}       
\usepackage{bookmark}       
\usepackage{amsfonts}       
\usepackage{amsmath}        
\usepackage{amsthm}         
\usepackage{amssymb}
\usepackage{nicefrac}       
\usepackage{microtype}      
\usepackage{xcolor}         
\usepackage{graphicx}       
\usepackage{subcaption}     
\usepackage{svg}            
\usepackage{tikz}           
\usepackage{mathtools}
\usepackage{dsfont}

\usepackage[createShortEnv]{proof-at-the-end}
\usepackage{aligned-overset}
\usepackage{cancel}
\usepackage{pdfpages}
\usepackage{pgfplots}
\usepackage{wrapfig}  
\pgfplotsset{width=10cm,compat=1.9}

\usepackage{stackengine} 
\newcommand\oast{\stackMath\mathbin{\stackinset{c}{0ex}{c}{0ex}{\ast}{\bigcirc}}}
\newtheorem*{Interchanneldistortion}{Inter-channel distortion}
\usepackage{xspace}
\newcommand{\MIMONets}{MIMONets\xspace}
\newcommand{\MIMOConv}{MIMOConv\xspace}
\newcommand{\MIMOFormer}{MIMOFormer\xspace}
\newcommand{\FAVORpS}{FAVOR+S\xspace}
\newcommand{\PWHRR}{PWHRR\xspace} 
\allowdisplaybreaks

\title{\MIMONets: Multiple-Input-Multiple-Output Neural Networks Exploiting Computation in Superposition}

\newgeometry{margin=1.5in}

\usepackage{fancyhdr}
\author{
Nicolas Menet$^{1,2}$\thanks{Research conducted at IBM Research -- Zurich.} \\
\texttt{menetn@ethz.ch} \\
\And
Michael Hersche$^{1,2}$\\
{\tt\small her@zurich.ibm.com}
\And 
Geethan Karunaratne$^{1}$\\
{\tt\small kar@zurich.ibm.com}
\And 
Luca Benini$^{2}$\\
{\tt\small lbenini@iis.ee.ethz.com}
\And 
Abu Sebastian$^{1}$\\
{\tt\small ase@zurich.ibm.com}
\And 
Abbas Rahimi$^{1}$\\
{\tt\small abr@zurich.ibm.com}
\And
{
\normalfont $^{1}$IBM Research -- Zurich, $^{2}$ETH Zurich }
}

\begin{document}

\maketitle
\begin{abstract}
With the advent of deep learning, progressively larger neural networks have been designed to solve complex tasks. We take advantage of these capacity-rich models to lower the cost of inference by exploiting \emph{computation in superposition}. To reduce the computational burden per input, we propose Multiple-Input-Multiple-Output Neural Networks (\MIMONets) capable of handling many inputs at once. \MIMONets augment various deep neural network architectures with variable binding mechanisms to represent an arbitrary number of inputs in a compositional data structure via fixed-width distributed representations. Accordingly, \MIMONets adapt nonlinear neural transformations to process the data structure holistically, leading to a speedup nearly proportional to the number of superposed input items in the data structure. After processing in superposition, an unbinding mechanism recovers each transformed input of interest. \MIMONets also provide a dynamic trade-off between accuracy and throughput by an instantaneous on-demand switching between a set of accuracy-throughput operating points, yet within a single set of fixed parameters. We apply the concept of \MIMONets to both CNN and Transformer architectures resulting in \MIMOConv and \MIMOFormer, respectively. Empirical evaluations show that \MIMOConv achieves $\approx2\,$--$\,4\times$ speedup at an accuracy delta within $[+0.68, -3.18]\%$ compared to WideResNet CNNs on CIFAR10 and CIFAR100.  
Similarly, \MIMOFormer can handle 2--4 inputs at once while maintaining a high average accuracy within a $[-1.07, -3.43]\%$ delta on the long range arena benchmark. 
Finally, we provide mathematical bounds on the interference between superposition channels in \MIMOFormer. Our code is available at \url{https://github.com/IBM/multiple-input-multiple-output-nets}.
\end{abstract}

\section{Introduction}
Driven by the successes of deep learning in image and natural language processing tasks, increasingly large neural network models have been developed to reach state-of-the-art performance~\cite{devlin2018bert,brown2020gpt3, Liu_2021_ICCV, liu2022convnet}. 
These large models, however, increase computational complexity in terms of operation count for every event of input processing. 
One viable option to reduce the computational cost of processing per input is to create a compositional data structure where a variable number of input items (i.e., values) can be bound to corresponding protection keys, creating key-value pairs that can coexist and be processed concurrently.
%
This variable-sized data structure can be represented by fixed-width distributed representations in vector-symbolic architectures (VSAs)~\cite{VSA_03, PlateHolographic1995, Kanerva2009}.
%
In VSAs, the composition of different items in the data structure is based on functional compositionality (i.e., key-value binding), which yields a dimensionality-preserving distributed representation, rather than concatenative compositionality.
Interestingly, the resulting fixed-width distributed data structure can be transformed by a one-time application of a function, whereby all input items are jointly transformed, leading to \emph{holistic transformation} or \emph{computation in superposition}~\cite{NeumannTransformation2000,NeumannTransformation2002, kleyko2022vector}.
This concept of computation in superposition can reduce the effective number of operations per input by a factor of the number of input items in the data structure, because the function is applied to the data structure holistically without decomposing the constituent items for individual transformations. 
%
However, processing the VSA data structure via computation in superposition has so far been limited to linear maps~\cite{NeumannTransformation2000,NeumannTransformation2002, kleyko2022vector}.

\begin{figure}
    \centering
\includegraphics{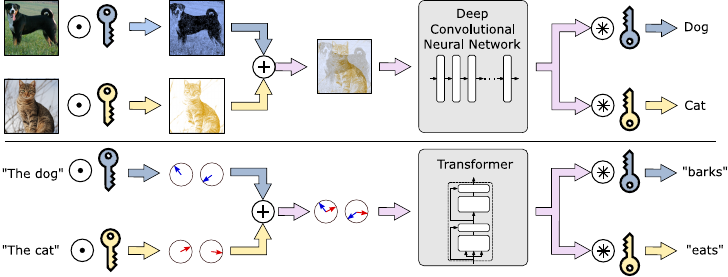}
    \caption{
    \MIMONets 
    simultaneously pass multiple inputs through a nonlinear function, e.g., a deep convolutional network (on top) or a Transformer (on bottom). Input samples are bound with high-dimensional keys to project the samples into quasi-orthogonal subspaces. 
    The results of the individual samples are retrieved at the end of the network by unbinding with corresponding keys.}
    \label{fig:mimo_concept}
\end{figure}

Motivated by these observations, we make the following contributions:

(1) We introduce a principled and transparent approach to Multiple-Input-Multiple-Output Neural Networks (\MIMONets) based on VSA, enabling computation in superposition for highly nonlinear transformations in neural networks (Section~\ref{sec:CIS_in_nonlinear_func}). 
The \MIMONets concept can be applied to various architectures, embracing the rich capacity provided by increasingly large models. 
The resulting network can handle many inputs at once, thus reducing the computational cost per input. We describe and overcome the challenges of computation in superposition, which originates from nonlinear interference of inputs in separate superposition channels.

(2) We propose \MIMOConv, a realization of \MIMONets for deep convolutional neural network (CNN) architectures (Section~\ref{sec:methods_MIMOConv}).
We provide several strategies for mitigating interference between different superposition channels, including a novel locality-preserving binding operation (\PWHRR) and isometry-inducing regularization. Empirical evaluations on CIFAR10 and CIFAR100 show that a \MIMOConv built on a WideResNet-28-10~\cite{BMVC2016_87} can process concurrently two inputs in superposition ($\approx2\times$ speedup) even with a slightly higher accuracy ($0.11$--$0.68\%$ gain), and four inputs ($\approx4\times$ speedup) with a marginal drop ($1.24$--$3.18\%$) (see Section~\ref{sec:results-MIMOConv}).

(3) We further extend the concept of \MIMONets to Transformer architectures, where the calculation of attention scores poses additional challenges for the computation in superposition paradigm. To this end, we propose \MIMOFormer, which relies on a 2D grid binding scheme for computing attention in superposition
(Section~\ref{sec:method_transformers}). 
We derive probabilistic tail bounds on the distortion caused by inter-channel interference and show that our method converges to noise-free attention in the limit of high dimension. Our method succeeds empirically ($\geq 96.52\%$ accuracy) at synthetic sequence modelling tasks~\cite{fu2022h3}, while previous work~\cite{murahari2022datamux} fails ($\leq 20.04\%$). We also provide evaluations on the long range arena (LRA) dataset~\cite{tay2021long} in Section~\ref{sec:results_transformer} using a \MIMOFormer that is based on the Performer architecture~\cite{choromanski2020performer}. Compared to the Performer, \MIMOFormer maintains a high average accuracy with a marginal drop of $1.07\%$ and $3.43\%$ when handling two and four inputs at once, respectively.

(4) \MIMONets allow a \emph{dynamic} trade-off at inference time between accuracy and speed, i.e., they offer an instantaneous on-demand switching between accuracy-throughput operating points using \emph{a single set of fixed parameters}. Experimental results in Section~\ref{sec:results-MIMOConv} show that a dynamic \MIMOConv can seamlessly operate at various modes ($\approx1\,$--$\, 4\times$ speedup) while maintaining a high accuracy compared to the best static models ($\leq 1.82\%$ drop). 

\section{MIMONets enabling Computation in Superposition}
\label{sec:CIS_in_nonlinear_func}
The central idea behind \MIMONets (see Figure~\ref{fig:mimo_concept}) is to simultaneously pass multiple inputs as a superposition through a nonlinear function $f_\theta$ parameterized by neural network weights $\theta$. We isolate the individual inputs into separate protected \textit{channels} by binding them with protection keys resulting in a key-value data structure~\cite{VSA_03, PlateHolographic1995, Kanerva2009}. 

\paragraph{Concept.}
Assuming two inputs ($x^{(1)}$ and $x^{(2)}$), which can be generic embeddings either from images or natural language, we define a unique high-dimensional key ($a^{(1)}$ and $a^{(2)}$) for each protected channel, drawn randomly at initialization. Owing to the Blessing of Dimensionality, randomly drawn high-dimensional vectors are quasi-orthogonal with high probability (see Appendix~A). 
Consequently, binding ($\odot$) the inputs with these keys yields quasi-orthogonal key-value pairs ($x^{(1)} \odot a^{(1)} $ and $x^{(2)} \odot a^{(2)}$), which enables one to superpose the pairs with low interference:
\begin{align}
    s = a^{(1)} \odot x^{(1)}  +  a^{(2)}  \odot x^{(2)}.
\end{align}
As discussed in Appendix~A, \(s\) admits a noisy retrieval of $x^{(1)}$ and $x^{(2)}$ through unbinding:
\begin{align}
    \hat{x}^{(1)} = a^{(1)} \oast s = a^{(1)} \oast a^{(1)} \odot x^{(1)} + a^{(1)} \oast a^{(2)} \odot x^{(2)} = x^{(1)} + noise
\end{align}

To accelerate computing, inspired by the above-mentioned noisy retrieval, we pass the superposition $s$ through a nonlinear function $f_\theta$ with parameters $\theta$, such as a neural network, before retrieval. The quasi-orthogonality of the bound inputs allows processing each in a separate \emph{protected} subspace---all with a single function call. 
To be able to recover the first processed sample $f_\theta (x^{(1)})$ from $f_\theta (s)$, we aim to find an unbinding key $\tilde{a}^{(1)}$ for which
\begin{align}
    \tilde{a}^{(1)} \oast f_\theta(s) &\approx \tilde{a}^{(1)} \oast f_\theta\left(a^{(1)} \odot x^{(1)}\right) +  \tilde{a}^{(1)} \oast f_\theta\left(a^{(2)} \odot x^{(2)}\right) \label{eq:linearity} \\
                                    &\approx f_\theta\left(x^{(1)}\right) + \tilde{a}^{(1)} \oast f_\theta\left(a^{(2)} \odot x^{(2)}\right). \label{eq:unbinding_linearity}
\end{align}
The first approximation holds exact for linear $f_\theta$.
As discussed in Section~\ref{sec:methods_MIMOConv}, a nonlinear $f_\theta$ can still be encouraged to allow such an approximation through appropriate weight regularization techniques and well-suited activation functions.
Further, by optimizing over unbinding keys ($\tilde{a}^{(i)}$), the second estimation (Eq.~\eqref{eq:unbinding_linearity}) can be achieved. Consequently, matching binding and unbinding keys ($a^{(i)}$ and $\tilde{a}^{(i)}$) that confirm the approximation~(Eq.~\eqref{eq:linearity} and \eqref{eq:unbinding_linearity}) set up a protected \emph{channel} through the nonlinear function $f_\theta(s)$.
Appendix~A lists the design choices of the adopted VSA, which define the operations of key-value binding and unbinding, for all MIMONet variants presented in this work. In the case of image embeddings, we use circular convolution~\cite{PlateHolographic1995} for binding and Matrix Binding of Additive Terms (MBAT)~\cite{gallant2013mbat} for unbinding. In the case of sequence tokens, we bind and unbind using the Hadamard product~\cite{MAP_1998}. Binding and unbinding keys are always data-independent, i.e., they depend only on the index of the protected channel.
See~\cite{kleyko2023survey} for alternative binding and unbinding options. 
\paragraph{Dynamic Inference.}\label{sec:dynamic_inference_theory}
Setting up $N$ protected channels through a neural network $f_\theta$ gives almost a speedup of $N\times$ due to most computations taking place in superposition. However, as is explored empirically, increasing $N$ adds inter-channel noise leading to a decrease in predictive accuracy. If a fixed trade-off is unsatisfactory, one can build a dynamic model capable of running a superposition of one up to $N$ different inputs. By inserting the same input into multiple channels and averaging the output, one effectively forms an in-network ensemble, similar to~\cite{havasi2021training, rame2021mixmo}. Using all protected channels for different inputs leads to a fast but less accurate model, whereas using all protected channels for the same input yields a slower but accurate ensemble model. By partitioning the superposition channels on demand, arbitrary configurations in between may be reached. 
Note that our method can instantaneously adapt to the current computational demand, without loading different model weights from the main memory. 
To perform across slow and fast configurations, the model should randomly switch between them during training. See Appendix~A for a more detailed explanation.

\section{MIMOConv}\label{sec:methods_MIMOConv}
\begin{figure}
    \centering
    \includegraphics{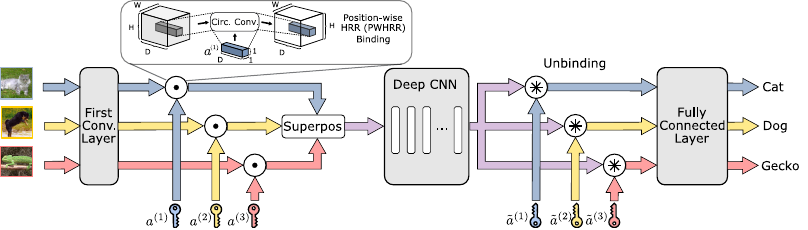}
    \caption{MIMOConv configured with $N$=$3$ channels. The input images are passed individually through the first convolutional layer before binding each feature value with a channel-specific high-dimensional key. The key-value pairs are superposed yielding a dimensionality-preserving composition and passed through the rest of the CNN layers. The output is unbound with corresponding keys, and the unbound representations are classified separately with a shared fully-connected layer. }
    \label{fig:MIMOConv_architecture}
\end{figure}

This section presents how the \MIMONets concept, introduced in Section~\ref{sec:CIS_in_nonlinear_func}, can be applied to construct a multiple-input-multiple-output CNN (\MIMOConv) capable of simultaneously processing multiple images in superposition. 
The \MIMOConv architecture is shown in Figure~\ref{fig:MIMOConv_architecture}. Multiple input samples ($N$) are passed through the first convolutional layer, bound with a unique high-dimensional key based on Holographic Reduced Representations (HRR)~\cite{PlateHolographic1995}, and superposed by element-wise addition. After passing the superposed tensors through the network's main CNN layers, we obtain a combined feature vector with information on all inputs. By unbinding with separately learned keys based on MBAT~\cite{gallant2013mbat}, which amounts to a matrix multiplication, we extract the individual processed information, which is then passed through a final fully-connected layer to produce logits for classification. In the following, we introduce our three main contributions that lead to a highly accurate \MIMOConv. 

\paragraph{Augmenting CNNs with locality-preserving variable bindings.}
We embrace a principled and transparent binding mechanism from 
HRR. 
Accordingly, binding is performed using circular convolution with a binding key of dimension $D$, drawn from an i.i.d. Gaussian distribution with zero mean and $1/D$ variance. Instead of convolving the flattened image tensor, 
we repeatedly apply circular convolution between the binding key and each pixel volume spanning across the feature maps ($D$$\times$$1$$\times1$). 
This binding operation, which we call \emph{position-wise HRR} (\PWHRR), is translation equivariant and maintains locality, an essential property for subsequent layers with limited receptive fields. 
More concretely, binding and unbinding are performed as
\begin{align}
(a^{(k)} \odot x^{(k)})_{ {:,w,h}} & = a^{(k)} * x^{(k)}_{:,w,h}\\
(\tilde{a}^{(k)} \oast h)_{ :,w,h} & = \tilde{a}^{(k)} \cdot h_{:,w,h},
\end{align}
with image tensors $x^{(k)} \in \mathbb{R}^{D \times W \times H}$, hidden representation $h \in \mathbb{R}^{D' \times W \times H}$, binding key $a^{(k)} \in \mathbb{R}^D$ and unbinding key $\tilde{a}^{(k)} \in \mathbb{R}^{D' \times D'}$. Here, $D, D', W,$ and $ H$ denote the hidden dimension at binding, the hidden dimension at unbinding (generally differs from $D$), image width, and image height, respectively. $*$ is the circular convolution, $\cdot$ the matrix multiplication, and $k$ indexes the superposition channel. Unbinding is applied after the global (average) pooling to reduce computational costs.
The binding keys can be either learned or fixed during training (see ablation study in Appendix~E).

\paragraph{Embracing high dimensional embeddings.}
\label{sec:channel_width_theory}
According to the Blessing of Dimensionality (see Appendix~A), random vectors quickly become quasi-orthogonal as their dimension increases. To reduce interference between protected channels, we increase the number of feature maps by adopting Wide Residual Networks~\cite{BMVC2016_87}, the most commonly used CNN architecture to achieve state-of-the-art accuracy on CIFAR100~\cite{krizhevsky2009learning}. The input tensors are passed individually through the first convolutional layer before being superposed in a suitably high dimensional space. We set the number of feature maps after the first convolutional layer to $D$=64. This is 4$\times$ more than the standard Wide-ResNet-28~\cite{BMVC2016_87}, which results in improved training stability at a marginally higher compute cost (see Section~\ref{sec:results-MIMOConv}).

\paragraph{Encouraging isometric layers.}
We aim to preserve the quasi-orthogonality of our protected channels as the superposition is passed through many network layers. To that end, residual connections are used and each subfunction $g_\theta$ of type (strided) spatial convolution or activation function is made approximately inner-product preserving, i.e.,  
\begin{equation}
    \langle g_\theta(x), g_\theta(y) \rangle \approx \langle x, y \rangle.
\end{equation}
%
%
%
Inspired by~\cite{qi2020deep}, we deploy a regularization to the CNN weights and use a parametric ReLU~\cite{he2015prelu} activation function, a learnable affine combination between identity and ReLU. 
Those adjustments lead to a near-isometric network.
The regularization for the individual CNN layers is determined by
\begin{align}
    L(W) & = \frac{\gamma}{2} \norm{Conv(W,W) - \delta^{C_o }}_F^2 & \text{where }  \quad\quad&\delta^{C_o }_{:, :, j, l}\!  =\! I_{C_o \times C_o}\! \cdot\! \mathds{1}_{j, l = \lfloor \tfrac{k}{2} \rfloor},
    \label{eq:isometry_regularization}\\
    L(W^T) & = \frac{\gamma}{2} \norm{Conv(W^T,W^T) - \delta^{C_i }}_F^2 & \text{where } \quad\quad &\delta^{C_i }_{:, :, j, l}\!  =\! I_{C_i \times C_i}\! \cdot\! \mathds{1}_{j, l = \lfloor \tfrac{k}{2} \rfloor}, \label{eq:isometry_regularization_T}
\end{align}
with $\gamma$ as hyperparameter.
We use $L(W)$ if $C_i > C_o$, else $L(W^T)$. 
See Appendix~B for more details.
 
\begin{wrapfigure}[28]{R}{0.5\textwidth}
    \vspace{10pt}
    \centering
    \includegraphics{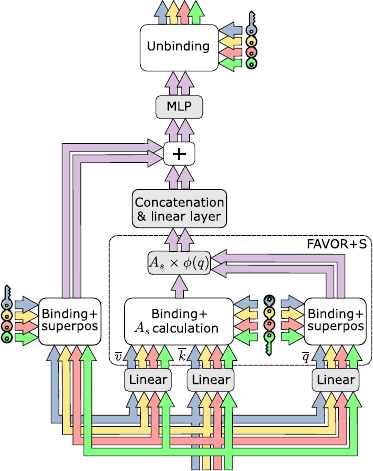}
    \caption{MIMOFormer layer applying computation in superposition to single-head FAVOR+S attention and to the MLP.
    This example passes four channels ($N\cdot M$=$2\cdot 2$=$4$) to the FAVOR+S attention (see Eq.~\eqref{eq:mimoformer_main}). 
    The individual outputs are retrieved by unbinding after the MLP. 
    The skip connection superposes the individual inputs for alignment, using the same protection keys as in unbinding.
    }
    \label{fig:mimoformer}
\end{wrapfigure}
\


\section{\MIMOFormer}\label{sec:method_transformers}
This section presents \MIMOFormer, which applies the principles of computation in superposition to dot-product self-attention~\cite{vaswani2017attention}.
%
Figure~\ref{fig:mimoformer} shows a \MIMOFormer layer with four protected channels, consisting of a single-head\footnote{The application to multi-head attention is straightforward; empirical results are shown in Section~\ref{sec:results_transformer}.} attention block, a concatenation, a linear layer, an MLP, and a skip connection.

Merely superposing protected attention keys\footnote{Protection keys are denoted by the letter $a$ and attention keys by the letter $k$ to distinguish between them.} and queries does not yield the desired result. As discussed in Appendix~F, with scalar attention scores between pairs of tokens, vanilla dot-product attention irreversibly combines attention scores of separate protected channels, effectively blurring the attention weights.
By building instead on linear Transformers~\cite{choromanski2020performer}~\cite{schlag2021fastweights}, attention scores are not collapsed to scalars, thus enabling computation in superposition.

Despite being compatible with other linear transformers (such as DPFP~\cite{schlag2021fastweights}), for concreteness we discuss changes to the Performer's FAVOR+ attention block~\cite{choromanski2020performer}. Enabling computation in superposition, we label the block as \mbox{FAVOR+S}. 

Given attention keys $(k_j)_{j=1}^L$, queries $(q_i)_{i=1}^L$, and values $(v_j)_{j=1}^L$, FAVOR+ estimates dot-product attention at sequence index $i$ through
\begin{align}
    o_i = \sum_{j=1}^L v_j \frac{\exp(\langle k_j, q_i \rangle / \sqrt{D})}{\sum_{l=1}^L \exp(\langle k_l, q_i \rangle / \sqrt{D})} \approx  \frac{1}{B_i}\underbrace{\bigg[\sum\limits_{j=1}^L v_j \phi(k_j)^T\bigg]}_{A} \times\ \phi(q_i),
    \label{method_attention_formulation}
\end{align}
where $\phi:\mathbb{R}^D \to \mathbb{R}_+^R$ approximates the softmax kernel $\exp \small(\langle k_j, q_i \rangle / \sqrt{D} )$ as an explicit inner product. 
%
%
Since computing $\phi$ has a computational complexity of $\mathcal{O}(DR)$, the construction of $A \in \mathbb{R}^{D \times R}$ takes $\mathcal{O}(L D R)$. Equally, multiplying $A \times \phi(q_i)\ \forall i$ takes $\mathcal{O}(LDR)$. Thus, FAVOR+ breaks the quadratic dependence on sequence length $L$ of attention. The denominator $B_i$ is discussed in Appendix~C. 
%
%
%

Our extension, \FAVORpS, computes attention in superposition, yielding a square-root speedup in the number of protected channels. It encodes them in an $M\times N$ grid, distributing the computational burden among the setup of the value-key matrix $A$ and its product with $\phi(q_i)$. 
Importantly, in the limit of high dimensional projections $D$ our mechanism converges to exact self-attention, completely separating the protected channels from one another. In the following, we assume the token values $\overline{v}_j$, keys $\overline{k}_j$, and queries $\overline{q}_j$ to be in protected subspaces:

\begin{equation}
    v_j^{(m,n)} := \overline{v}_j^{(m,n)} \odot a^{(m,n)}, \qquad k_j^{(m,n)} := \overline{k}_j^{(m,n)} \odot a^{(m,n)}, \qquad q_j^{(m,n)} := \overline{q}_j^{(m,n)} \odot a^{(m,n)},
\end{equation}
where $a^{(m,n)}$ are i.i.d. bipolar vectors of Rademachers~\cite{MAP_1998} and ($m,n$) denotes a channel.
%

\MIMOFormer benefits from the low time complexity ($\mathcal{O}(D)$) of the Hadamard product, especially since binding and unbinding are performed in every \MIMOFormer layer. 
%
Our derivations rely on two estimates:
\begin{equation}
    \phi(k)^T \phi(q) \overset{P}{\approx} \exp(\langle k, q \rangle / \sqrt{D}) \qquad
 \langle \sum\limits_{w=1}^N k_j^{(u, w)},\ \sum\limits_{t=1}^M q_i^{(t, n)} \rangle
 \overset{H}{\approx} \underbrace{\langle k_j^{(u, n)},\ q_i^{(u, n)} \rangle}_{\text{intended signal}}
\end{equation}
The approximation $P$, which improves with increasing $R = \dim (\phi(\overline{q}_i))$, is due to FAVOR+ and is quantified in \cite{choromanski2020performer}. On the other hand, the approximation $H$ follows from:

\begin{Interchanneldistortion}\label{informal_statement_attention_inter_channel_distortion}
    The probability that inter-channel attention distorts the intended signal of the dot-product by a factor outside $[1-\alpha, 1+\alpha]$ has various upper bounds, most notably decaying exponentially w.r.t. $D\alpha^2 \cos^2(\measuredangle( \overline{k}_j^{(u,n)}, \overline{q}_i^{(u,n)} ))/(NM-1)^2$. See Appendix~D for the full theorem.
\end{Interchanneldistortion}

\subsection{\FAVORpS: Computing self-attention in superposition}\label{sec:FAVOR+S_explanation}
We first discuss separately how to use a one-dimensional grid to carry out either the multiplication ($\times$) or the construction of $A$ in superposition.
Finally, the integration into a 2D grid will be shown. 
\paragraph{Placing multiple queries in superposition.}
We set up channels $1, \ldots, M$ by simultaneously generating a superposition in the construction of $A_s$ and of the queries to be applied. To avoid inter-channel attention we superpose value-key tensor products, i.e., we do not construct tensor products between superposed values and superposed keys:
%
\begin{align}
    S_i = &\underbrace{\bigg[\sum\limits_{j=1}^L\sum\limits_{u=1}^M  v_j^{(u)} \ \phi (k_j^{(u)})^T\bigg]}_{{A_s}} \ \ \times\ \phi( \sum\limits_{t=1}^M q_i^{(t)})
    = \sum\limits_{j=1}^L\sum\limits_{u=1}^M  v_j^{(u)} \ \bigg(\phi (k_j^{(u)})^T \ \phi (\sum\limits_{t=1}^M q_i^{(t)}) \bigg)\\
    \overset{P}{\approx} & \sum\limits_{j=1}^L\sum\limits_{u=1}^M  v_j^{(u)} \ \exp \big(\langle k_j^{(u)},\ \sum\limits_{t=1}^M q_i^{(t)} \rangle /\sqrt{D}\big)
    \overset{H}{\approx} \sum\limits_{j=1}^L\sum\limits_{u=1}^M  v_j^{(u)} \ \exp\big( \langle k_j^{(u)},\ q_i^{(u)} \rangle /\sqrt{D}\big)\\
    = & \sum\limits_{u=1}^M \underbrace{\sum\limits_{j=1}^L v_j^{(u)} \ \exp\big( \langle k_j^{(u)},\ q_i^{(u)} \rangle /\sqrt{D}\big)}_{\text{unnormalized } o_i \text{ of channel } u}. 
\end{align}
We obtain a superposition of bound output values; hence, the cost of computing $A \cross \phi(q_i)$ for all $i$ is amortized across channels. However, the construction complexity of $A_s$ is increased $M$-fold to \(\mathcal{O}(LDR\cdot M)\), hence the complexity per protected channel remains at \(\mathcal{O}(LDR)\).
%

\paragraph{Constructing value-key tensor products in superposition.}
Next, we demonstrate a value-key tensor product ($A_s$) shared across all channels, but with a mere $\mathcal{O}(L D (R+N))$ setup complexity.
In $\mathcal{O}(L N D)$ we compute $\sum_{q=1}^N v_j^{(q)}$ and $\sum_{w=1}^N k_j^{(w)}$ for all $j$. These are then (re)used to build $A_s$. 
\begin{align}
     S_i^{(n)} = & \underbrace{\bigg[\sum\limits_{j=1}^L  \big(\sum_{q=1}^N v_j^{(q)}\big)  \phi( \sum\limits_{w=1}^N k_j^{(w)})^T \bigg]}_{A_s} \times\ \phi(q_i^{(n)})
    = \sum\limits_{j=1}^L \sum_{q=1}^N v_j^{(q)}  \bigg(\phi( \sum\limits_{w=1}^N k_j^{(w)})^T \phi(q_i^{(n)}) \bigg)\\
    \overset{P}{\approx} & \sum\limits_{j=1}^L \sum_{q=1}^N v_j^{(q)}  \exp \big( \langle \sum\limits_{w=1}^N k_j^{(w)},\ q_i^{(n)} \rangle  /\sqrt{D} \big) \overset{H}{\approx} \sum\limits_{j=1}^L \sum_{q=1}^N v_j^{(q)}  \exp \big( \langle k_j^{(n)},\ q_i^{(n)} \rangle  /\sqrt{D} \big)\\
    = & \underbrace{\sum\limits_{j=1}^L v_j^{(n)} \exp \big( \langle k_j^{(n)},\ q_i^{(n)} \rangle  /\sqrt{D} \big)}_{\text{unnormalized } o_i \text{ of channel } n } + \underbrace{\sum_{q\not= n} \sum\limits_{j=1}^L v_j^{(q)} \exp \big( \langle k_j^{(n)},\ q_i^{(n)} \rangle  /\sqrt{D} \big).}_{\text{noise in separate protected subspace}}
\end{align}
The output contains the $n^{th}$ channel together with noise. 
The operation $A_s \cross \phi(q_i^{(n)})$, which takes $\mathcal{O}(L D R)$, must be repeated $N$ times to produce outputs for all $N$ channels, causing a bottleneck.

\paragraph{Simultaneous superposition of queries and value-key tensor products using a 2D grid.}\label{sec:acc_att_full}
Finally, we combine the two previously described approaches to encode the superposition channels in a 2D grid of size $N \times M$. 
We multiply a constant matrix ($A_s$) with features derived from a superposition of queries to get the superposition vector $S_i^{(n)}$\begin{equation}
    S_i^{(n)} = \underbrace{\bigg[\sum\limits_{j=1}^L \sum\limits_{u=1}^M \left(\sum_{q=1}^N v_j^{(u,q)}\right)  \phi( \sum\limits_{w=1}^N k_j^{(u, w)})^T \bigg]}_{\text{construct } A_s \text{ in } \mathcal{O}(LMD(R + N))}  \quad \times \quad \underbrace{\phi( \sum\limits_{t=1}^M q_i^{(t, n)}).}_{\mathclap{\text{construct } \forall i, n \text{ in } \mathcal{O}(LND(R + M))}}\label{eq:mimoformer_main}
\end{equation}
Computing the multiplication $\times\ \forall i,n$ takes $\mathcal{O}(L N D R)$. If we set $M$=$N$, we can evaluate $S_i^{(n)}\ \forall i, n$ using only $\mathcal{O}(L N D R + LN^2 D)$ instead of the usual $\mathcal{O}(LDR \cdot N^2)$. Thus, one may achieve a speedup of $\mathcal{O}(\min(\sqrt{N^2}, R))$ compared to FAVOR+. Since $R$ is normally in the hundreds~\cite{choromanski2020performer}, we can assume improvements of $\mathcal{O}(\sqrt{N^2})$ for reasonably large $N^2 = M \cdot N$.
Eq.~\eqref{eq:mimoformer_main} simplifies to:
\begin{align}
     S_i^{(n)} = & \sum\limits_{j, u, q} v_j^{(u, q)}  \big(\phi( \sum\limits_{w} k_j^{(u, w)})^T \phi( \sum\limits_{t} q_i^{(t, n)}) \big) \overset{P}{\approx} \sum\limits_{j, u, q} v_j^{(u, q)}  \exp \bigg( \frac{\langle \sum\limits_{w} k_j^{(u, w)},\ \sum\limits_{t} q_i^{(t, n)} \rangle}{\sqrt{D}} \bigg)\\
    \overset{H}{\approx} & \sum\limits_{j=1}^L \sum\limits_{u=1}^M \sum_{q=1}^N v_j^{(u, q)}  \exp \left( \langle k_j^{(u, n)},\ q_i^{(u, n)} \rangle  /\sqrt{D} \right)\\
    = & \sum\limits_{u=1}^M \underbrace{\sum\limits_{j=1}^L v_j^{(u, n)} \exp \bigg( \frac{\langle k_j^{(u, n)},\ q_i^{(u, n)} \rangle}{\sqrt{D}}  \bigg)}_{\text{unnormalized } o_i \text{ of channel } (u, n) } + \underbrace{\sum_{q\not= n} \sum\limits_{u=1}^M \sum\limits_{j=1}^L v_j^{(u, q)} \exp \bigg( \frac{\langle k_j^{(u, n)},\ q_i^{(u, n)} \rangle}{\sqrt{D}}\bigg).}_{\text{noise in separate protected subspace}}
\end{align}

\subsection{Integrating \FAVORpS into \MIMOFormer}
%
As is apparent in Eq.~\eqref{eq:mimoformer_main}, the query superposition is along a different axis ($M$) than the key and value superpositions ($N$). The output of attention, however, is only superposed along a single axis ($M$). 
To be able to set up superpositions along both axes ($M$ and $N$) at the next layer, we require all channels (i.e., keys, queries, and values) in separation, i.e., not superposed, at the interface between \MIMOFormer layers. 
%
%
We present two variants of \MIMOFormer with different speedups.

The first computes in superposition exclusively during the attention mechanism. 
The individual tokens of the channel ($n,m$) are directly retrieved from $S_i^{(n)}$ by unbinding with the key $\tilde{a}^{(n,m)} = a^{(n,m)}$, and the remaining computational steps within \FAVORpS are performed separately. 

The second (and faster) \MIMOFormer instance additionally performs the concatenation, the linear layer, as well as the MLP in superposition (shown in Figure~\ref{fig:mimoformer}). 
%
%
%
Unlike in the first variant, the skip connection around the attention block must account for the introduced superposition. 
%
%
To allow a potential embedding dimension mismatch,
we instantiate two different sets of randomly drawn bipolar keys: one for the skip connection and post-MLP unbinding, and one for \FAVORpS binding. 
All keys are frozen during training; it is up to the trainable weights in the linear layer after concatenation to find the relationship between the binding and unbinding keys.

The function $\phi$ in the self-attention block consists of an $R$=256 dimensional projection and a ReLU activation ~\cite{choromanski2020performer}. 
%
%
Appendix~C provides theoretical justification for using ReLU in $\phi$ and its benefits for MIMOFormer. Empirically, ReLU shows better numerical stability than unbiased softmax FAVOR+.

\section{Empirical Results}\label{sec:results}
This section evaluates the proposed \MIMONets on various model architectures and benchmarks. 
%
Appendix~E and Appendix~F describe the experimental setup for \MIMOConv and \MIMONets, respectively. 
All experiments are repeated five times with a different random seed. We report the mean and standard deviations of accuracy to account for variability in training.
\subsection{\MIMOConv}\label{sec:results-MIMOConv}

\textbf{CIFAR10 and CIFAR100.} 
%
%
Our main baseline, which we adapt to \MIMOConv, is a reproduced WideResNet-28-10~\cite{BMVC2016_87}, i.e., a 28-layer CNN with the number of feature maps of each convolutional layer enlarged by $10\times$. 
Moreover, as a stronger baseline, we include the isometry regularization in the training of WideResNet-28-10, and call the resulting network WideIsoNet-28-10. 
Both baselines demand 5.251 GMACs (Giga multiply-accumulate operations) per sample.
\MIMOConv's inference complexity per sample is $5.335$\,GMACs for $N$=$1$, $2.667$\,GMACs for $N$=$2$, and $1.334$\,GMACs for $N$=$4$; hence, we get a $\approx$$N$$\times$ speedup despite not accelerating the first and last layer.
See Appendix~E.
Table~\ref{tab:MIMOConv_comparison_dynamic_static} shows the accuracy of the static \MIMOConv, which is exclusively trained to support either 1, 2, or 4 channels.
The \MIMOConv with one channel ($N$=1) outperforms both baselines, which may be attributed to regularizing effects of the key-value binding. 
%
\MIMOConv with $N$=$2$ channels still outperforms WideResNet-28-10, while reducing the inference complexity by 2$\times$. 
%
The complexity can be further reduced by increasing the number of superpositions to $N$=$4$ at a slight accuracy drop of $\leq$3.18\%, compared to WideResNet-28-10. 

Next, we evaluate the dynamic partitioning of the superposition channels to select a speed-accuracy operating point instantaneously, which is a main feature of our approach and sets it apart from other static approaches that opt for a fixed performance point like model downsizing, quantization aware training, and pruning (see Appendix~A for a discussion).
We set up a model with four channels, but evaluate its performance in different configurations: a fast (4 inputs/pass), a normal (2 inputs/pass), and a slow mode (1 input/pass). 
The fast mode maps each input to one channel; the medium mode distributes two inputs over pairs of channels; and the slow mode uses all channels for the same input.
The models are trained on 80\% of the batches in fast mode and on 20\% of the batches in slow mode.
Appendix~E provides more details on the trade-off between fast and slow mode training.
As Table~\ref{tab:MIMOConv_comparison_dynamic_static} shows, a single dynamic model can seamlessly switch between operation points while maintaining a high accuracy compared to the static models ($\leq$$1.82\%$ drop). 

\begin{table}[h]
\centering
 \caption{Average accuracy (\%) of WideResNet-28-10 variants and our \MIMOConv. Static models are trained to process $N$ inputs in one pass, speeding up inference by $N\times$. Dynamic models are trained with a variable number of inputs ($N$=$1$--$4$), and can process a variable number of inputs per pass. We report the average accuracy $\pm$ the standard deviation over five runs with different seeds.}
  \label{tab:MIMOConv_comparison_dynamic_static}
  \resizebox{\linewidth}{!}{
\begin{tabular}{lcccccc}
\toprule
        & \multicolumn{3}{c}{CIFAR10} & \multicolumn{3}{c}{CIFAR100} \\
\cmidrule(r){2-4}\cmidrule(r){5-7}
 \# inputs/pass        & 1    & 2   & 4  & 1  & 2   & 4   \\
\cmidrule(r){1-1}\cmidrule(r){2-4}\cmidrule(r){5-7}
WideResNet-28-10 & $96.82^{\pm0.06}$  &     n.a.      &   n.a.    &    $81.62^{\pm0.07}$     &     n.a.      &   n.a.  \\
WideIsoNet-28-10 & $97.31^{\pm0.11}$  &     n.a.      &   n.a.     &    $82.38^{\pm0.20}$     &     n.a.      &   n.a.  \\
\cmidrule(r){1-1}\cmidrule(r){2-4}\cmidrule(r){5-7}
MIMOConv static ($N$=$1$)  &    $97.49^{\pm0.08}$  &     n.a.      &  n.a.     &    $83.19^{\pm0.17}$     &     n.a.      &   n.a.      \\
MIMOConv static ($N$=$2$)  &  n.a.       &   $96.93^{\pm 0.13}$        &    n.a.    &     n.a.    & $82.30^{\pm0.19}$       &    n.a.     \\
MIMOConv static ($N$=$4$)  &  n.a.       &    n.a.       &  $95.58^{\pm0.23}$      &    n.a.     &    n.a.       &    $78.44^{\pm0.30}$     \\
\cmidrule(r){1-1}\cmidrule(r){2-4}\cmidrule(r){5-7}
MIMOConv dynamic ($N$=$1$--$4$) &   $97.13^{\pm0.11}$      &   $96.41^{\pm0.14}$        &   $95.43^{\pm0.07}$      &  $82.52^{\pm0.09}$        &    $80.48^{\pm0.08}$        &  $78.19^{\pm0.10}$ \\
\bottomrule
\end{tabular}
}
\end{table}

The detailed ablation study in Appendix~E makes the following findings: 
(1)~isometry regularization improves accuracy for any number of channels; (2)~training \MIMOConv for more epochs closes the performance gap to the single-input baseline; (3)~an appropriate number of feature maps (32 or 64) in the first layer stabilizes training; and~(4) the binding keys can be frozen during training without loss (whereas unbinding keys are never frozen). 

\begin{minipage}{0.49\textwidth}
\paragraph{MNIST and SVHN.}
Figure~\ref{fig:mnist} compares \MIMOConv with DataMux~\cite{murahari2022datamux} on the MNIST dataset
. Even with a trivial downsizing for fair comparison from a 28-layer very-wide ($10\times$) ResNet to a 10-layer narrow ($1\times$) network, \MIMOConv scales much better to high superposition channels ($N$) than DataMUX does. Indeed, our model shows an accuracy of $80.4\%$ against their $52.9\%$ in case of $N$=$16$ superposition channels (highest number of channels reported by DataMUX for vision tasks), despite being computationally cheaper ($0.47$\,MMAC/s vs. $0.65$\,MMAC/s). Also, DataMux's binding overhead results in a mere $1.35\times$ reduction in MACs compared to our $10.9\times$ as $N$ goes from $1$ to $16$. Ergo, our method scales better in accuracy and throughput as $N$ increases.
\end{minipage}\hfill
\begin{minipage}{0.49\textwidth}
    \centering
    \includegraphics[width=\textwidth]{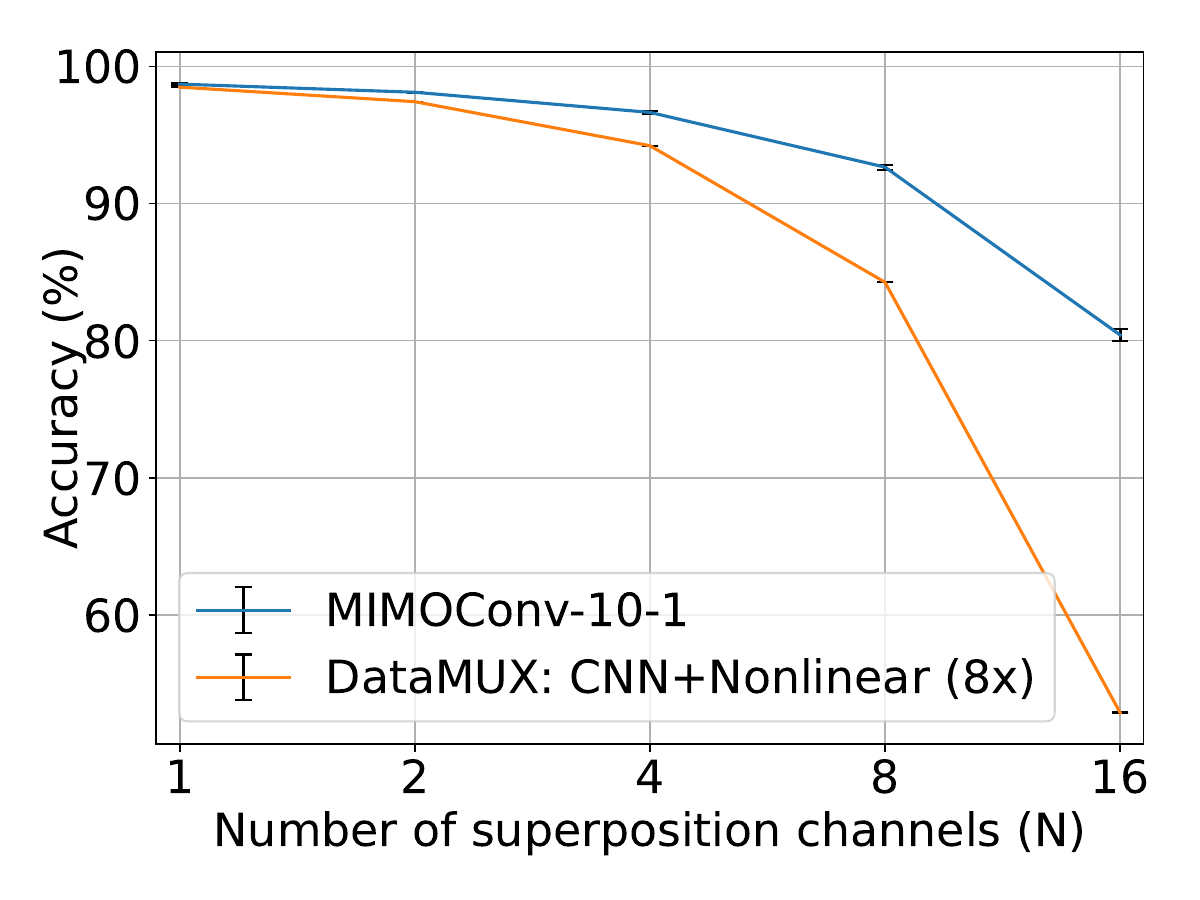}
    \vspace{-.4cm}
    \captionof{figure}{Classification accuracy (\%) on MNIST for downsized model.
    }
    \label{fig:mnist}
\end{minipage}
\

Finally, we tested \MIMOConv on the SVHN dataset. Despite limited hyperparameter tuning, MIMOConv achieves a high accuracy of $97.17\%$ ($N$=1), and can maintain its performance with larger superpositions ($97.05\%$ and $96.84\%$ for $N$=$2$ and $N$=$4$, respectively).

%


%

%

\subsection{\MIMOFormer}\label{sec:results_transformer}
\textbf{LRA.} We evaluate \MIMOFormer on five tasks from LRA~\cite{tay2021long}, and compare against the vanilla Transformer~\cite{vaswani2017attention} and the Performer~\cite{choromanski2020performer} using FAVOR+ attention with ReLU projection. 
Moreover, we also consider wide Transformer variants~\cite{brown2022wide}, consisting of only one layer but as many heads as their deep counterparts. 
%
%
Task-specific architectures and training hyperparameters are kept the same for the Performer and the \MIMOFormer (see Appendix F). 

\begin{table}[h] 
\centering
\caption{Test accuracy (\%) on the long range arena (LRA). \MIMOFormer uses an equal number of query superpositions ($M$) and value-key superpositions ($N$), i.e., $N$=$M$. Computation in superposition is performed either in attention only (att.) or in both attention and MLP (att.+MLP). $L$~is the number of layers, $H$ the number of heads, and $^*$ indicates curriculum learning.}
\label{tab:lra}
\resizebox{\linewidth}{!}{
\begin{tabular}{lllllll}
\toprule
                        & \multicolumn{1}{c}{ListOps} & \multicolumn{1}{c}{Text}  & \multicolumn{1}{c}{Retrieval} & \multicolumn{1}{c}{Image} & \multicolumn{1}{c}{Pathfinder} & \multicolumn{1}{c}{Avg.}  \\
\cmidrule(r){1-1}\cmidrule(r){2-7}
\textbf{Deep models} & $L$=$6$, $H$=$8$ & $L$=$6$, $H$=$8$ & $L$=$4$, $H$=$4$ & $L$=$3$, $H$=$4$ & $L$=$4$, $H$=$8$ &\\
\cmidrule(r){1-1}\cmidrule(r){2-7}
Transformer~\cite{vaswani2017attention}             & $36.37$   & $64.27$ & $57.46$     & $42.44$ & $71.40$      & $53.39$ \\
Performer~\cite{choromanski2020performer}               & $18.01$   & $65.40$ & $53.82$     & $42.77$ & $77.05$      & $51.41$ \\
Performer (reproduced)        &  $38.94^{\pm0.23}$	&	$65.70^{\pm0.31}$	&	$81.58^{\pm0.18}$	&	$40.14^{\pm0.86}$	&	$73.82^{\pm0.78}$	&	$60.04^{\pm0.47}$\\					
MIMOFormer (N=2, att.) &  $38.08^{\pm0.21}$	&	$65.00^{\pm0.28}$	&	$79.37^{\pm0.81}$	&	$38.21^{\pm0.63}$	&	$72.36^{\pm0.54}$	&	$58.61^{\pm0.49}$\\
MIMOFormer (N=2, att.+MLP) &  $37.65^{\pm0.33}$	&	$64.39^{\pm0.22}$	&	$76.02^{\pm0.27}$	&	$33.85^{\pm0.55}$	& 
$67.98^{\pm0.47}$	&	$55.98^{\pm0.37}$\\
MIMOFormer (N=4, att.) & $37.22^{\pm0.33}$	&	$64.59^{\pm0.14}$	&	$60.99^{\pm9.06}$	&	$28.16^{\pm0.08}$	&	$55.50^{\pm4.95}$	&	
$49.29^{\pm2.91}$ \\	
MIMOFormer (N=4, att.)$^*$ & $37.64^{\pm0.73}$	&	$64.46^{\pm0.15}$	&	$74.38^{\pm0.82}$	&	$30.52^{\pm0.77}$	&	$67.10^{\pm0.45}$	&	$ 54.82^{\pm0.58}$ \\	
MIMOFormer (N=4, att.+MLP) & $17.74^{\pm0.63}$	&	$60.71^{\pm5.14}$	&	$72.20^{\pm0.28}$	&	$24.01^{\pm0.47}$	&	$50.33^{\pm0.16}$	&	$45.00^{\pm1.34}$ \\
\cmidrule(r){1-1}\cmidrule(r){2-7}
\textbf{Wide models} & $L$=$1$, $H$=$48$ & $L$=$1$, $H$=$48$ & $L$=$1$, $H$=$16$ & $L$=$1$, $H$=$12$ & $L$=$1$, $H$=$32$ &\\
\cmidrule(r){1-1}\cmidrule(r){2-7}
Performer (reproduced)    &  $39.40^{\pm0.51}$	&	$65.73^{\pm0.32}$	&	$83.67^{\pm0.25}$	&	$41.67^{\pm0.44}$	&	$74.11^{\pm0.33}$	&	$60.93^{\pm0.37}$ \\
MIMOFormer (N=2, att.) & $38.90^{\pm0.53}$	&	$65.39^{\pm0.18}$	&	$81.27^{\pm0.28}$	&	$40.25^{\pm0.21}$	&	$73.51^{\pm0.23}$	&	$59.86^{\pm0.29}$\\
MIMOFormer (N=2, att.+MLP) & $37.59^{\pm0.17}$	&	$64.64^{\pm0.25}$	&	$78.30^{\pm0.32}$	&	$36.69^{\pm0.76}$	&	$68.22^{\pm0.18}$	&	$57.09^{\pm0.34}$ \\					
MIMOFormer (N=4, att.) & $37.71^{\pm0.24}$	&		$64.22^{\pm0.14}$	&		$74.99^{\pm0.36}$	&		$35.43^{\pm0.60}$	&		$69.52^{\pm0.40}$	&		$56.37^{\pm0.35}$ \\
MIMOFormer (N=4, att.)$^*$ & $37.68^{\pm0.36}$	&	
$64.56^{\pm0.25}$	&	$76.37^{\pm0.50}$	&	$35.53^{\pm0.48}$	&	
$73.37^{\pm0.22}$   & 
$57.50^{\pm0.36}$	\\
MIMOFormer (N=4, att.+MLP) & $18. 52^{\pm0.98}$	&	$63.53^{\pm0.12}$	&	$74.30^{\pm0.26}$	&	$26.54^{\pm0.28}$	&	$56.33^{\pm0.17}$	&	$47.84^{\pm0.36}$ \\		
\bottomrule
\end{tabular}
}
\end{table}

Owing to an improved training setup, our replicated deep and wide Performer baselines substantially outperform the results reported in~\cite{tay2021long} (see Table~\ref{tab:lra}).
Moreover, \MIMOFormer enables accurate computation in superposition for both deep and wide attention models.
%
%
%
The performance drop is less pronounced in wide models (only $1.07\%$ drop compared to Performer with $N$=$2$, att.), which may be attributed to the larger number of heads, increasing the effective dimension ($D_{\mathrm{tot}}= H \cdot D_{\mathrm{head}}$). 

When computing both attention and the MLP in superposition (att.+MLP), we observe better scaling (in $N$) for wide models. Also, \MIMOFormer reduces the gap to the baseline as the number of epochs increases (see Appendix F).

To stabilize training in the case of $N$=$4$, we implemented a curriculum training procedure where the number of superpositions is reduced to $N$'=$N/2$ during a warmup phase (1/6th of the training steps), improving  
%
%
the average accuracy of MIMOFormer in both wide and deep models.
%

Comparing against a reproduced DataMUX~\cite{murahari2022datamux}, MIMOFormer (att.) outperforms it on ListOps (38.08\% vs. 30.54\% accuracy) when using models of similar size and $N$=2, see Appendix~F.

\paragraph{Synthetic sequence modeling.}
\begin{wraptable}[9]{R}{0.57\textwidth}
 \vspace{-.4cm}
  \captionof{table}{Accuracy (\%) on synthetic sequence modelling.}
  \label{tab:synthetic}
 \resizebox{0.57\textwidth}{!}{
    \begin{tabular}{llrr}
    \toprule
                   Architecture &     Attention & \multicolumn{1}{c}{\begin{tabular}[c]{@{}c@{}}Associative \\ recall~\cite{fu2022h3}\end{tabular}} & \multicolumn{1}{l}{\begin{tabular}[c]{@{}c@{}}Induction \\ head~\cite{fu2022h3}\end{tabular}} \\
    \cmidrule(r){1-2}\cmidrule(r){3-4}
    
    Transformer     & Softmax      &  $98.48^{\pm1.87}$                                                                                & $100^{\pm0.0}$                                                                           \\
    Performer      & FAVOR+       & $96.32^{\pm6.26}$                                                                               & $31.58^{\pm33.67}$                                                                          \\
     \cmidrule(r){1-2}\cmidrule(r){3-4}
    MIMOFormer (N=2, att.) & FAVOR+  & $96.52^{\pm2.79}$                                                                              & $99.40^{\pm0.13}$                                                                          \\
    MIMOFormer (N=2, att.) & DPFP~\cite{schlag2021fastweights}  & $93.64^{\pm 12.66}$ & $98.56^{\pm 0.86}$                                                                          \\
    DataMUX (N=2)~\cite{murahari2022datamux}    & Softmax             & $20.04^{\pm 1.72}$& $6.06^{\pm 2.24}$        \\
    \bottomrule
    \end{tabular}
    }
\end{wraptable}	

Table~\ref{tab:synthetic} reports the accuracy on two synthetic sequence modeling tasks, which Transformer alternatives such as S4 ~\cite{gu2022efficiently} have difficulties solving ~\cite{fu2022h3}. 
On these more nuanced NLP tasks, the accuracy of DataMUX~\cite{murahari2022datamux} drops to $20.04\%$ and $6.06\%$ for $N$=$2$ despite significant efforts in training, while MIMOFormer, at a score of $96.52\%$ and $99.40\%$ respectively, succeeds. We attribute this difference in performance to attention score blurring in DataMux, discussed in Appendix~F. Contrastingly, our method converges to exact attention without blurring.
It is versatile and can be adjusted to other linear Transformers such as DPFP~\cite{schlag2021fastweights}, achieving a score of $93.64\%$ and $98.56\%$.

\section{Related Work}

%
So far, superposition principles have been applied in order to store the weights of multiple models in a single neural network~\cite{cheung2019superposition, hersche2020compressing,zeman2021compressed}, to circumvent catastrophic forgetting in continual learning~\cite{wortsman2020supermasks, zeman2023superformer}, and to render symbolic reasoning tractable~\cite{Hersche_NMI2023}.
%
%
%
%
To address privacy concerns when running remote inference, single inputs were bound with random channels to implement pseudo-encryption~\cite{alam2022deploying}.
%
%
%
Recently in~\cite{alam2022hrrformer}, HRR was used to define an unconventional version of self-attention, whose attention scores are processed to a diagonal matrix. The value vectors are scaled according to their importance in the sequence instead of being combined in a weighted sum. 
%
In contrast to us, none of these works superpose multiple inputs into a data structure to speed up computation. 

In~\cite{havasi2021training, rame2021mixmo}, an ensemble of CNN models was fit into one network. 
However, by only broadcasting a single input over the channels and by averaging all the outputs, this approach collapses to a single-input-single-output (SISO) network. 
On the contrary, we explore using protected channels for \textit{different} inputs at inference, resulting in an actual multiple-input-multiple-output (MIMO) network.


%
%

There has also been a line of work to accelerate Transformers using inputs in superposition~\cite{murahari2022datamux}~\cite{murahari2023muxplms}. DataMux~\cite{murahari2022datamux} claims to retain high performance for language understanding tasks, even when using up to $40$ inputs in superposition. However, none of the reported tasks require attention layers at all~\cite{hassid2022does}. 
In Section~5.2 we show failure of their method when actual attention is required (see also Appendix~F). MUX-PLMs~\cite{murahari2023muxplms} improves on DataMux with contextual binding and replaces token prefixes with unbinding keys, but does not address the blurry attention mechanism. In contrast to DataMUX and MUX-PLMs, our work approximates true attention and our theoretical derivations show convergence to actual dot-product attention as the dimension of attention projections increases, giving us an even stronger case for applicability to large language models.

%

\section{Conclusion}
We present \MIMONets that simultaneously process multiple inputs by performing computation in superposition. 
\MIMONets bind arbitrary inputs with high-dimensional keys, which projects them to orthogonal subspaces that, together with near-isometric subfunctions, guarantee low interference through all nonlinear layers.  
Unbinding with (learned) keys can safely retrieve information on individual channels.
We provide two \MIMONets instances, \MIMOConv and \MIMOFormer, that show the effectiveness of computation in superposition through two dominant operations in neural network architectures: convolution and attention. 
Further investigations could 
%
explore the MIMO-capability of architectures that contain additional nonlinearities (e.g., max-pooling) and use different input modalities. 
%
\MIMONets could be suitable candidates to accelerate dynamically and on-demand the inference of foundation models~\cite{bommasani2021foundation}.

%

\section*{Acknowledgement}
This work is supported by the Swiss National Science foundation (SNF), grant 200800.
We thank Dario Bolli for conducting initial experiments and Aleksandar Terzić for helping with ablation studies.

\bibliographystyle{IEEEtran}
\bibliography{references} 

\begin{thebibliography}{10}
\providecommand{\url}[1]{#1}
\csname url@samestyle\endcsname
\providecommand{\newblock}{\relax}
\providecommand{\bibinfo}[2]{#2}
\providecommand{\BIBentrySTDinterwordspacing}{\spaceskip=0pt\relax}
\providecommand{\BIBentryALTinterwordstretchfactor}{4}
\providecommand{\BIBentryALTinterwordspacing}{\spaceskip=\fontdimen2\font plus
\BIBentryALTinterwordstretchfactor\fontdimen3\font minus
  \fontdimen4\font\relax}
\providecommand{\BIBforeignlanguage}[2]{{%
\expandafter\ifx\csname l@#1\endcsname\relax
\typeout{** WARNING: IEEEtran.bst: No hyphenation pattern has been}%
\typeout{** loaded for the language `#1'. Using the pattern for}%
\typeout{** the default language instead.}%
\else
\language=\csname l@#1\endcsname
\fi
#2}}
\providecommand{\BIBdecl}{\relax}
\BIBdecl

\bibitem{devlin2018bert}
J.~Devlin, M.-W. Chang, K.~Lee, and K.~Toutanova, ``Bert: Pre-training of deep
  bidirectional transformers for language understanding,'' \emph{arXiv preprint
  arXiv:1810.04805}, 2018.

\bibitem{brown2020gpt3}
T.~Brown, B.~Mann, N.~Ryder, M.~Subbiah, J.~D. Kaplan, P.~Dhariwal,
  A.~Neelakantan, P.~Shyam, G.~Sastry, A.~Askell \emph{et~al.}, ``Language
  models are few-shot learners,'' \emph{Advances in Neural Information
  Processing Systems (NeurIPS)}, vol.~33, pp. 1877--1901, 2020.

\bibitem{Liu_2021_ICCV}
Z.~Liu, Y.~Lin, Y.~Cao, H.~Hu, Y.~Wei, Z.~Zhang, S.~Lin, and B.~Guo, ``Swin
  transformer: Hierarchical vision transformer using shifted windows,'' in
  \emph{Proceedings of the IEEE/CVF International Conference on Computer Vision
  (ICCV)}, October 2021, pp. 10\,012--10\,022.

\bibitem{liu2022convnet}
Z.~Liu, H.~Mao, C.-Y. Wu, C.~Feichtenhofer, T.~Darrell, and S.~Xie, ``A convnet
  for the 2020s,'' in \emph{Proceedings of the IEEE/CVF Conference on Computer
  Vision and Pattern Recognition (CVPR)}, 2022, pp. 11\,976--11\,986.

\bibitem{VSA_03}
R.~W. Gayler, ``Vector symbolic architectures answer {J}ackendoff's challenges
  for cognitive neuroscience,'' in \emph{{Joint International Conference on
  Cognitive Science (ICCS/ASCS)}}, 2003.

\bibitem{PlateHolographic1995}
T.~A. Plate, ``Holographic reduced representations,'' \emph{{IEEE Transactions
  on Neural Networks}}, vol.~6, no.~3, pp. 623--641, 1995.

\bibitem{Kanerva2009}
P.~Kanerva, ``Hyperdimensional computing: An introduction to computing in
  distributed representation with high-dimensional random vectors,''
  \emph{Cognitive Computation}, vol.~1, no.~2, pp. 139--159, 2009.

\bibitem{NeumannTransformation2000}
J.~Neumann, ``Learning holistic transformation of {HRR} from examples,'' in
  \emph{{International Conference on Knowledge-Based Intelligent Engineering
  Systems and Allied Technologies (KES)}}, 2000, pp. 557--560.

\bibitem{NeumannTransformation2002}
------, ``Learning the systematic transformation of holographic reduced
  representations,'' \emph{Cognitive Systems Research}, vol.~3, no.~2, pp.
  227--235, 2002.

\bibitem{kleyko2022vector}
D.~Kleyko, M.~Davies, E.~P. Frady, P.~Kanerva, S.~J. Kent, B.~A. Olshausen,
  E.~Osipov, J.~M. Rabaey, D.~A. Rachkovskij, A.~Rahimi, and F.~T. Sommer,
  ``Vector symbolic architectures as a computing framework for emerging
  hardware,'' \emph{Proceedings of the IEEE}, vol. 110, no.~10, pp. 1538--1571,
  2022.

\bibitem{BMVC2016_87}
S.~Zagoruyko and N.~Komodakis, ``Wide residual networks,'' in \emph{Proceedings
  of the British Machine Vision Conference (BMVC)}.\hskip 1em plus 0.5em minus
  0.4em\relax BMVA Press, September 2016, pp. 87.1--87.12.

\bibitem{fu2022h3}
D.~Y. Fu, T.~Dao, K.~K. Saab, A.~W. Thomas, A.~Rudra, and C.~Re, ``Hungry
  hungry hippos: Towards language modeling with state space models,'' in
  \emph{The Eleventh International Conference on Learning Representations
  (ICLR)}, 2022.

\bibitem{murahari2022datamux}
V.~Murahari, C.~Jimenez, R.~Yang, and K.~Narasimhan, ``{DataMUX}: Data
  multiplexing for neural networks,'' \emph{Advances in Neural Information
  Processing Systems (NeurIPS)}, vol.~35, pp. 17\,515--17\,527, 2022.

\bibitem{tay2021long}
Y.~Tay, M.~Dehghani, S.~Abnar, Y.~Shen, D.~Bahri, P.~Pham, J.~Rao, L.~Yang,
  S.~Ruder, and D.~Metzler, ``Long range arena: A benchmark for efficient
  transformers,'' in \emph{International Conference on Learning Representations
  (ICLR)}, 2021.

\bibitem{choromanski2020performer}
K.~M. Choromanski, V.~Likhosherstov, D.~Dohan, X.~Song, A.~Gane, T.~Sarlos,
  P.~Hawkins, J.~Q. Davis, A.~Mohiuddin, L.~Kaiser \emph{et~al.}, ``Rethinking
  attention with performers,'' in \emph{International Conference on Learning
  Representations (ICLR)}, 2020.

\bibitem{gallant2013mbat}
S.~I. Gallant and T.~W. Okaywe, ``Representing objects, relations, and
  sequences,'' \emph{Neural Computation}, vol.~25, no.~8, pp. 2038--2078, 2013.

\bibitem{MAP_1998}
R.~W. Gayler, ``Multiplicative binding, representation operators \& analogy,''
  in \emph{Advances in Analogy Research: Integration of Theory and Data from
  the Cognitive, Computational, and Neural Sciences}, 1998.

\bibitem{kleyko2023survey}
D.~Kleyko, D.~Rachkovskij, E.~Osipov, and A.~Rahimi, ``A survey on
  hyperdimensional computing aka vector symbolic architectures, part {I}:
  Models and data transformations,'' \emph{ACM Computing Surveys}, vol.~55,
  no.~6, 2022.

\bibitem{havasi2021training}
M.~Havasi, R.~Jenatton, S.~Fort, J.~Z. Liu, J.~Snoek, B.~Lakshminarayanan,
  A.~M. Dai, and D.~Tran, ``Training independent subnetworks for robust
  prediction,'' in \emph{International Conference on Learning Representations
  (ICLR)}, 2021.

\bibitem{rame2021mixmo}
A.~Ram{\'e}, R.~Sun, and M.~Cord, ``Mixmo: Mixing multiple inputs for multiple
  outputs via deep subnetworks,'' in \emph{Proceedings of the IEEE/CVF
  International Conference on Computer Vision (CVPR)}, 2021, pp. 823--833.

\bibitem{krizhevsky2009learning}
A.~Krizhevsky, ``Learning multiple layers of features from tiny images,''
  \emph{University of Toronto}, 2009.

\bibitem{qi2020deep}
H.~Qi, C.~You, X.~Wang, Y.~Ma, and J.~Malik, ``Deep isometric learning for
  visual recognition,'' in \emph{International Conference on Machine Learning
  (ICML)}.\hskip 1em plus 0.5em minus 0.4em\relax PMLR, 2020, pp. 7824--7835.

\bibitem{he2015prelu}
K.~He, X.~Zhang, S.~Ren, and J.~Sun, ``Delving deep into rectifiers: Surpassing
  human-level performance on imagenet classification,'' in \emph{Proceedings of
  the IEEE International Conference on Computer Vision (ICCV)}, 2015, pp.
  1026--1034.

\bibitem{vaswani2017attention}
A.~Vaswani, N.~Shazeer, N.~Parmar, J.~Uszkoreit, L.~Jones, A.~N. Gomez,
  {\L}.~Kaiser, and I.~Polosukhin, ``Attention is all you need,''
  \emph{Advances in Neural Information Processing Systems (NeurIPS)}, vol.~30,
  2017.

\bibitem{schlag2021fastweights}
I.~Schlag, K.~Irie, and J.~Schmidhuber, ``Linear transformers are secretly fast
  weight programmers,'' in \emph{International Conference on Machine Learning
  (ICML)}.\hskip 1em plus 0.5em minus 0.4em\relax PMLR, 2021, pp. 9355--9366.

\bibitem{brown2022wide}
J.~R. Brown, Y.~Zhao, I.~Shumailov, and R.~D. Mullins, ``Wide attention is the
  way forward for transformers?'' in \emph{NeurIPS '22 Workshop on All Things
  Attention: Bridging Different Perspectives on Attention}, 2022.

\bibitem{gu2022efficiently}
A.~Gu, K.~Goel, and C.~Re, ``Efficiently modeling long sequences with
  structured state spaces,'' in \emph{International Conference on Learning
  Representations (ICLR)}, 2022.

\bibitem{cheung2019superposition}
B.~Cheung, A.~Terekhov, Y.~Chen, P.~Agrawal, and B.~Olshausen, ``Superposition
  of many models into one,'' \emph{Advances in Neural Information Processing
  Systems (NeurIPS)}, vol.~32, 2019.

\bibitem{hersche2020compressing}
M.~Hersche, P.~Rupp, L.~Benini, and A.~Rahimi, ``Compressing subject-specific
  brain-computer interface models into one model by superposition in
  hyperdimensional space,'' in \emph{2020 Design, Automation \& Test in Europe
  Conference \& Exhibition (DATE)}.\hskip 1em plus 0.5em minus 0.4em\relax
  IEEE, 2020, pp. 246--251.

\bibitem{zeman2021compressed}
M.~Zeman, E.~Osipov, and Z.~Bosni{\'c}, ``Compressed superposition of neural
  networks for deep learning in edge computing,'' in \emph{2021 International
  Joint Conference on Neural Networks (IJCNN)}.\hskip 1em plus 0.5em minus
  0.4em\relax IEEE, 2021, pp. 1--8.

\bibitem{wortsman2020supermasks}
M.~Wortsman, V.~Ramanujan, R.~Liu, A.~Kembhavi, M.~Rastegari, J.~Yosinski, and
  A.~Farhadi, ``Supermasks in superposition,'' \emph{Advances in Neural
  Information Processing Systems (NeurIPS)}, vol.~33, pp. 15\,173--15\,184,
  2020.

\bibitem{zeman2023superformer}
M.~Zeman, J.~F. Pucer, I.~Kononenko, and Z.~Bosni{\'c}, ``Superformer:
  Continual learning superposition method for text classification,''
  \emph{Neural Networks}, vol. 161, pp. 418--436, 2023.

\bibitem{Hersche_NMI2023}
M.~Hersche, M.~Zeqiri, L.~Benini, A.~Sebastian, and A.~Rahimi, ``A
  neuro-vector-symbolic architecture for solving {R}aven's progressive
  matrices,'' \emph{Nature Machine Intelligence}, vol.~5, no.~4, pp. 363--375,
  2023.

\bibitem{alam2022deploying}
M.~M. Alam, E.~Raff, T.~Oates, and J.~Holt, ``Deploying convolutional networks
  on untrusted platforms using {2D} holographic reduced representations,'' in
  \emph{International Conference on Machine Learning (ICML)}.\hskip 1em plus
  0.5em minus 0.4em\relax PMLR, 2022, pp. 367--393.

\bibitem{alam2022hrrformer}
------, ``Recasting self-attention with holographic reduced representations,''
  in \emph{Proceedings of 8TH SIGKDD International Workshop on Mining and
  Learning from Time Series -- Deep Forecasting: Models, Interpretability, and
  Applications (MiLeTS 2022)}, 2022.

\bibitem{murahari2023muxplms}
Y.~Su, V.~Murahari, K.~Narasimhan, and K.~Li, ``{PruMUX}: Augmenting data
  multiplexing with model compression,'' \emph{arXiv preprint
  arXiv:2305.14706}, 2023.

\bibitem{hassid2022does}
M.~Hassid, H.~Peng, D.~Rotem, J.~Kasai, I.~Montero, N.~A. Smith, and
  R.~Schwartz, ``How much does attention actually attend? {Q}uestioning the
  importance of attention in pretrained transformers,'' in \emph{Findings of
  the Association for Computational Linguistics: EMNLP 2022}, 2022, pp.
  1403--1416.

\bibitem{bommasani2021foundation}
R.~Bommasani, D.~A. Hudson, E.~Adeli, R.~Altman, S.~Arora, S.~von Arx, M.~S.
  Bernstein, J.~Bohg, A.~Bosselut, E.~Brunskill \emph{et~al.}, ``On the
  opportunities and risks of foundation models,'' \emph{arXiv preprint
  arXiv:2108.07258}, 2021.

\end{thebibliography}


\begin{thebibliography}{10}
\providecommand{\url}[1]{#1}
\csname url@samestyle\endcsname
\providecommand{\newblock}{\relax}
\providecommand{\bibinfo}[2]{#2}
\providecommand{\BIBentrySTDinterwordspacing}{\spaceskip=0pt\relax}
\providecommand{\BIBentryALTinterwordstretchfactor}{4}
\providecommand{\BIBentryALTinterwordspacing}{\spaceskip=\fontdimen2\font plus
\BIBentryALTinterwordstretchfactor\fontdimen3\font minus
  \fontdimen4\font\relax}
\providecommand{\BIBforeignlanguage}[2]{{%
\expandafter\ifx\csname l@#1\endcsname\relax
\typeout{** WARNING: IEEEtran.bst: No hyphenation pattern has been}%
\typeout{** loaded for the language `#1'. Using the pattern for}%
\typeout{** the default language instead.}%
\else
\language=\csname l@#1\endcsname
\fi
#2}}
\providecommand{\BIBdecl}{\relax}
\BIBdecl

\bibitem{kleyko2023survey}
D.~Kleyko, D.~Rachkovskij, E.~Osipov, and A.~Rahimi, ``A survey on
  hyperdimensional computing aka vector symbolic architectures, part {I}:
  Models and data transformations,'' \emph{ACM Computing Surveys}, vol.~55,
  no.~6, 2022.

\bibitem{PlateHolographic1995}
T.~A. Plate, ``Holographic reduced representations,'' \emph{{IEEE Transactions
  on Neural Networks}}, vol.~6, no.~3, pp. 623--641, 1995.

\bibitem{gallant2013mbat}
S.~I. Gallant and T.~W. Okaywe, ``Representing objects, relations, and
  sequences,'' \emph{Neural Computation}, vol.~25, no.~8, pp. 2038--2078, 2013.

\bibitem{MAP_1998}
R.~W. Gayler, ``Multiplicative binding, representation operators \& analogy,''
  in \emph{Advances in Analogy Research: Integration of Theory and Data from
  the Cognitive, Computational, and Neural Sciences}, 1998.

\bibitem{Hu2020Provable}
W.~Hu, L.~Xiao, and J.~Pennington, ``Provable benefit of orthogonal
  initialization in optimizing deep linear networks,'' in \emph{International
  Conference on Learning Representations}, 2020.

\bibitem{he2015prelu}
K.~He, X.~Zhang, S.~Ren, and J.~Sun, ``Delving deep into rectifiers: Surpassing
  human-level performance on imagenet classification,'' in \emph{Proceedings of
  the IEEE International Conference on Computer Vision (ICCV)}, 2015, pp.
  1026--1034.

\bibitem{qi2020deep}
H.~Qi, C.~You, X.~Wang, Y.~Ma, and J.~Malik, ``Deep isometric learning for
  visual recognition,'' in \emph{International Conference on Machine Learning
  (ICML)}.\hskip 1em plus 0.5em minus 0.4em\relax PMLR, 2020, pp. 7824--7835.

\bibitem{choromanski2020performer}
K.~M. Choromanski, V.~Likhosherstov, D.~Dohan, X.~Song, A.~Gane, T.~Sarlos,
  P.~Hawkins, J.~Q. Davis, A.~Mohiuddin, L.~Kaiser \emph{et~al.}, ``Rethinking
  attention with performers,'' in \emph{International Conference on Learning
  Representations (ICLR)}, 2020.

\bibitem{krizhevsky2009learning}
A.~Krizhevsky, ``Learning multiple layers of features from tiny images,''
  \emph{University of Toronto}, 2009.

\bibitem{lecun1998gradient}
Y.~LeCun, L.~Bottou, Y.~Bengio, and P.~Haffner, ``Gradient-based learning
  applied to document recognition,'' \emph{Proceedings of the IEEE}, vol.~86,
  no.~11, pp. 2278--2324, 1998.

\bibitem{yuval2011reading}
N.~Yuval, ``Reading digits in natural images with unsupervised feature
  learning,'' in \emph{Proceedings of the NeurIPS Workshop on Deep Learning and
  Unsupervised Feature Learning}, 2011.

\bibitem{smith2019clr}
L.~N. Smith and N.~Topin, ``Super-convergence: Very fast training of neural
  networks using large learning rates,'' in \emph{Artificial intelligence and
  machine learning for multi-domain operations applications}, vol. 11006, 2019,
  pp. 369--386.

\bibitem{zhang2018mixup}
H.~Zhang, M.~Cisse, Y.~N. Dauphin, and D.~Lopez-Paz, ``mixup: Beyond empirical
  risk minimization,'' in \emph{International Conference on Learning
  Representations (ICLR)}, 2018.

\bibitem{murahari2022datamux}
V.~Murahari, C.~Jimenez, R.~Yang, and K.~Narasimhan, ``{DataMUX}: Data
  multiplexing for neural networks,'' \emph{Advances in Neural Information
  Processing Systems (NeurIPS)}, vol.~35, pp. 17\,515--17\,527, 2022.

\bibitem{de2020batch}
S.~De and S.~Smith, ``Batch normalization biases residual blocks towards the
  identity function in deep networks,'' \emph{Advances in Neural Information
  Processing Systems (NeurIPS)}, vol.~33, pp. 19\,964--19\,975, 2020.

\bibitem{BMVC2016_87}
S.~Zagoruyko and N.~Komodakis, ``Wide residual networks,'' in \emph{Proceedings
  of the British Machine Vision Conference (BMVC)}.\hskip 1em plus 0.5em minus
  0.4em\relax BMVA Press, September 2016, pp. 87.1--87.12.

\bibitem{tay2021long}
Y.~Tay, M.~Dehghani, S.~Abnar, Y.~Shen, D.~Bahri, P.~Pham, J.~Rao, L.~Yang,
  S.~Ruder, and D.~Metzler, ``Long range arena: A benchmark for efficient
  transformers,'' in \emph{International Conference on Learning Representations
  (ICLR)}, 2021.

\bibitem{nangia2018listops}
N.~Nangia and S.~R. Bowman, ``Listops: A diagnostic dataset for latent tree
  learning,'' \emph{arXiv preprint arXiv:1804.06028}, 2018.

\bibitem{maas2011learning}
A.~Maas, R.~E. Daly, P.~T. Pham, D.~Huang, A.~Y. Ng, and C.~Potts, ``Learning
  word vectors for sentiment analysis,'' in \emph{Proceedings of the 49th
  annual meeting of the association for computational linguistics: Human
  language technologies}, 2011, pp. 142--150.

\bibitem{radev2013acl}
D.~R. Radev, P.~Muthukrishnan, V.~Qazvinian, and A.~Abu-Jbara, \emph{Language
  Resources and Evaluation}, vol.~47, pp. 919--944, 2013.

\bibitem{linsley2018learning}
D.~Linsley, J.~Kim, V.~Veerabadran, C.~Windolf, and T.~Serre, ``Learning
  long-range spatial dependencies with horizontal gated recurrent units,''
  \emph{Advances in neural information processing systems}, vol.~31, 2018.

\bibitem{fu2022h3}
D.~Y. Fu, T.~Dao, K.~K. Saab, A.~W. Thomas, A.~Rudra, and C.~Re, ``Hungry
  hungry hippos: Towards language modeling with state space models,'' in
  \emph{The Eleventh International Conference on Learning Representations
  (ICLR)}, 2022.

\bibitem{xiong2021nystromformer}
Y.~Xiong, Z.~Zeng, R.~Chakraborty, M.~Tan, G.~Fung, Y.~Li, and V.~Singh,
  ``Nystr{\"o}mformer: A nystr{\"o}m-based algorithm for approximating
  self-attention,'' in \emph{Proceedings of the AAAI Conference on Artificial
  Intelligence}, vol.~35, no.~16, 2021, pp. 14\,138--14\,148.

\bibitem{vaswani2017attention}
A.~Vaswani, N.~Shazeer, N.~Parmar, J.~Uszkoreit, L.~Jones, A.~N. Gomez,
  {\L}.~Kaiser, and I.~Polosukhin, ``Attention is all you need,''
  \emph{Advances in Neural Information Processing Systems (NeurIPS)}, vol.~30,
  2017.

\bibitem{wang2019glue}
A.~Wang, A.~Singh, J.~Michael, F.~Hill, O.~Levy, and S.~R. Bowman, ``{GLUE}: A
  multi-task benchmark and analysis platform for natural language
  understanding,'' in \emph{International Conference on Learning
  Representations (ICLR)}, 2019.

\bibitem{sang2003introduction}
E.~F. Sang and F.~De~Meulder, ``Introduction to the {CoNLL}-2003 shared task:
  Language-independent named entity recognition,'' \emph{arXiv preprint
  cs/0306050}, 2003.

\bibitem{hassid2022does}
M.~Hassid, H.~Peng, D.~Rotem, J.~Kasai, I.~Montero, N.~A. Smith, and
  R.~Schwartz, ``How much does attention actually attend? {Q}uestioning the
  importance of attention in pretrained transformers,'' in \emph{Findings of
  the Association for Computational Linguistics: EMNLP 2022}, 2022, pp.
  1403--1416.

\bibitem{gu2022efficiently}
A.~Gu, K.~Goel, and C.~Re, ``Efficiently modeling long sequences with
  structured state spaces,'' in \emph{International Conference on Learning
  Representations (ICLR)}, 2022.

\bibitem{bandeira2020}
\BIBentryALTinterwordspacing
A.~S. Bandeira, \emph{Mathematics of Data Science}, 2020, book draft version
  0.1. [Online]. Available:
  \url{https://people.math.ethz.ch/~abandeira/teaching.html}
\BIBentrySTDinterwordspacing

\bibitem{hua2022transformer}
W.~Hua, Z.~Dai, H.~Liu, and Q.~Le, ``Transformer quality in linear time,'' in
  \emph{International Conference on Machine Learning (ICML)}.\hskip 1em plus
  0.5em minus 0.4em\relax PMLR, 2022, pp. 9099--9117.

\end{thebibliography}

\end{document}


\maketitle

\appendix 
\setcounter{figure}{0}
\renewcommand{\thefigure}{A\arabic{figure}}
\setcounter{table}{0}
\renewcommand{\thetable}{A\arabic{table}}

\vspace{-3cm}
\renewcommand\contentsname{}
\tableofcontents

\section{\MIMONets Details}

\subsection{VSA representations and operations in \MIMONets}
There are numerous available options for binding and unbinding depending on the VSA models being used~\cite{kleyko2023survey}. Table~\ref{tab:representation} summarizes the VSA representations and operations used in \MIMOConv and \MIMOFormer.
\MIMOConv relies on holographic reduced representation (HRR)~\cite{PlateHolographic1995} for binding, and matrix binding of additive terms (MBAT)~\cite{gallant2013mbat} for unbinding. 
The binding and unbinding keys are real-valued and trainable. 
At initialization, each element in the $D$-dimensional key vector is drawn from an i.i.d. Gaussian distribution with zero mean and $1/D$ variance.
Optionally, the binding keys can be frozen during training while maintaining a high accuracy (see Appendix~\ref{sec:ablation_mimoconv}).
\MIMOConv's binding relies on our proposed position-wise HRR (\PWHRR) binding, which maintains the image's local structure. 
The unbinding is implemented with MBAT, which computes the vector-matrix multiplication between the CNN's $D_o$-dimensional output feature vector and an unbinding matrix $\tilde{A}^{(i)}\in \mathbb{R}^{D_o \times D_o}$. 
The output dimension is $D_o$=$640$ in WideResNet-28-10. 
The MBAT unbinding provides a higher degree of freedom by having $D_o^2$ trainable parameters, whereas HRR's unbinding would generate a circulant matrix with $D_o$ trainable parameters. 
However, it requires only 409,600 parameters per superposition channel, which is negligible compared to the remaining layers in \MIMOConv, which have 36.54\,M trainable parameters. Due to deep neural networks being highly nonlinear, we thus opt for this more flexible variant of unbinding by arbitrary linear transformations.

\MIMOFormer uses the multiply-add-permute (MAP)~\cite{MAP_1998} model, which uses bipolar keys and the element wise multiplication (Hadamard product) for binding and unbinding. 
The bipolar keys are drawn from a Rademacher distribution and are frozen during training and inference. 

\begin{table}[h]
\centering
\caption{Summary of VSA representations and operations used in \MIMOConv and \MIMOFormer.}
\label{tab:representation}
\resizebox{\linewidth}{!}{
\begin{tabular}{lcccccc}
\toprule
           &  &  & \multicolumn{2}{c}{Binding}         & \multicolumn{2}{c}{Unbinding}    \\
\cmidrule(r){4-5}\cmidrule(r){6-7}
           & VSA framework              & Key representation                   & Operation        & Keys             & Operation            & Keys      \\
\cmidrule(r){1-1}\cmidrule(r){2-2}\cmidrule(r){3-3}\cmidrule(r){4-4}\cmidrule(r){5-5}\cmidrule(r){6-6}\cmidrule(r){7-7}
MIMOConv   & HRR/MBAT           & real-valued        & PWHRR            & trainable/frozen & MBAT & trainable \\
MIMOFormer & MAP           & bipolar            & Hadamard product & frozen           & Hadamard product     & frozen   \\
\bottomrule
\end{tabular}
}
\end{table}

\subsection{Illustration of the Blessing of Dimensionality}\label{subsec:blessing}
VSA builds upon the mathematical concept of the Blessing of Dimensionality. According to it, random vectors are (quasi-)orthogonal with high probability. Let us illustrate one version of it.
Suppose independent random bipolar vectors $x,y\in \lbrace -1, +1 \rbrace^D$ with i.i.d. Rademacher distributed components as used in \MIMOFormer. It holds by Hoeffding's inequality (Appendix~\ref{app:theorems})
\begin{equation}
    \mathbb{P}\left(\abs{\cos \measuredangle (x,y)} \geq \alpha \right) = \mathbb{P}\left(\abs{\langle x , y \rangle }\geq \alpha D \right) \leq 2 e^{-D\alpha^2/2} \qquad \forall \alpha \geq 0
\end{equation}
Similar bounds exist for Gaussian random vectors, which are used to generate keys for MIMOConv. Let us set some cutoff for interference events (IE), namely, we consider two vectors to interfere with each other if the angle between them is less than $70^\circ$  (corresponding to $\alpha = \cos(70^\circ)$), already $20^\circ$ off from exact orthogonality.
The bound demonstrates that the probability for two vectors (with i.i.d. Rademacher distributed components) to interfere (IE) is less than 0.785 for vectors in 16 dimensions. As such, high levels of interference could still occur frequently.
In contrast, for 64 dimensions, the probability that two vectors interfere (IE) is already known to be less than 0.0474. Consequently, we are much more certain that interference occurs with low probability.

\subsection{Noisy retrieval of values from a key-value superposition}\label{sec:comparison_dictionary}

Consider a superposition of $N$ bound values $x^{(i)}$
 \begin{align}
     s = a^{(1)} \odot x^{(1)}  +  a^{(2)}  \odot x^{(2)} + \ldots + a^{(N)}  \odot x^{(N)}
 \end{align}
 where the binding keys $a^{(i)}$ are (for instance) independent bipolar vectors of i.i.d. Rademacher entries. Unbinding with \(a^{(k)}\) produces the signal of interest \(x^{(k)}\) together with a noise vector orthogonal to it:
 \begin{align}
     a^{(k)} \oast s &= a^{(k)} \oast a^{(1)} \odot x^{(1)}  +  a^{(k)} \oast a^{(2)}  \odot x^{(2)} + \ldots + a^{(k)} \oast a^{(N)}  \odot x^{(N)} \\
     &= x^{(k)} + noise.
 \end{align}
    The noise vector stems from the approximate unbinding ($ a^{(k)} \oast  a^{(k)} \odot x^{(k)} \approx x^{(k)}$) as well as randomized value vectors ($a^{(k)} \oast a^{(j)} \odot x^{(j)}$). Importantly, $noise$ becomes orthogonal to $x^{(1)}$, hence distinguishable, with a growing embedding dimension according to the Blessing of Dimensionality. The effect of noise is mitigated after comparing against a dictionary of known outputs, based on a notion of inner product. Such a comparison naturally, but not exclusively, arises in classification tasks. Concretely, comparing against $a^{(k)} \odot \Omega$ for an $\Omega$ aligned with $x^{(k)}$ returns
 \begin{align}
     \langle a^{(k)} \oast s, \Omega \rangle &= \langle s, a^{(k)} \odot \Omega \rangle \\
     &= \sum\limits_{i = 1}^{N} \langle a^{(i)} \odot x^{(i)}, a^{(k)} \odot \Omega \rangle \\
     &\approx \langle a^{(k)} \odot x^{(k)}, a^{(k)} \odot \Omega \rangle \\
     &= \langle x^{(k)}, \Omega \rangle
 \end{align}
where we still assume bipolar binding and where the approximation relies on the Blessing of Dimensionality producing orthogonal vectors. For the VSA framework MAP, which uses bipolar keys of i.i.d. Rademacher entries, we provide a more precise formulation:

\begin{thm}[Dictionary Cleanup Noise]
    Let $\Omega \in \mathbb{R}^D$ be an element of a dictionary and consider the superposition of bound values $x^{(i)} \in \mathbb{R}^D$
    \begin{equation}
        s = a^{(1)} \odot x^{(1)}  +  a^{(2)}  \odot x^{(2)} + \ldots + a^{(N)}  \odot x^{(N)}
    \end{equation}
    where the binding keys $a^{(i)} \in \{-1, 1\}^D$ are independent bipolar vectors of i.i.d. Rademacher entries. It then holds
   
    \begin{align}
        \mathbb{P}\Bigg\{\left\langle s, a^{(k)} \odot \Omega \right\rangle \not\in [1-\alpha, 1+\alpha] \cdot
        \left\langle x^{(k)}, \Omega \right \rangle \Bigg\}
        \leq 2 \exp\left( - \frac{\alpha^2 \abs{\langle x^{(k)}, \Omega \rangle}^2}{2 \sum_{i \not= k} \norm{x^{(i)} \odot \Omega}_2^2}\right)
    \end{align}
    with the exponent, given
    $x^{(i)}$ are all of roughly equal norm, according to Theorem \ref{thm:norm_hadamard_product} in Appendix~\ref{app:theorems} typically scaling as
\begin{equation}
    -D \alpha^2 \cos^2(\measuredangle (x^{(k)}, \Omega)) \bigg/ 2(N-1)
\end{equation}
Thus, for $D \gg N$ comparison against elements from a dictionary allows faithful retrieval.
\end{thm}

\begin{proof}
    By the following equivalence relation
    \begin{equation}
        \left\langle s, a^{(k)} \odot \Omega \right\rangle \not\in [1-\alpha, 1+\alpha] \cdot
        \left\langle x^{(k)}, \Omega \right \rangle
        \iff \abs{\sum\limits_{i \not=k}\left\langle a^{(i)} \odot x^{(i)}, a^{(k)} \odot \Omega \right\rangle} > \alpha \abs{
        \left\langle x^{(k)}, \Omega \right\rangle}
    \end{equation}
it suffices to derive tail bounds on 
\begin{multline}
    \mathbb{P}\left\{ 
    \abs{\sum\limits_{i \not=k}\left\langle a^{(i)} \odot x^{(i)}, a^{(k)} \odot \Omega \right\rangle} > \alpha \abs{\left\langle x^{(k)}, \Omega
        \right\rangle} \right\}\\ = \mathbb{P}\left\{ 
    \abs{\sum\limits_{i \not=k} \sum\limits_{d} a^{(i)}_d a^{(k)}_d x^{(i)}_d \Omega_d} > \alpha \abs{\left\langle x^{(k)}, \Omega
        \right\rangle} \right\}
\end{multline}
Since $\{ a^{(i)}_d \cdot a^{(k)}_d \}_{d,i\not=k}$ for $k$ fixed is a set of i.i.d. Rademacher random variables, we are in a position to apply Hoeffding's inequality (see Appendix, Theorem \ref{thm:hoeffding's inequality}) which gives
\begin{align}
    \mathbb{P}\left\{ 
    \abs{\sum\limits_{i \not=k} \sum\limits_{d} a^{(i)}_d a^{(k)}_d x^{(i)}_d \Omega_d} > \alpha \abs{\left\langle x^{(k)}, \Omega
        \right\rangle} \right\} & \leq 2 \exp\left( - \frac{\alpha^2 \abs{\langle x^{(k)}, \Omega \rangle}^2}{2 \sum_{i\not=k} \sum_d \abs{x_d^{(i)} \Omega_d}^2 } \right)
\end{align}
\end{proof}

\begin{figure}[t]
    \centering
    \includegraphics[width=\textwidth]{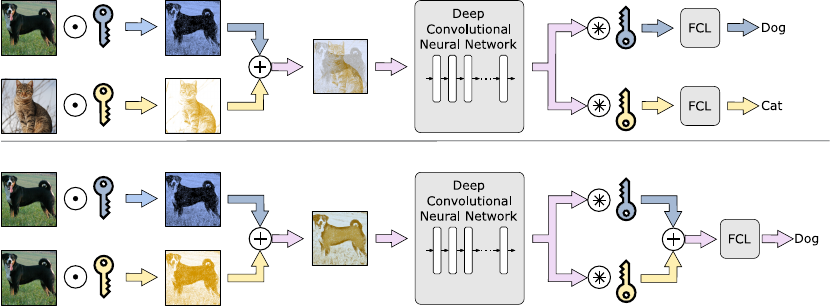}
    \caption{Depiction of a single trained \MIMOConv performing dynamic inference. Instead of the fast $N$=$2$ mode (above) the same input can be inserted twice for the slow $N$=$1$ mode (below) effectively implementing an ensemble method. We can instantaneously switch between the modes.}
    \label{fig:illustration_dynamic_inference}
\end{figure}

\subsection{Illustration of dynamic inference}
To explore the idea of dynamic inference, suppose only two superposition channels are used with binding keys $a^{(1)}, a^{(2)}$ and unbinding keys $\tilde{a}^{(1)}, \tilde{a}^{(2)}$. We already know how the model performs standard computation in superposition (see Eq.~(1) -- (4) in the main text). Let us thus examine how a network with the same parameters can instead be used as an ensemble-method with higher accuracy, but lower throughput. A superposition is established of twice the same input $x$:

\begin{align}
    s = a^{(1)} \odot x  +  a^{(2)}  \odot x
\end{align}

After applying the deep neural network $f_\theta$ to the superposition, we may unbind as

\begin{align}
    \tilde{a}^{(1)} \oast f_\theta(s) &\approx \tilde{a}^{(1)} \oast f_\theta\left(a^{(1)} \odot x\right) +  \tilde{a}^{(1)} \oast f_\theta\left(a^{(2)} \odot x\right) \\
    &\approx f_\theta\left(x\right) + \tilde{a}^{(1)} \oast f_\theta\left(a^{(2)} \odot x\right)
\end{align}
and

\begin{align}
    \tilde{a}^{(2)} \oast f_\theta(s) &\approx \tilde{a}^{(2)} \oast f_\theta\left(a^{(1)} \odot x\right) +  \tilde{a}^{(2)} \oast f_\theta\left(a^{(2)} \odot x\right) \\
    &\approx \tilde{a}^{(2)} \oast f_\theta\left(a^{(1)} \odot x\right) + f_\theta\left(x\right).
\end{align} 

After averaging the two expressions, we get
\begin{align}
 \frac{1}{2}\left( \tilde{a}^{(1)} \oast f_\theta(s)  + \tilde{a}^{(2)} \oast f_\theta(s) \right) \approx f_\theta\left(x\right) + noise   
\end{align}

where $noise$ is a random noise vector and $f_\theta\left(x\right)$ is approximated as an average of two predictions. Owing to the introduction of stochasticity by the binding and unbinding process these predictions are decorrelated, i.e., each superposition channel is processed to some degree differently. See Figure~\ref{fig:illustration_dynamic_inference} for an illustration.

\subsection{Alternative throughput-increasing methods}
Although not a focus of this work, we believe that computation in superposition can be combined with other throughput-increasing methods such as model-downsizing, quantization aware training, and pruning. 

The Blessing of Dimensionality gives, in terms of dimensionality, an exponentially decreasing probability of interference for superpositions, even for (2-bit quantized) Rademachers. The extent to which these superpositions can be kept intact as linear layers act on them depends on the conditioning of the matrix (ideally nearly-isometric) not on the fidelity of its entries. As such we suspect that MIMOConv can be mixed with quantization, weight pruning, etc.

Regarding MIMOFormer, we can give quantitative insights. As is evident from Theorem~\ref{thm:attention_cross_talk}, the error bounds have no dependence on the precision of projection weights, but depend only on the embedding dimensionality, the size of keys and queries, and the angles between them. Consequently, quantization, pruning, etc. are not in competition with our approach and can be easily combined.

Naturally, when combining different methods not only the gains but also the errors add up. However, with diminishing returns of each method we believe the combination of several to be most effective, especially given that our method is not competing with alternatives for the same resources of a model and allows it to conduct dynamic inference.

\newpage

\section{\MIMOConv Details}

\subsection{Inner-product preserving activation functions}
Any inner-product preserving map is linear (see Appendix~\ref{app:theorems} Theorem \ref{thm:inner_product_preserving_map_is_linear}).
With activation functions being introduced to break the linearity of neural networks, they are innately at odds with inner-product preservation. According to~\cite{Hu2020Provable}, a trade-off can be reached between preserving inner products and introducing nonlinearities by replacing the ReLU activation function with shifted ReLU (sReLU)
\begin{equation}
    sReLU_b(x) = ReLU(x-b)+b = \max (x, b),
\end{equation}
where the trainable bias $b$ determines the trade-off and is initialized to $-1$.
However, in our experiments (see Appendix~\ref{results:isometry_at_activation_function}), replacing sReLU with parametric ReLU (pReLU)~\cite{he2015prelu}, another activation function capable of choosing the extent of nonlinearity, defined as
\begin{equation}
    pReLU_b(x) = ReLU(x) - b \cdot ReLU(-x) = max(x,0) + b\cdot\  min(x,0)
\end{equation}
gives higher performance. 
The trainable parameter $b \in [-1, 1]$ controls the degree of linearity, where $b$=$1$ indicates fully linear behavior. It is initialized to $b$=$0.5$ at the beginning of training. 

\subsection{Details on isometric convolutional layers}\label{app:isometry_conv_explanation}

As elaborated in the main text, we strive for inner-product preserving maps. With inner-product preserving maps being norm-preserving and by extension distance-preserving if linear, and with linear distance-preserving maps being norm preserving and according to the polarization identity also inner-product preserving, it holds that inner-product preserving maps are equivalent to linear isometries. Hence the name of the regularization term being \textit{isometry regularization term}.
%
%
The adopted regularization takes the form
\begin{align}
    L(W) & = \tfrac{\gamma}{2} \norm{Conv(W,W) - \delta_{C_o }}_F^2 \, ,  &\delta_{C_o }[:, :, j, l]\!  &=\! I_{C_o \times C_o}\! \cdot\! \mathds{1}_{j, l = \lfloor \tfrac{k}{2} \rfloor} \label{eq:isometry}\\
    L(W^T) & = \tfrac{\gamma}{2} \norm{Conv(W^T,W^T) - \delta_{C_i }}_F^2 \, , &\delta_{C_i }[:, :, j, l]\!  &=\! I_{C_i \times C_i}\! \cdot\! \mathds{1}_{j, l = \lfloor \tfrac{k}{2} \rfloor} \label{eq:isometry_transposed}
\end{align}
where $W \in \mathbb{R}^{C_o \times C_i \times k \times k}$ contains the weights of a convolutional layer. $C_o$ denotes the number of output feature maps, $C_i$ the number of input feature maps, and $k$ the (square) kernel size. $W^T$ refers to a kernel with the first two dimensions of $W$ transposed.
In the notation of Einstein summations, the 2D convolution $Conv(U, V)$ evaluates to
\begin{equation}
    O_{a,b,c,d} = U_{a,r,c+s,d+t} \cdot V_{b,r,s,t}
\end{equation}
This is implemented in Pytorch by the usual zero-padded spatial 2D convolution taking an input in the first argument (with ranks: batch size, fmaps, height, width) and a convolutional kernel in the second argument (with ranks: output fmaps, input fmaps, kernel height, kernel width).
 
Unless the number of input fmaps and output fmaps coincide, only one of $W$ and its adjoint $W^T$ may be isometries. Thus we use $L(W)$ when $C_i > C_o$ and $L(W^T)$ otherwise. 

For more information on why such a regularization term may help to preserve inner products, see~\cite{qi2020deep} where it was first proposed.

\subsection{Binding key regularization}\label{sec:binding_key_reg_term}
 We use a regularization term to keep the binding vectors ($a^{(i)}$) orthonormal:
\begin{equation}
    L({a^{(1)}, \ldots, a^{(N)}}) = \tfrac{\mu}{\binom{N}{2}}\sum_{i=1}^N \sum_{j=i+1}^N  (\tfrac{\langle a^{(i)}, a^{(j)} \rangle}{\norm{a^{(i)}} \norm{a^{(j)}}})^2 + \tfrac{\mu}{N}\sum\limits_{i=1}^N (\norm{a^{(i)}} -1)^2, 
\end{equation}
with hyperparameter $\mu$. 
A grid search on the validation set found a value of $\mu$=$0.1$ to give the best results. Alternatively, the binding keys may be frozen after random (Gaussian) initialization, guaranteeing orthogonality in the limit of high key dimension (see Appendix~\ref{subsec:blessing}).

\newpage

\section{\MIMOFormer Details}

\subsection{FAVOR+ in the Performer}\label{sec:favor+_description}
Here, we revisit the Performer's FAVOR+ attention block~\cite{choromanski2020performer} and in the next subsection we validate the use of the ReLU activation in the projection. 
FAVOR+ takes advantage of the fact that $a,b \mapsto \exp(a^T b / \sqrt{D})$ is a kernel and can be represented as an explicit inner product (inverse kernel trick) in an infinite-dimensional space of transformed inputs. The mapping to this infinite-dimensional space is approximated with a randomized feature map $\phi:\mathbb{R}^D \to \mathbb{R}_+^R$ of finitely many entries. More explicitly, since
\begin{align}
    & \mathbb{E}_{w \sim \mathcal{N}(0, I_D)}\left[\exp(\tfrac{w_i^T q}{\sqrt[4]{D}}-\tfrac{\norm{q}_2^2}{2\sqrt{D}}) \cdot \exp(\tfrac{w_i^T k}{\sqrt[4]{D}}-\tfrac{\norm{k}_2^2}{2\sqrt{D}})\right]\\
    = & \exp(\tfrac{-\norm{q}_2^2-\norm{k}_2^2}{2\sqrt{D}}) \cdot \mathbb{E}_{w \sim \mathcal{N}(0, I_D)}[\exp(w_i^T \tfrac{q+k}{\sqrt[4]{D}})]\\
    = & \exp(\tfrac{-\norm{q}_2^2-\norm{k}_2^2}{2\sqrt{D}}) \cdot \exp(\tfrac{\norm{q+k}_2^2}{2\sqrt{D}}) = \exp(q^T k / \sqrt{D}),
\end{align}
drawing $w_1, \ldots, w_R \sim \mathcal{N}(0, I_D)$ i.i.d. induces a function $\phi:\mathbb{R}^D \to \mathbb{R}_+^R$ with components given by $\phi_i(x) = \exp(\tfrac{w_i^T x}{\sqrt[4]{D}}-\tfrac{\norm{x}_2^2}{2\sqrt{D}}) / \sqrt{R}$ that, by the law of large number, approximates $\exp(q^T k / \sqrt{D})$, i.e.
\begin{equation}
    \langle \phi(k),\ \phi(q) \rangle \approx \exp(q^T k / \sqrt{D}) .
\end{equation}
Alternatively, by partitioning $w_1, \ldots, w_R$ into subsets of cardinality $D$ and drawing each such subset from the orthogonal group before rescaling each $w_i$ according to the $\chi_D$-distribution it still holds $w_i \sim \mathcal{N}(0, I_D)$, but the entries are no longer independent. Using such orthogonal features, one obtains an unbiased estimate of $\exp(q^T k / \sqrt{D})$ with lower variance than independently drawing $w_1, \ldots, w_R \sim \mathcal{N}(0, I_D)$, see \cite{choromanski2020performer}. 

The inverse kernel trick then allows FAVOR+ to take advantage of the associativity of matrix multiplication to give the following factored expression of dot-product attention:
\begin{align}
    o_i & = \sum\limits_{j=1}^L v_j \tfrac{\langle \phi(k_j), \phi(q_i) \rangle}{\sum\limits_{j=1}^L \langle \phi(k_j), \phi(q_i) \rangle} = \frac{\sum\limits_{j=1}^L v_j \left(\phi(k_j)^T \phi(q_i)\right)}{\underbrace{\sum\limits_{j=1}^L  \left(\phi(k_j)^T \phi(q_i)\right)}_{B_i}} 
    = \frac{\overbrace{\sum\limits_{j=1}^L v_j\  \phi(k_j)^T}^{A} \cross \phi(q_i)}{\underbrace{\sum\limits_{j=1}^L \phi(k_j)^T}_{C} \cross \phi(q_i)},
\end{align}
where $A$ and $C$ must only be computed once. With the computational complexity of evaluating $\phi$ being $\mathcal{O}(DR)$, this takes in both cases $\mathcal{O}(L D R)$. Computing $\phi(q_i)$ for all $i$ requires another $\mathcal{O}(L D R)$. Finally, the matrix-vector product $\cross$  for a given query position $i$ takes $\mathcal{O}(D R)$ for the numerator and $\mathcal{O}(R)$ for the denominator. Thus, computing outputs $o_i$ for all $i$ takes only $\mathcal{O}(LDR)$, which can be a considerable improvement over the usual $\mathcal{O}(L^2 D)$ for long sequence lengths ($L$).

\subsection{ReLU activation in \FAVORpS}

To increase training stability, we use a ReLU-based projection $\phi_i(x) = ReLU(w_i^T x) / \sqrt{R\sqrt{D}}$ instead of the above-mentioned unbiased approximation of softmax through $\phi_i(x) = \exp(\tfrac{w_i^T x}{\sqrt[4]{D}}-\tfrac{\norm{x}_2^2}{2\sqrt{D}}) / \sqrt{R}$. Such an approach was already mentioned in the Performer \cite{choromanski2020performer}. Here, we derive (new to our knowledge) a theoretical justification. As the following theorem expresses, the ReLU approximation leads to a {(bi-)linear} dependency on the norm of keys and queries while retaining a roughly polynomial dependency (with exponent $\log_2(\pi) \approx 1.65$) on the query-key alignment $\rho = \langle x, y \rangle /(\norm{x} \norm{y})$. 

\begin{samepage}
\begin{thm}[ReLU approximation]
    Let $w \sim \mathcal{N}(0, I_D)$. Then 
    \begin{align}
         \mathbb{E}[ReLU(w^T x) \cdot ReLU(w^T y)] = \frac{\langle x, y\rangle + \norm{x} \norm{y} g(\tfrac{\langle x, y \rangle}{\norm{x} \norm{y}})}{4} \approx &\norm{x}\norm{y} \tfrac{(\rho+1)^{\log_2(\pi)}}{2\pi}\label{eq:first_interpretation_approx_relu}
     \end{align}
     for
    \begin{equation}
         g(\rho) = \tfrac{2}{\pi} \left[\sqrt{1-\rho^2} + \abs{\rho} \arctan(\tfrac{\abs{\rho}}{\sqrt{1-\rho^2}}) \right] \text{ with } \rho = \langle x, y \rangle / (\norm{x} \norm{y})
     \end{equation}
\end{thm}
\end{samepage}

\begin{figure}
    \centering
    \includegraphics[width=0.5\linewidth]{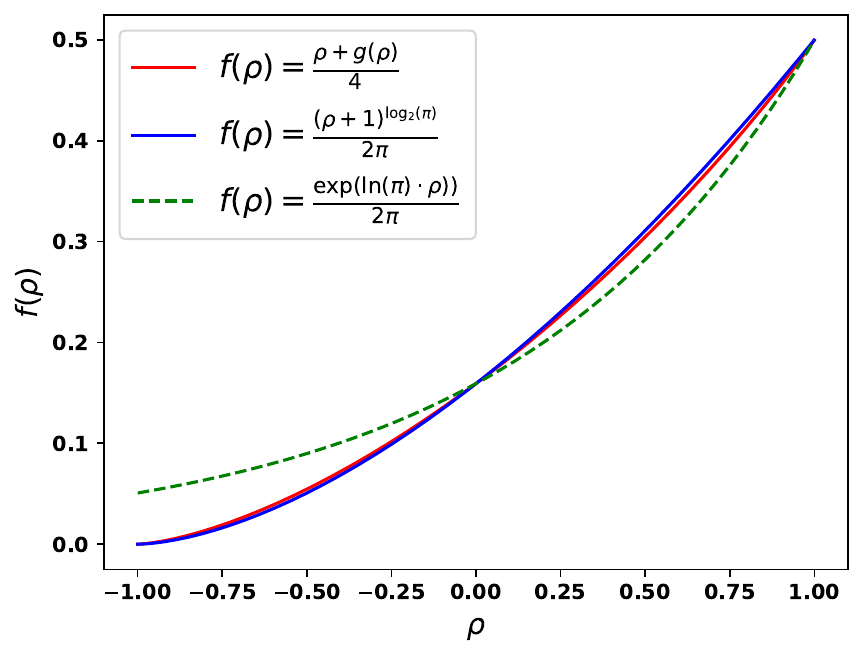}
      \caption{Approximation of $\mathbb{E}[ReLU(w^Tx) \cdot ReLU(w^T y)] / (\norm{x} \norm{y})$ (red line) with a polynomial (blue line). Softmax function is shown with green dashed line. }
      \label{fig:relu}
\end{figure}

Figure~\ref{fig:relu} illustrates the validity of interpreting $\mathbb{E}[ReLU(w^Tx) \cdot ReLU(w^T y)] / (\norm{x} \norm{y})$ as a polynomial. The general behaviour is also similar to the usual softmax function (green dashed line), although the norm of $x,y$ no longer affecting the exponent gives greater stability.
\begin{proof}
    \begin{align}
        & \mathbb{E}[ReLU(w^T x) \cdot ReLU(w^T y)] = \tfrac{1}{4} \mathbb{E}[(w^T x + \abs{w^T x})(w^T y + \abs{w^T y})]\\
        = & \frac{\mathbb{E}[ w^T x \cdot w^T y] + \cancel{\mathbb{E}[ w^T x \cdot \abs{w^T y}]} + \cancel{\mathbb{E}[ \abs{w^T x} \cdot w^T y]} + \mathbb{E}[ \abs{w^T x \cdot w^T y}]}{4}\\
        = & \frac{\mathbb{E}[ x^T w \cdot w^T y] + \mathbb{E}[ \abs{w^T x \cdot w^T y}]}{4}
        = \frac{x^T\mathbb{E}[ w w^T ]y + \mathbb{E}[ \abs{w^T x \cdot w^T y}]}{4}\\
        = & \frac{ \langle x, y \rangle + \mathbb{E}[ \abs{w^T x \cdot w^T y}]}{4}
    \end{align}
    Let us now proceed to evaluate $\mathbb{E}[ \abs{w^T x \cdot w^T y}]$. First notice that $X = w^T x, Y = w^T y$ are jointly Gaussian with
    \begin{equation}
        \begin{bmatrix} X\\Y \end{bmatrix} \sim \mathcal{N}\left(0, \begin{bmatrix} \sigma_X^2 & \rho\ \sigma_X \sigma_Y\\ \rho\ \sigma_X \sigma_Y & \sigma_Y^2\end{bmatrix} \right) = \mathcal{N}\left(0, \begin{bmatrix} \norm{x}^2 & \langle x, y \rangle\\ \langle x, y \rangle & \norm{y}^2\end{bmatrix} \right) .
    \end{equation}
    where $\rho = \tfrac{Cov(X,Y)}{\sigma_X\ \sigma_Y} = \langle x, y \rangle / (\norm{x} \norm{y})$ measures the alignment between $x$ and $y$. Hence, by the innovations form for centered jointly Gaussians it holds
    \begin{equation}
        (X | Y = y) \sim \mathcal{N}(\mu_{x|y}(y), \sigma^2_{x|y}(y)) = \mathcal{N}(\rho\ \sigma_X \sigma_Y^{-1} y, (1-\rho^2) \sigma_X^2).
    \end{equation}
    Thus, the expectation of the folded normal distribution is given by
    \begin{align}
        \mathbb{E}[\abs{X} | Y = y] = & \sigma_{x|y}(y) \sqrt{\tfrac{2}{\pi}} \exp(-\tfrac{\mu_{x|y}^2(y)}{2\sigma^2_{x|y}(y)}) + \mu_{x|y}(y) (1- 2 \Phi(- \tfrac{\mu_{x|y}(y)}{\sigma_{x|y}(y)}))\\
        = & \sqrt{1-\rho^2} \sigma_X \sqrt{\tfrac{2}{\pi}} \exp(-\tfrac{\rho^2 \sigma_X^2 \sigma_Y^{-2}y^2}{2(1-\rho^2) \sigma_X^2}) + \rho \sigma_X \sigma_Y^{-1}y (1- 2 \Phi(- \tfrac{\rho \sigma_X \sigma_Y^{-1}y}{\sqrt{1-\rho^2} \sigma_X}))\\
        = & \sqrt{1-\rho^2} \sigma_X \sqrt{\tfrac{2}{\pi}}  \ \cdot \exp(-\tfrac{\rho^2 y^2}{2(1-\rho^2) \sigma_Y^2}) + \abs{\rho} \sigma_X\ \cdot\ \sigma_Y^{-1}y (2 \Phi(\tfrac{\abs{\rho}\ y}{\sqrt{1-\rho^2} \sigma_Y}) - 1)
    \end{align}
    where $\Phi(s) = \tfrac{1}{\sqrt{2\pi}} \int_{-\infty}^s \exp(-t^2/2) dt$ denotes the cumulative distribution function of the standard normal distribution. Thus, we split the target expression in two:
    \begin{align}
        \mathbb{E}[ \abs{w^T x \cdot w^T y}] = &\mathbb{E}[\abs{X} \abs{Y}] = \mathbb{E}[\mathbb{E}[\abs{X} \cdot \abs{Y}\ \big|\ Y]] = \mathbb{E}[\abs{Y} \cdot\mathbb{E}[\abs{X}\ \big|\ Y] ]\\
        = & \sqrt{1-\rho^2} \sigma_X \sqrt{\tfrac{2}{\pi}}  \ \cdot \mathbb{E}\left[ \abs{Y} \exp(-\tfrac{\rho^2 Y^2}{2(1-\rho^2) \sigma_Y^2})\right]\\
        + & \abs{\rho} \sigma_X\ \cdot\ \mathbb{E}\left[ \abs{Y} \sigma_Y^{-1} Y (2 \Phi(\tfrac{\abs{\rho}\ Y}{\sqrt{1-\rho^2} \sigma_Y}) - 1) \right].
    \end{align}
    For the first term, we get
    \begin{align}
        \mathbb{E}\left[ \abs{Y} \exp(-\tfrac{\rho^2 Y^2}{2(1-\rho^2) \sigma_Y^2})\right] = & \int |y| \exp(-\tfrac{\rho^2 y^2}{2(1-\rho^2) \sigma_Y^2}) \tfrac{1}{\sqrt{2\pi}\sigma_Y} \exp{-\tfrac{y^2}{2\sigma_Y^2}} dy\\
        = & \sqrt{1-\rho^2}\int |y| \tfrac{1}{\sqrt{2\pi} \sqrt{1-\rho^2} \sigma_Y} \exp(-\tfrac{y^2}{2(1-\rho^2) \sigma_Y^2}) dy\\
        = & \sqrt{1-\rho^2}\ \mathbb{E}[|Z|] \ \text{ where } Z \sim \mathcal{N}(0, (1-\rho^2)\sigma_Y^2)\\
        = & \sqrt{1-\rho^2} \sqrt{1-\rho^2} \sigma_Y \sqrt{\tfrac{2}{\pi}} = (1-\rho^2) \sigma_Y \sqrt{\tfrac{2}{\pi}},
    \end{align}
    using again the formula for the expectation of a folded normal distribution.
    The second term is a bit more involved to evaluate. We substitute $S = \tfrac{Y}{\sigma_Y}$, i.e. $S \sim \mathcal{N}(0, 1)$:
    \begin{align}
    \mathbb{E}[\abs{Y} \tfrac{Y}{\sigma_Y}(2 \Phi(\tfrac{\abs{\rho}\ Y}{\sqrt{1-\rho^2} \sigma_Y}) - 1)] = & \sigma_Y \mathbb{E}[\abs{S} S (2 \Phi(\tfrac{\abs{\rho}\ S}{\sqrt{1-\rho^2}}) - 1)]\\
     = & \sigma_Y \mathbb{E}[S^2 (2 \Phi(\tfrac{\abs{\rho}\ \abs{S}}{\sqrt{1-\rho^2}}) - 1)]\\
     = & \sigma_Y \int p_s(s) s^2 \cdot (2 \Phi(\tfrac{\abs{\rho}\ \abs{s}}{\sqrt{1-\rho^2}}) - 1) ds\\
     = & 2\sigma_Y \int p_s(s) s^2 \Phi(\tfrac{\abs{\rho}\ \abs{s}}{\sqrt{1-\rho^2}}) ds - \sigma_Y\\
     = & 4\sigma_Y \int_0^\infty s^2 p_s(s) \Phi(b s) ds - \sigma_Y\\
     = & 4 \sigma_Y ( \tfrac{1}{4} + \tfrac{1}{2\pi} (\tfrac{b}{1+b^2} + \arctan (b))) - \sigma_Y\\
     = & 4 \sigma_Y ( \tfrac{1}{4} + \tfrac{1}{2\pi} \left(\tfrac{b}{1+b^2} + \arctan (b)\right)) - \sigma_Y\\
     = & \tfrac{2 \sigma_Y}{\pi}\left( \abs{\rho} \sqrt{1-\rho^2}+ \arctan(\tfrac{\abs{\rho}}{\sqrt{1-\rho^2}})\right).
     \end{align}
     where $b = \tfrac{\abs{\rho}}{\sqrt{1-\rho^2}}$ and Owen's extensive list of integrals of Gaussian functions was employed.
     Therefore, we get
     \begin{align}
         \mathbb{E}[ \abs{w^T x \cdot w^T y}] = & \norm{x} \cdot \norm{y} \cdot \tfrac{2}{\pi} \left[(1-\rho^2)^{3/2} + \rho^2 \sqrt{1-\rho^2} + \abs{\rho} \arctan(\tfrac{\abs{\rho}}{\sqrt{1-\rho^2}}) \right]\\
         = & \norm{x} \cdot \norm{y} \cdot \underbrace{\tfrac{2}{\pi} \left[\sqrt{1-\rho^2} + \abs{\rho} \arctan(\tfrac{\abs{\rho}}{\sqrt{1-\rho^2}}) \right]}_{g(\rho)},
     \end{align}
     which yields for the full expression
     \begin{equation}
         \mathbb{E}[ReLU(w^T x) \cdot ReLU(w^T y)] = \frac{\langle x, y\rangle + \norm{x} \cdot \norm{y} \cdot g(\tfrac{\langle x, y \rangle}{\norm{x} \cdot \norm{y}})}{4}.
     \end{equation}
\end{proof}

\subsection{Attention normalization in \FAVORpS}\label{sec:acc_attention_normalization}
We present the computation of the normalization term in \FAVORpS. 
Recall the usual FAVOR+ formulation from Appendix~\ref{sec:favor+_description}
\begin{equation}
    o_i  = \overbrace{\sum\limits_{j=1}^L v_j\  \phi(k_j)^T}^{A} \cross \phi(q_i) \bigg /\underbrace{\overbrace{\sum\limits_{j=1}^L \phi(k_j)^T}^{C} \cross \phi(q_i)}_{B_i}.
\end{equation}
As was remarked in Appendix~\ref{sec:favor+_description}, construction of $A$, $C$, and $\phi(q_i)\ \forall i$ each requires $\mathcal{O}(LDR)$ computational complexity. Therefore, it is central to also compute the normalization factor $B_i$ in superposition.
We start with the construction of:
\begin{equation}
    C^{(m)}_s = \underbrace{\left(\sum\limits_{j=1}^L \phi(\sum\limits_{w=1}^N k_j^{(m, w)})^T \right)}_{\text{construct } \forall m \text{ in } \mathcal{O}(L M(DR + ND))}  \qquad \qquad \qquad Q^{(n)}_{i} =  \underbrace{\phi( \sum\limits_{t=1}^M q_i^{(t, n)})}_{\text{construct } \forall i, n \text{ in } \mathcal{O}(LN(DR + MD))}
\end{equation}
Since $C^{(m)}_s \in \mathbb{R}^{1\times R}$, the evaluation of $\cross$ for all query positions $i$ and channels $(m,n)$ is relatively inexpensive, demanding only $\mathcal{O}(LMNR)$ operations. According to the two approximations ($P$) from FAVOR+ and ($H$), which is explored thoroughly in Appendix~\ref{theoretical_results_acc_attention}, it holds:
\begin{align}
  B_i^{(m,n)} =& \ C^{(m)}_s \times Q^{(n)}_{i} \\
  =& \sum\limits_{j=1}^L \phi\left(\sum\limits_{w=1}^N k_j^{(m, w)}\right)^T \phi\left(\sum\limits_{t=1}^M q_i^{(t, n)}\right) \\
  \overset{P}{\approx}& \sum\limits_{j=1}^L \exp\left(\langle \sum\limits_{w=1}^N k_j^{(m, w)},\ \sum\limits_{t=1}^M q_i^{(t, n)}\rangle / \sqrt{D}\right) \\
    \overset{H}{\approx} & \sum\limits_{j=1}^L \exp\left(\langle k_j^{(m, n)},\ q_i^{(m, n)} \rangle / \sqrt{D}\right)\label{att_clean_solution_evaluation}
\end{align}
In total, while still setting $M$=$N$ to balance the load, computing $B_i^{(m,n)}$ for all channels $(m,n)$ and query positions $i$ demands a runtime of $\mathcal{O}(LNDR + LN^2 D + LN^2R)$, a significant improvement over $\mathcal{O}(LN^2 DR)$ for computing $B_i$ 
separately for each channel $(m,n) \in \{1, \ldots, N\}^2$.

How the normalization is incorporated depends on the \MIMOFormer instantiation.
In the first case using superposition exclusively for the attention block (att.), the normalization scalar is directly applied on the output tokens after unbinding, i.e., 
\begin{align}
    o_i^{(m,n)} = \frac{S_i^{(n)}\odot \tilde{a}^{(m,n)}}{B_i^{(m,n)}}, 
\end{align}
where $\tilde{a}^{(m,n)}$ is the unbinding key. 
In the second instantiation, where additionally the MLP computes in superposition (att.+MLP), we jointly normalize the output by the sum of all normalization scalars over $m$ (where we enjoy additional computational savings by in fact already summing over $m$ in the construction of $C_s = \sum_m C_s^{(m)}$): 
\begin{align}
    \bar{S}_i^{(n)} = \frac{S_i^{(n)}}{ \sum_{m=1}^{M} B_i^{(m,n)}}.  
\end{align}

\newpage

\section{Theoretical Basis for Noise Mitigation in FAVOR+S}\label{theoretical_results_acc_attention}

As mentioned in the main text, our derivations in Section~4.1 rely on two estimates:
\begin{equation}
    \phi(k)^T \phi(q) \overset{P}{\approx} \exp(\langle k, q \rangle / \sqrt{D}) \qquad
 \langle \sum\limits_{w=1}^N k_j^{(u, w)},\ \sum\limits_{t=1}^M q_i^{(t, n)} \rangle
 \overset{H}{\approx} \underbrace{\langle k_j^{(u, n)},\ q_i^{(u, n)} \rangle}_{\text{intended signal}}
\end{equation}
The approximation $P$, which improves with increasing $R = \dim (\phi(\overline{q}_i))$, is due to FAVOR+ and is quantified in \cite{choromanski2020performer} whereas the approximation $H$ follows from:

\begin{Inter_channeldistortion}\label{informal_statement_attention_inter_channel_distortion}
    The probability that inter-channel attention distorts the intended signal of the dot product by a factor outside $[1-\alpha, 1+\alpha]$ has various upper bounds, most notably decaying exponentially w.r.t. $D\alpha^2 \cos^2(\measuredangle( \overline{k}_j^{(u,n)}, \overline{q}_i^{(u,n)} ))/(NM-1)^2$.
\end{Inter_channeldistortion}

We proceed to make this statement exact:

\begin{thm}[FAVOR+S Inter-Channel Noise]\label{thm:attention_cross_talk}
The probability that inter-channel attention distorts the true signal by a factor outside $[1-\alpha, 1+\alpha]$ shows the following tail bounds.
Denote by
\begin{equation}
    P = \mathbb{P}\left\{ \left\langle \sum\limits_{w=1}^N k_j^{(u,w)} ,\  \sum\limits_{t=1}^M q_i^{(t,n)}  \right\rangle \Bigg / \sqrt{D} \not\in \left[1 -\alpha, 1 + \alpha\right] \cdot \langle \overline{k}_j^{(u,n)},\ \overline{q}_i^{(u,n)} \rangle / \sqrt{D} \right\}
\end{equation}
where 
\begin{equation}
    k_j^{(m,n)} := \overline{k}_j^{(m,n)} \odot a^{(m,n)} \qquad q_j^{(m,n)} := \overline{q}_j^{(m,n)} \odot a^{(m,n)}
\end{equation}
with $a^{(m,n)}$ being i.i.d. bipolar vectors of Rademachers and ($m,n$) denoting a channel. It holds 
\begin{align}
    & P \leq \sum\limits_{w=1}^N \sqrt{\sum\limits_{t=1,\ldots, M}^{(u,w) \neq (t,n)} 1/\Xi_{(u,w)}^{(t,n)}}
    & P \leq \sum\limits_{t=1}^M \sqrt{\sum\limits_{w=1,\ldots, N}^{(u,w) \neq (t,n)} 1/\Xi_{(u,w)}^{(t,n)}} &\\
    & P \leq \sum\limits_{\substack{w=1,\ldots, N\\t=1,\ldots, M}}^{(u,w) \neq (t,n)} 1/\Xi_{(u,w)}^{(t,n)}
    & P \leq 2 \sum\limits_{\substack{w=1,\ldots, N\\t=1,\ldots, M}}^{(u,w) \neq (t,n)}
    \exp \left( - \tfrac{\Xi_{(u,w)}^{(t,n)}}{2(NM-1)^2} \right) &
\end{align}
for 
\begin{equation}
    \Xi_{(u,w)}^{(t,n)} = \tfrac{\alpha^2\abs{\langle \overline{k}_j^{(u,n)},\ \overline{q}_i^{(u,n)}\rangle}^2}{ \sum_{p=1}^D \left(\overline{k}_j^{(u,w)}\right)_p^2 \left(\overline{q}_i^{(t,n)}\right)_p^2} = \tfrac{\alpha^2 \cos^2(\measuredangle( \overline{k}_j^{(u,n)}, \overline{q}_i^{(u,n)})) \norm{\overline{k}_j^{(u,n)}}_2^2 \norm{\overline{q}_i^{(u,n)}}_2^2}{ \norm{ \overline{k}_j^{(u,w)} \odot  \overline{q}_i^{(t,n)}}_2^2}
\end{equation}
which for keys/queries of similar size according to Theorem \ref{thm:norm_hadamard_product} in Appendix~\ref{app:theorems} typically scales as
\begin{equation}
    \Xi_{(u,w)}^{(t,n)} \sim D\ \alpha^2 \cos^2(\measuredangle (\overline{k}_j^{(u,n)}, \overline{q}_i^{(u,n)}))
\end{equation}
\end{thm}

\begin{proof}
    First notice that since
    \begin{gather}
    \sum\limits_{\substack{w=1,\ldots, N\\t=1,\ldots, M}} \left\langle k_j^{(u,w)} ,\  q_i^{(t,n)}  \right\rangle \not\in \left[1 -\alpha, 1 + \alpha\right] \cdot \langle \overline{k}_j^{(u,n)},\ \overline{q}_i^{(u,n)} \rangle\\
    \iff \\
    \abs{\sum\limits_{\substack{w=1,\ldots, N\\t=1,\ldots, M}}^{(u,w) \neq (t,n)} \left\langle k_j^{(u,w)} ,\  q_i^{(t,n)}  \right\rangle} > \alpha \abs{\langle \overline{k}_j^{(u,n)},\ \overline{q}_i^{(u,n)}\rangle},
    \end{gather}
    we may instead derive tail bounds on
    \begin{equation}
        \mathbb{P}\left\{ \abs{\sum\limits_{\substack{w=1,\ldots, N\\t=1,\ldots, M}}^{(u,w) \neq (t,n)} \left\langle k_j^{(u,w)} ,\  q_i^{(t,n)}  \right\rangle} > \alpha \abs{\langle \overline{k}_j^{(u,n)},\ \overline{q}_i^{(u,n)}\rangle}\right\}.
    \end{equation}
    We shall derive the tail bounds for a threshold $\alpha$ and only replace it by $\alpha \abs{\langle \overline{k}_j^{(u,n)},\ \overline{q}_i^{(u,n)}\rangle}$ in a final step.
    Applying Markov, the triangle inequality, and linearity of expectation gives
    \begin{align}
    & \ \mathbb{P}\left\{ \abs{\sum\limits_{\substack{w=1,\ldots, N\\t=1,\ldots, M}}^{(u,w) \neq (t,n)} \left\langle k_j^{(u,w)} ,\  q_i^{(t,n)}  \right\rangle} > \alpha\right\}\\
    \leq & \ \mathbb{E}\left[ \abs{\sum\limits_{\substack{w=1,\ldots, N\\t=1,\ldots, M}}^{(u,w) \neq (t,n)} \left\langle k_j^{(u,w)} ,\  q_i^{(t,n)}  \right\rangle}\right]\Bigg/\alpha\\
    \leq & \sum\limits_{w=1}^N \mathbb{E}\left[ \abs{\sum\limits_{t=1,\ldots, M}^{(u,w) \neq (t,n)} \left\langle k_j^{(u,w)} ,\  q_i^{(t,n)}  \right\rangle}\right]\Bigg/\alpha\\
    = & \sum\limits_{w=1}^N \mathbb{E}\left[ \abs{\sum\limits_{t=1,\ldots, M}^{(u,w) \neq (t,n)} \left\langle \overline{k}_j^{(u,w)} \odot a^{(u,w)} ,\  \overline{q}_i^{(t,n)} \odot a^{(t,n)} \right\rangle}\right]\Bigg/\alpha,
    \label{eq:proof_attention_main}
\end{align}
where $a^{(\cdot, \cdot)}$ denotes one of the (independent) binding vectors with entries given by independent Rademacher random variables and $\bar{k}_{(\cdot)}^{(\cdot, \cdot)}$,  $\bar{q}_{(\cdot)}^{(\cdot, \cdot)}$ the unbound keys and queries respectively of a given channel and token position. For $\epsilon_p^{(t)}$ denoting independent Rademacher random variables we may simplify to 
\begin{align}
    & \ \mathbb{E}\left[ \abs{\sum\limits_{t=1,\ldots, M}^{(u,w) \neq (t,n)} \left\langle \overline{k}_j^{(u,w)} \odot a^{(u,w)} ,\  \overline{q}_i^{(t,n)} \odot a^{(t,n)} \right\rangle}\right]\\
    = & \ \mathbb{E}\left[ \abs{\sum\limits_{t=1,\ldots, M}^{(u,w) \neq (t,n)} \sum\limits_{p=1}^D \left(\overline{k}_j^{(u,w)}\right)_pa_p^{(u,w)}\left(\overline{q}_i^{(t,n)}\right)_p a_p^{(t,n)} } \right]\\
    = & \ \mathbb{E}\left[ \abs{\sum\limits_{t=1,\ldots, M}^{(u,w) \neq (t,n)} \sum\limits_{p=1}^D \left(\overline{k}_j^{(u,w)}\right)_p \left(\overline{q}_i^{(t,n)}\right)_p \epsilon_p^{(t)} } \right].
\end{align}
The famous Khintchine inequality determines up to a constant the behavior of the expectation as 
\begin{align}
    \tfrac{1}{\sqrt{2}}\sqrt{\sum\limits_{t=1,\ldots, M}^{(u,w) \neq (t,n)} \sum\limits_{p=1}^D \left(\left(\overline{k}_j^{(u,w)}\right)_p \left(\overline{q}_i^{(t,n)}\right)_p\right)^2}
    \leq & \ \mathbb{E}\left[ \abs{\sum\limits_{t=1,\ldots, M}^{(u,w) \neq (t,n)} \sum\limits_{p=1}^D \left(\overline{k}_j^{(u,w)}\right)_p \left(\overline{q}_i^{(t,n)}\right)_p \epsilon_p^{(t)} } \right]\\ 
    \leq & \sqrt{\sum\limits_{t=1,\ldots, M}^{(u,w) \neq (t,n)} \sum\limits_{p=1}^D \left(\left(\overline{k}_j^{(u,w)}\right)_p \left(\overline{q}_i^{(t,n)}\right)_p\right)^2}\label{eq:proof_attention_khintchine}.
\end{align}
Hence, the Markov bound gives
\begin{align}
    \mathbb{P}\left\{ \abs{\sum\limits_{\substack{w=1,\ldots, N\\t=1,\ldots, M}}^{(u,w) \neq (t,n)} \left\langle k_j^{(u,w)} ,\  q_i^{(t,n)}  \right\rangle} > \alpha\right\} \leq  & \sum\limits_{w=1}^N \sqrt{\sum\limits_{t=1,\ldots, M}^{(u,w) \neq (t,n)} \sum\limits_{p=1}^D \left(\overline{k}_j^{(u,w)}\right)_p^2 \left(\overline{q}_i^{(t,n)}\right)_p^2} \Bigg/ \alpha\\
    \text{and}\\
    \mathbb{P}\left\{ \abs{\sum\limits_{\substack{w=1,\ldots, N\\t=1,\ldots, M}}^{(u,w) \neq (t,n)} \left\langle k_j^{(u,w)} ,\  q_i^{(t,n)}  \right\rangle} > \alpha\right\} \leq & \sum\limits_{t=1}^M \sqrt{\sum\limits_{w=1,\ldots, N}^{(u,w) \neq (t,n)} \sum\limits_{p=1}^D \left(\overline{k}_j^{(u,w)}\right)_p^2 \left(\overline{q}_i^{(t,n)}\right)_p^2} \Bigg/ \alpha
\end{align}
where the latter follows by symmetry. Alternatively, we could also apply Chebyshev to the problem, i.e.
\begin{align}
     & \ \mathbb{P}\left\{ \abs{\sum\limits_{\substack{w=1,\ldots, N\\t=1,\ldots, M}}^{(u,w) \neq (t,n)} \left\langle k_j^{(u,w)} ,\  q_i^{(t,n)}  \right\rangle} > \alpha\right\}\\
     = & \ \mathbb{P}\left\{ \left(\sum\limits_{\substack{w=1,\ldots, N\\t=1,\ldots, M}}^{(u,w) \neq (t,n)} \left\langle k_j^{(u,w)} ,\  q_i^{(t,n)}  \right\rangle\right)^2 > \alpha^2\right\}\\
     \leq & \ \mathbb{E}\left[ \left(\sum\limits_{\substack{w=1,\ldots, N\\t=1,\ldots, M}}^{(u,w) \neq (t,n)} \left\langle k_j^{(u,w)} ,\  q_i^{(t,n)}  \right\rangle\right)^2 \right] \bigg/ \alpha^2\\
     = & \ \mathbb{E}\left[ \left(\sum\limits_{\substack{w=1,\ldots, N\\t=1,\ldots, M}}^{(u,w) \neq (t,n)}  \left\langle \overline{k}_j^{(u,w)} \odot a^{(u,w)} ,\  \overline{q}_i^{(t,n)} \odot a^{(t,n)} \right\rangle\right)^2 \right] \bigg/ \alpha^2\\
     = & \ \mathbb{E}\left[ \left(\sum\limits_{\substack{w=1,\ldots, N\\t=1,\ldots, M}}^{(u,w) \neq (t,n)} \sum\limits_{p=1}^D \left(\overline{k}_j^{(u,w)}\right)_pa_p^{(u,w)}\left(\overline{q}_i^{(t,n)}\right)_p a_p^{(t,n)}\right)^2 \right] \bigg/ \alpha^2\\
     = & \ \begin{multlined} \sum\limits_{\substack{w=1,\ldots, N\\t=1,\ldots, M}}^{(u,w) \neq (t,n)} \sum\limits_{p=1}^D \sum\limits_{\substack{w'=1,\ldots, N\\t'=1,\ldots, M}}^{(u,w') \neq (t',n)} \sum\limits_{p'=1}^D \left(\overline{k}_j^{(u,w)}\right)_p \left(\overline{q}_i^{(t,n)}\right)_p
     \left(\overline{k}_j^{(u,w')}\right)_{p'} \left(\overline{q}_i^{(t',n)}\right)_{p'} \cdot \\
     \cdot \mathbb{E}\left[ a_p^{(u,w)} a_p^{(t,n)}  a_{p'}^{(u,w')} a_{p'}^{(t',n)} \right] \bigg/ \alpha^2 \end{multlined}
\end{align}
Now, since for $(u,w) \neq (t,n)$, $(u,w') \neq (t',n)$ and $(t,w) \neq (t',w')$ the cardinality of $\{(u,w), (t,n), (u, w'), (t',n)\}$ is at least three, at least one entry has no duplicate. Let w.l.o.g. be that entry $a_p^{(u,w)}$. Then by independence, one has
\begin{equation}
    \mathbb{E} \left[ a_p^{(u,w)} a_p^{(t,n)} a_{p'}^{(u,w')} a_{p'}^{(t',n)}\right] = \mathbb{E} \left[ a_p^{(u,w)} \right] \mathbb{E} \left[ a_p^{(t,n)} a_{p'}^{(u,w')} a_{p'}^{(t',n)}\right] = 0
\end{equation}
Consequently, all terms with $(t,w) \neq (t',w')$ vanish. Also, for $(u,w) \neq (t,n)$ and $p \neq p'$ all four entries are independent, i.e.
\begin{equation}
    \mathbb{E} \left[ a_p^{(u,w)} a_p^{(t,n)} a_{p'}^{(u,w)} a_{p}^{(t,n)}\right] = \mathbb{E} \left[ a_p^{(u,w)} \right] \mathbb{E} \left[ a_p^{(t,n)} \right] \mathbb{E} \left[a_{p}^{(u,w)}\right] \mathbb{E} \left[a_{p}^{(t,n)}\right] = 0
\end{equation}
Hence, we have that all cross-terms vanish, which gives
\begin{align}
    & \ \mathbb{P}\left\{ \abs{\sum\limits_{\substack{w=1,\ldots, N\\t=1,\ldots, M}}^{(u,w) \neq (t,n)} \left\langle k_j^{(u,w)} ,\  q_i^{(t,n)}  \right\rangle} > \alpha\right\}\\
    \leq & \sum\limits_{\substack{w=1,\ldots, N\\t=1,\ldots, M}}^{(u,w) \neq (t,n)} \sum\limits_{p=1}^D \left(\overline{k}_j^{(u,w)}\right)_p^2 \left(\overline{q}_i^{(t,n)}\right)_p^2
     \mathbb{E}\left[ \left(a_p^{(u,w)}\right)^2 \left(a_p^{(t,n)}\right)^2 \right] \bigg/ \alpha^2\\
     = & \sum\limits_{\substack{w=1,\ldots, N\\t=1,\ldots, M}}^{(u,w) \neq (t,n)} \sum\limits_{p=1}^D \left(\overline{k}_j^{(u,w)}\right)_p^2 \left(\overline{q}_i^{(t,n)}\right)_p^2 \bigg/ \alpha^2
\end{align}
For higher orders than two, we can no longer rely on the absence of duplicates in the set of multiplied binding vectors. For example, a cross-term such as 
\begin{equation}
    \mathbb{E} \left[ \underbrace{a_p^{(u,w)} a_p^{(t,n)}}_1 \underbrace{a_{p'}^{(u,w')} a_{p'}^{(t',n)}}_2 \underbrace{a_p^{(u,w)} a_p^{(t,n)}}_3 \underbrace{a_{p'}^{(u,w')} a_{p'}^{(t',n)}}_4\right] = 1
\end{equation}
is non-vanishing even for $(t,w,p) \neq (t', w', p')$. Hence, for higher orders, we will have to work with 
\begin{align}
    & \ \mathbb{P}\left\{ \abs{\sum\limits_{\substack{w=1,\ldots, N\\t=1,\ldots, M}}^{(u,w) \neq (t,n)} \left\langle k_j^{(u,w)} ,\  q_i^{(t,n)}  \right\rangle} > \alpha\right\} \leq \mathbb{P}\left\{ \sum\limits_{\substack{w=1,\ldots, N\\t=1,\ldots, M}}^{(u,w) \neq (t,n)} \abs{\left\langle k_j^{(u,w)} ,\  q_i^{(t,n)}  \right\rangle} > \alpha\right\}\\
    \leq &\  \mathbb{P}\left\{ \exists (u,w) \neq (t,n) : \abs{\left\langle k_j^{(u,w)} ,\  q_i^{(t,n)}  \right\rangle} > \alpha/(NM-1)\right\}\\
    \leq & \sum\limits_{\substack{w=1,\ldots, N\\t=1,\ldots, M}}^{(u,w) \neq (t,n)} \mathbb{P}\left\{ \abs{\left\langle k_j^{(u,w)} ,\  q_i^{(t,n)}  \right\rangle} > \alpha/(NM-1)\right\}\\
    = & \sum\limits_{\substack{w=1,\ldots, N\\t=1,\ldots, M}}^{(u,w) \neq (t,n)} \mathbb{P}\left\{ \abs{\sum\limits_{p=1}^D \left(\overline{k}_j^{(u,w)}\right)_p a_p^{(u,w)} \left(\overline{q}_i^{(t,n)}\right)_p a_p^{(t,n)}} > \alpha/(NM-1)\right\}\\
    = & \sum\limits_{\substack{w=1,\ldots, N\\t=1,\ldots, M}}^{(u,w) \neq (t,n)} \mathbb{P}\left\{ \abs{\sum\limits_{p=1}^D \left(\overline{k}_j^{(u,w)}\right)_p \left(\overline{q}_i^{(t,n)}\right)_p \epsilon_p} > \alpha/(NM-1)\right\}
\end{align}
for $\{\epsilon_p\}_{p=1}^D$ independent Rademacher random variables. Finally, we may apply Hoeffding's inequality (Theorem \ref{thm:hoeffding's inequality}), which gives
\begin{equation}
    \mathbb{P}\left\{ \abs{\sum\limits_{\substack{w=1,\ldots, N\\t=1,\ldots, M}}^{(u,w) \neq (t,n)} \left\langle k_j^{(u,w)} ,\  q_i^{(t,n)}  \right\rangle} \geq \alpha\right\} \leq 2 \sum\limits_{\substack{w=1,\ldots, N\\t=1,\ldots, M}}^{(u,w) \neq (t,n)}
    \exp \left( - \tfrac{\alpha^2}{2(NM-1)^2 \sum_{p=1}^D \left(\overline{k}_j^{(u,w)}\right)_p^2 \left(\overline{q}_i^{(t,n)}\right)_p^2} \right)
\end{equation}
\end{proof}

\newpage

\section{Experimental Setup and Ablation Study on \MIMOConv}\label{sec:results}

\subsection{Experimental setup}

\subsubsection*{Datasets}

\paragraph{CIFAR10 and CIFAR100.} The CIFAR10~\cite{krizhevsky2009learning} dataset contains 60,000 images, each of resolution $32\times32$, divided into 50,000 training and 10,000 test images.
The images are grouped into ten classes, each with 6000 examples. 
CIFAR100 has the same number of images and resolutions, but contains 100 classes each with 600 examples.

\paragraph{MNIST.} The MNIST dataset~\cite{lecun1998gradient} provides grey-scale images, each of resolution 28$\times$28, containing hand-written digits. 
The 60,000 training samples are divided into a training and validation set containing 55,000 and 5000 samples, respectively. 
Finally, the test accuracy is reported on the test set containing 10,000 samples. 

\paragraph{SVHN.} The street view house number (SVHN) dataset~\cite{yuval2011reading} provides cropped images of house number plates, each of resolution 32$\times$32. 
As in MNIST, the task is to classify the printed digits (from 0 to 9). 
It contains 73,257 RGB images for training and 26,032 for testing. 

\subsubsection*{Training setups}

The experiments are run on an NVIDIA A100 Tensor Core GPU with 80\,GB memory and 8 CPU cores.
All experiments are repeated five times with different random seeds. We report the mean and standard deviations of accuracy to account for variability in training.
Overall, all \MIMOConv experiments together required 3290 GPU hours when accumulating all the training runs required for generating the main results and the ablations study. 

\paragraph{CIFAR10 and CIFAR100.} In all experiments on CIFAR10 and CIFAR100, stochastic gradient descent (SGD) with momentum is used. 
Unless otherwise noted, we train for 1200 epochs using the OneCycleLR policy~\cite{smith2019clr} with cosine annealing for two phases (30\% increase, 70\% decrease of learning rate). 
The initial learning rate is set to 0.008, the maximal learning rate to 0.2, and the final learning rate to 2e-5. 
Momentum is cycled inversely with base momentum set to 0.85 and maximal momentum set to 0.95. 
Due to overfitting, WideResNet28-10 shows higher test accuracy when trained with 200 epochs than 1200 epochs; hence, Table~1 in the main text shows WideResNet28-10's performance with 200 training epochs. 
For all parameters, except for the bias in shifted ReLU, the slope in parametric ReLU, and binding/unbinding keys, weight decay with value 1e-5 is applied.

The images are standardized on each color channel by subtracting the mean and dividing by the standard deviation. To augment data, the images are randomly flipped horizontally, and a random $32\times32$ crop is taken after zero padding the images on each side by four pixels. Furthermore, the data agnostic augmentation strategy mixup~\cite{zhang2018mixup} is employed with parameter $\alpha$=$1$, which is decisive for obtaining high accuracy. 

The batch size is set to $128$ elements per superposition channel (i.e., $128N$). 
Thus, after binding, a batch of 128 superpositions traverses the CNN. 
Increasing superposition channels would decrease the number of update steps per epoch. 
To correct for that, in each epoch, the dataset is traversed as often as the number of superposition channels used. 
%
%
While the results presented in this paper come from a train/test split of the datasets, the training dataset is split into a 90/10 train/validation split for all model design and hyperparameter choices.
Furthermore, to decrease the degrees of freedom in the experiments, the remaining hyperparameters (learning rate, weight decay, mixup parameters) are tuned to yield good performance on the base model WideResNet28-10.
Finally, to stabilize training, the average gradient norm of each epoch is recorded, and the batches of the subsequent epoch are discarded (without repetition) if their update gradient norm exceeds the recorded average of the last epoch by a factor of 10.

Training \MIMOConv for 1200 epochs takes 11 hours independent of the number of superposition channels owing to the batch loading corrections that account for the same number of training steps. 

\paragraph{MNIST.} The experiments on MNIST use a similar setup to the ones on CIFAR: the same learning rate scheduler, batch size, weight decay, and mixup coefficients are used. 
In contrast to CIFAR, the number of training epochs is reduced to 50. 
Moreover, the images are center-cropped to  20$\times$20 pixels both in training and testing. 
A random horizontal flip serves as data augmentation during training. 

We reduce the depth of \MIMOConv from 28 to 10 layers, and the width factor from $10\times$ to  $1\times$, hence we call it \MIMOConv-10-1. 
Moreover, the initial width factor of the first convolutional layer is also set to $1\times$. 

\paragraph{SVHN.} In the SVHN experiments, we train the standard \MIMOConv-28-10 architecture for 200 epochs. 
The remaining hyperparameters are kept the same as in CIFAR. 
During training, random crop with padding is used. 

\subsection{Computational complexity}\label{sec:complexity}

\begin{table}[t]
\caption{Millions of multiply-accumulate (MMAC) operations per sample on CIFAR-100. Number in parenthesis shows the relative share of the overall complexity.}
\label{tab:mac_mimoconv}
\centering
\resizebox{\linewidth}{!}{
\begin{tabular}{lllllll}
\toprule
                 & \multicolumn{1}{l}{\begin{tabular}[c]{@{}c@{}}First conv. \\ layer\end{tabular}} & Binding   & \multicolumn{1}{l}{\begin{tabular}[c]{@{}c@{}}Rest of\\ conv. layers\end{tabular}}  & Unbinding & FCL        & \textbf{Total}     \\
\cmidrule(r){1-1}\cmidrule(r){2-7}
WideResNet-28-10 & 0.49 (0.009\%)        & n.a.          & 5245 (99.99\%)   & n.a.         & 0.064 (0.001\%)  & \textbf{5251}   \\
WideIsoNet-28-10 & 0.49 (0.009\%)         & n.a.          & 5245 (99.99\%)  & n.a.        & 0.064 (0.001\%)  & \textbf{5251}  \\
\cmidrule(r){1-1}\cmidrule(r){2-7}
MIMOConv (N=1)   & 1.97 (0.04\%)           & 4.19 (0.08\%)  & 5329 (99.88\%)  & 0.41 (0.008\%)  & 0.064 (0.001\%)  & \textbf{5335}   \\
MIMOConv (N=2)   & 1.97 (0.07\%)          & 4.19 (0.15\%)  & 2664 (99.75\%)  & 0.41 (0.015\%)  & 0.064 (0.002\%)  & \textbf{2671}   \\
MIMOConv (N=4)   & 1.97 (0.15\%)          & 4.19 (0.31\%)   & 1332 (99.50\%)  & 0.41 (0.031\%)   & 0.064 (0.005\%)  & \textbf{1339}  \\
\bottomrule
\end{tabular}
}
\end{table}

Table~\ref{tab:mac_mimoconv} breaks down \MIMOConv's computational benefits (in MMAC) on CIFAR100, as $N$ is increased from $1$ to $4$.  
As shown, the integration of variable binding mechanisms via binding and unbinding operations is inconsequential, amounting to only between 0.008\% ($N$=$1$) and 0.031\% ($N$=$4$) of the total MACs for MIMOConv. 

On MNIST, \MIMOConv-10-1 has a computational cost of 5.10\,MMAC per sample at $N$=$1$ superposition channels and manages to reduce the cost to 0.47\,MMAC per sample at $N$=$16$, an effective reduction of $10.9\times$. 
The reduction is smaller than $N$ due to the computationally dominating first convolutional layer, which is not operating in superposition. 
Yet, \MIMOConv shows a notably higher reduction than the LeNet-like model (CNN+nonlinear ($8\times$)) from DataMUX~\cite{murahari2022datamux}, a key competitor, which requires 0.88\,MMAC and 0.65\,MMAC per sample at $N$=$1$ and $N$=$16$, respectively. 

\subsection{The effectiveness of position-wise binding (\PWHRR) and isometry regularization}
DataMux~\cite{murahari2022datamux} also explored CNNs that compute in superposition and reported its findings on MNIST. Even with a trivial downsizing for fair comparison from a 28-layer very-wide ($10\times$) MIMOConv to a 10-layer narrow ($1\times$) MIMOConv, our method scales much better to high superposition channels (N) than DataMUX does. Indeed, our model shows an accuracy of $80.4\%$ against their $52.9\%$ in case of $N$=$16$ superposition channels (highest number of channels reported by DataMUX for vision tasks), despite being computationally cheaper ($0.47$\,MMAC/s vs. $0.65$\,MMAC/s). DataMux's binding overhead results in a mere $1.35\times$ reduction in MACs compared to our $10.9\times$ as $N$ goes from $1$ to $16$ demonstrating superior scaling of our method.

We attribute the improved performance to a set of innovations which we reiterate here: MIMOConv applies position-wise binding (\PWHRR), thus retaining the locality property present in natural images and vital for CNNs, whilst as discussed by Murahari, Vishvak, et al. their primary binding does not. As a workaround they proposed binding via two layers CNNs each outputting 8 feature maps. The resulting (pixel-wise) superposition in a low-dimensional space (8-D) leads to high interference. Additionally to using an expensive binding mechanism, it also makes the first layer of the model 8 times as expensive no matter the number of superpositions. We are able to circumvent this issue by applying the first layer of the CNN \textit{before} the pixel-wise binding, increasing the dimensionality of each pixel in an easy-to-understand manner. 

Another difference to their work is our use of \textit{isometric neural networks} to further reduce interference during the processing of superposed images.

\subsection{Dynamic inference}\label{results:dynamic_inference}
Dynamic inference enables the instantaneous on-demand partitioning of the superposition channels to select an operating point with a suitable speed/accuracy trade-off. 
Even though every \MIMOConv can be configured to perform dynamic inference at any time, exposing \MIMOConv to dynamic switching between different modes during training is beneficial. 
We set up a model with four channels and consider a fast (4 inputs/pass), normal (2 inputs/pass), and slow mode (1 input/pass). 
The fast mode maps each input to one channel; the normal mode distributes two inputs over pairs of channels; and the slow mode uses all channels for the same input.
We then train the models for different frequencies in fast and slow modes. 
The normal mode is not used during training of the model. 
The potential switching between the modes happens between every batch. 

\begin{figure}[t]
 \centering
 \begin{subfigure}{0.49\linewidth}
 \centering
 \resizebox{\columnwidth}{!}{
\includegraphics{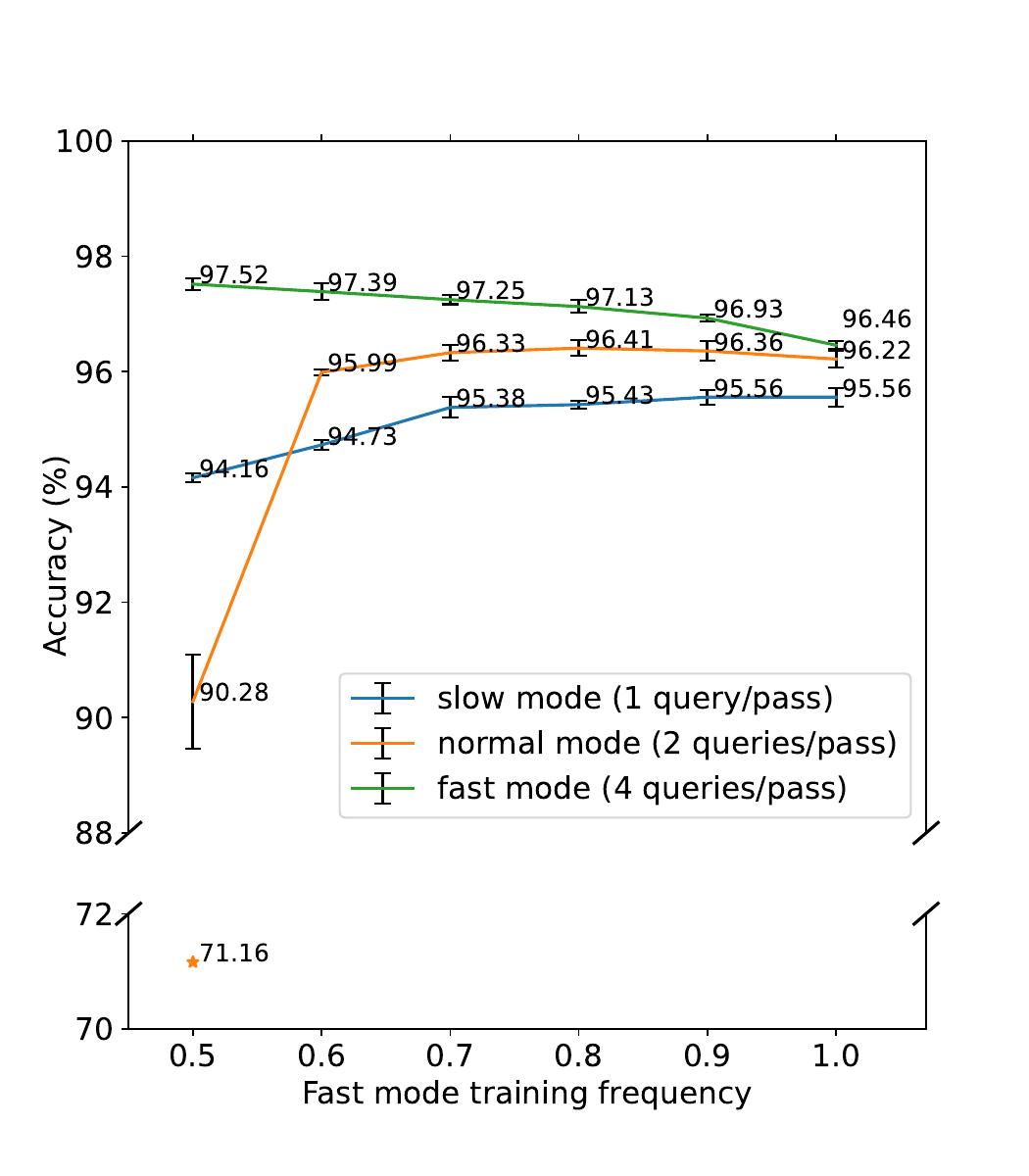}
}
 \caption{CIFAR10}
 \label{experiment2_cifar10}
 \end{subfigure}
 \begin{subfigure}{0.49\linewidth}
 \centering
 \resizebox{\columnwidth}{!}{
\includegraphics{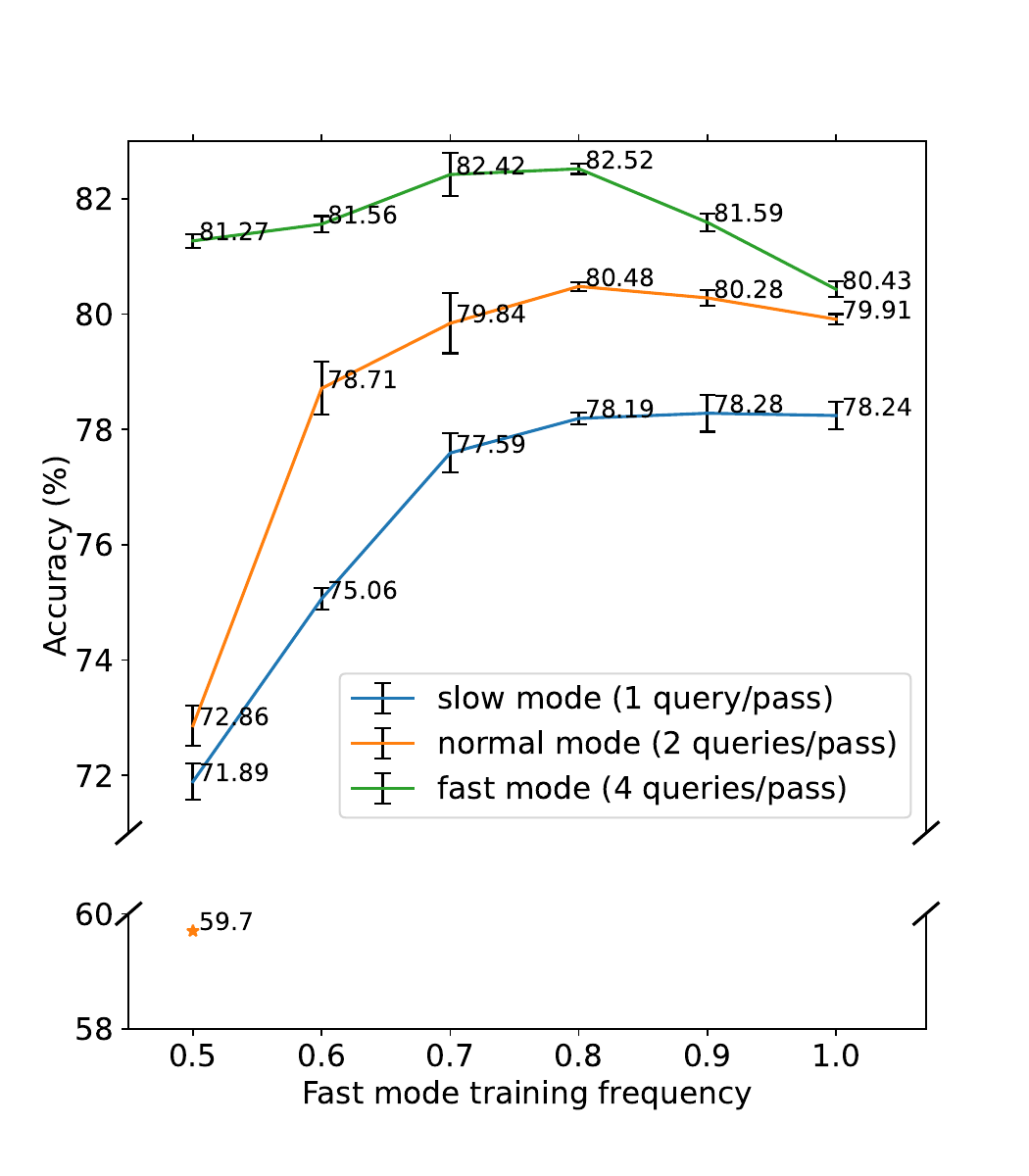}
}
 \caption{CIFAR 100}
 \label{fig:dynamic_inference}
 \end{subfigure}
 \caption{Dynamic inference with \MIMOConv trained for 1200 epochs in slow and fast mode depending on the fast mode training frequency. Each model is evaluated in slow (1 input/pass), normal (2 inputs/pass), and fast mode (4 inputs/pass). We report the average accuracy and the standard deviation (error bars) over five runs with different seeds. Outliers in normal mode (indicated with $\star$) are not used for standard deviation computation. }
 \label{experiment2}
 \end{figure}

Figure~\ref{experiment2} shows the classification accuracy on CIFAR10 (a) and CIFAR100 (b) when using dynamic models trained with varying fast mode frequencies. 
The models, which are trained with a different fraction of inputs in fast mode, are evaluated in slow, normal, and fast modes. 
As is expected, increasing the fast mode training frequency is beneficial for both datasets when only looking at the fast mode inference.
Conversely, the normal inference mode benefits from a mixture of fast and slow mode training, whereby a fast mode training frequency of 80\% achieves the highest accuracy on both datasets. 
There is a notable volatility in the performance of normal mode at a fast mode training frequency of $0.5$, which could be due to the optimization getting stuck in a local optimum exclusively learning a slow mode instance.

\subsection{Ablation study on CIFAR10/100}\label{sec:ablation_mimoconv}

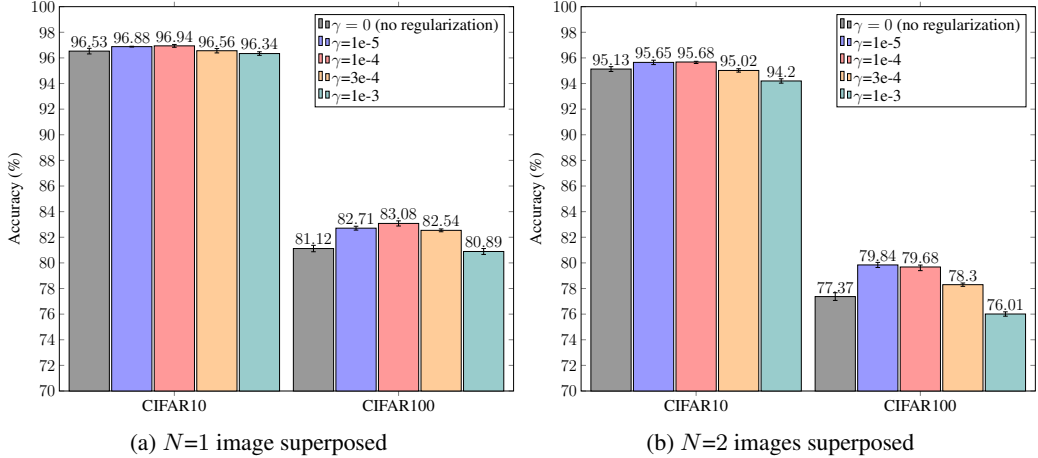
\begin{figure}[t]
\centering
\begin{subfigure}{0.49\linewidth}
\resizebox{\linewidth}{!}{
    \begin{tikzpicture}
    \begin{axis}[
        style={font=\Large},
        symbolic x coords={CIFAR10,CIFAR100},
        xtick=data,
        ybar,
        bar width=35pt,
        nodes near coords,
        legend cell align={left},
        enlarge x limits={abs=100pt},
        ymin=70,
        ymax=100,
        ylabel={Accuracy (\%)},
        width = 440pt
    ]
    \addplot[style={fill=black!40, draw=black}, error bars/.cd, y dir=both, y explicit] coordinates 
    {(CIFAR10,96.53) += (0, 0.22) -= (0, 0.22)
     (CIFAR100,81.12) += (0, 0.25) -= (0, 0.25)};
    \addplot[style={fill=blue!40, draw=black}, error bars/.cd, y dir=both, y explicit] coordinates 
    {(CIFAR10,96.88) += (0, 0.03) -= (0, 0.03)
     (CIFAR100,82.71) += (0, 0.15) -= (0, 0.15)};
    \addplot[style={fill=red!40, draw=black}, error bars/.cd, y dir=both, y explicit] coordinates 
     {(CIFAR10,96.94) += (0, 0.11) -= (0, 0.11)
     (CIFAR100,83.08) += (0, 0.20) -= (0, 0.20)};
    \addplot[style={fill=orange!40, draw=black}, error bars/.cd, y dir=both, y explicit] coordinates 
    {(CIFAR10,96.56) += (0, 0.17) -= (0, 0.17)
     (CIFAR100,82.54) += (0, 0.11) -= (0, 0.11)};
    \addplot[style={fill=teal!40, draw=black}, error bars/.cd, y dir=both, y explicit] coordinates 
    {(CIFAR10,96.34) += (0, 0.14) -= (0, 0.14)
     (CIFAR100,80.89) += (0, 0.23) -= (0, 0.23)};
    \legend{$\gamma=0$ (no regularization), $\gamma$=1e-5, $\gamma$=1e-4, $\gamma$=3e-4, $\gamma$=1e-3}
    \end{axis}
\end{tikzpicture}
}
\caption{$N$=$1$ image superposed}
\label{experiment10_one_img}
\end{subfigure}
\begin{subfigure}{0.49\linewidth}
\resizebox{\linewidth}{!}{
    \begin{tikzpicture}
    \begin{axis}[
        style={font=\Large},
        symbolic x coords={CIFAR10,CIFAR100},
        xtick=data,
        ybar,
        bar width=35pt,
        nodes near coords,
        legend cell align={left},
        enlarge x limits={abs=100pt},
        ymin=70,
        ymax=100,
        ylabel={Accuracy (\%)},
        width = 440pt
    ]
    \addplot[style={fill=black!40, draw=black}, error bars/.cd, y dir=both, y explicit] coordinates 
    {(CIFAR10,95.13) += (0, 0.19) -= (0, 0.19)
     (CIFAR100,77.37) += (0, 0.30) -= (0, 0.30)};
    \addplot[style={fill=blue!40, draw=black}, error bars/.cd, y dir=both, y explicit] coordinates 
    {(CIFAR10,95.65) += (0, 0.17) -= (0, 0.17)
     (CIFAR100,79.84) += (0, 0.20) -= (0, 0.20)};
    \addplot[style={fill=red!40, draw=black}, error bars/.cd, y dir=both, y explicit] coordinates 
     {(CIFAR10,95.68) += (0, 0.055) -= (0, 0.115)
     (CIFAR100,79.68) += (0, 0.15) -= (0, 0.285)};
    \addplot[style={fill=orange!40, draw=black}, error bars/.cd, y dir=both, y explicit] coordinates 
    {(CIFAR10,95.02) += (0, 0.135) -= (0, 0.13)
     (CIFAR100,78.30) += (0, 0.13) -= (0, 0.13)};
    \addplot[style={fill=teal!40, draw=black}, error bars/.cd, y dir=both, y explicit] coordinates 
    {(CIFAR10,94.20) += (0, 0.17) -= (0, 0.17)
     (CIFAR100,76.01) += (0, 0.17) -= (0, 0.17)};
    \legend{$\gamma=0$ (no regularization), $\gamma$=1e-5, $\gamma$=1e-4, $\gamma$=3e-4, $\gamma$=1e-3}
    \end{axis}
\end{tikzpicture}
}
\caption{$N$=$2$ images superposed}
\label{experiment10_two_imgs}
\end{subfigure}
\caption{\MIMOConv with varying orthogonal regularization coefficient ($\gamma$) for $N$=$1$ and $N$=$2$ superpositions. All models are trained for 200 epochs. We report the average accuracy and the standard deviation (error bars) over five runs with different seeds.}
\label{experiment10}
\end{figure}

\paragraph{Isometry regularization of CNN weights.}
We evaluate the impact of isometry regularization to the CNN weights by varying the orthogonal regularization coefficient ($\gamma$), described in Eq.~\eqref{eq:isometry} and Eq.~\eqref{eq:isometry_transposed}.
All models are trained for 200 epochs to reduce the training time. 
As can be observed in Figure \ref{experiment10}, orthogonal regularization enables the network to perform notably better both for configurations with a single superposition channel and two superposition channels. 
However, a strong regularization hinders the ability of the network to adapt to the task. With two superposition channels, the performance difference is more striking, i.e., orthogonal regularization is more important. In all other experiments, an orthogonal regularization coefficient of $\gamma$=1e-4 was used. 

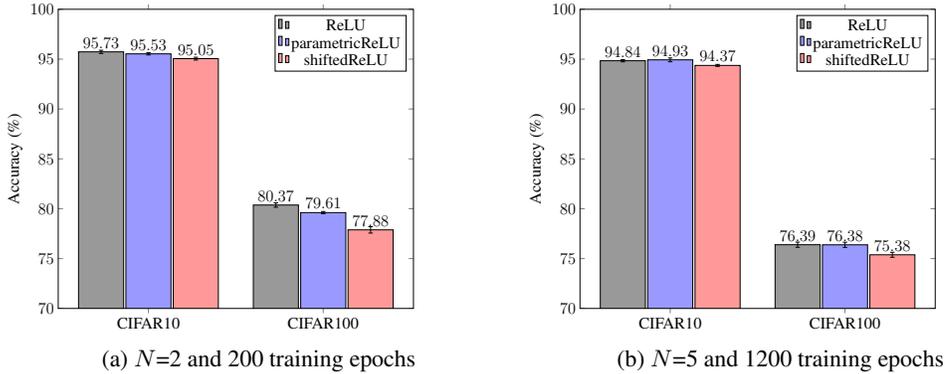
\begin{figure}[t]
\centering
\begin{subfigure}{0.49\linewidth}
    \resizebox{0.8\linewidth}{!}{
    \begin{tikzpicture}
        \begin{axis}[
            style={font=\Large},
            symbolic x coords={CIFAR10,CIFAR100},
            xtick=data,
        	ybar,
            bar width=40pt,
            nodes near coords,
            enlarge x limits={abs=80pt},
            ymin=70,
            ymax=100,
            ylabel={Accuracy (\%)},
            width = 360pt
        ]
        \addplot[style={fill=black!40, draw=black}, error bars/.cd, y dir=both, y explicit] coordinates 
        {(CIFAR10,95.73) += (0, 0.16) -= (0, 0.16)
        (CIFAR100,80.37) += (0, 0.22) -= (0, 0.22)};
        \addplot[style={fill=blue!40, draw=black}, error bars/.cd, y dir=both, y explicit] coordinates 
        {(CIFAR10,95.53) += (0, 0.11) -= (0, 0.11)
        (CIFAR100,79.61) += (0, 0.11) -= (0, 0.11)};
        \addplot[style={fill=red!40, draw=black}, error bars/.cd, y dir=both, y explicit] coordinates 
        {(CIFAR10,95.05) += (0, 0.14) -= (0, 0.14)
        (CIFAR100,77.88)  += (0, 0.32) -= (0, 0.32)};
        \legend{ReLU, parametricReLU, shiftedReLU}
        \end{axis}
    \end{tikzpicture}}
    \caption{$N$=2 and 200 training epochs}
    \label{experiment5_comparison200}
\end{subfigure}
\begin{subfigure}{0.49\linewidth}
    \resizebox{0.8\linewidth}{!}{
    \begin{tikzpicture}
        \begin{axis}[
            style={font=\Large},
            symbolic x coords={CIFAR10,CIFAR100},
            xtick=data,
        	ybar,
            bar width=40pt,
            nodes near coords,
            enlarge x limits={abs=80pt},
            ymin=70,
            ymax=100,
            ylabel={Accuracy (\%)},
            width = 360pt
        ]
        \addplot[style={fill=black!40, draw=black}, error bars/.cd, y dir=both, y explicit] coordinates 
        {(CIFAR10,94.84) += (0, 0.11) -= (0, 0.11)
        (CIFAR100,76.39) += (0, 0.26) -= (0, 0.26)};
        \addplot[style={fill=blue!40, draw=black}, error bars/.cd, y dir=both, y explicit] coordinates 
        {(CIFAR10,94.93) += (0, 0.19) -= (0, 0.19)
         (CIFAR100,76.38) += (0, 0.26 ) -= (0, 0.26)};
        \addplot[style={fill=red!40, draw=black}, error bars/.cd, y dir=both, y explicit] coordinates 
        {(CIFAR10,94.37) += (0, 0.10) -= (0, 0.10)
        (CIFAR100,75.38)  += (0, 0.25) -= (0, 0.25)};
        \legend{ReLU, parametricReLU, shiftedReLU}
        \end{axis}
    \end{tikzpicture}}
    \caption{$N$=5 and 1200 training epochs}
    \label{experiment5_comparison1200}
\end{subfigure}
\caption{\MIMOConv with different activation functions. We report the average accuracy and the standard deviation (error bars) over five runs with different seeds.}
\label{experiment5_comparison}
\end{figure}

\paragraph{Isometry at activation functions.}\label{results:isometry_at_activation_function}
We investigate the ReLU, shifted ReLU, and parametric ReLU activation functions to give the model more control over the extent of isometry. 
Each activation function owns separate, trainable parameters which are not shared between the feature maps and layers. 
%
%
Experimental results are shown in Figure~\ref{experiment5_comparison}. 
When using $N$=$2$ superposition channels and 200 training epochs, ReLU outperforms both parametric ReLU and shifted ReLU. 
For longer training times (1200 epochs) and more superposition channels ($N$=$5$), the network prefers parametric ReLU.  

\begin{figure}[t]
\centering
       \includegraphics[width=0.45\linewidth]{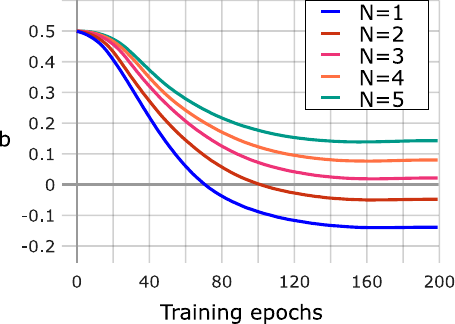}
    \caption{Average parametric ReLU parameter ($b$) during training on CIFAR10 for different number of superposition channels ($N$). 
    }
    \label{experiment1_prelu_evolution}
\end{figure}

Surprisingly, in the case of low superposition counts, the model develops highly non-isometric parametric ReLU activation functions, as seen in Figure~\ref{experiment1_prelu_evolution}. 
With the convolutional layers being pushed toward isometry due to the isometry regularization term and residual skip connections increasing isometry further, the network seeks balance through strongly non-isometric activation functions. 
Nevertheless, increasing the number of images superposed incentivizes the network to learn isometric activation functions.
It is unclear if the performance degradation induced by superpositions originates in interference or in the attempt of the network to reduce interference through isometry, and it is likely that some balance between the two is reached. Further research will be needed to gain more insight into the benefits and drawbacks of isometry. We employ parametric ReLU in all other experiments as its performance is comparable to ReLU, but it allows more degrees of freedom and hence could give additional performance benefits under different network configurations.

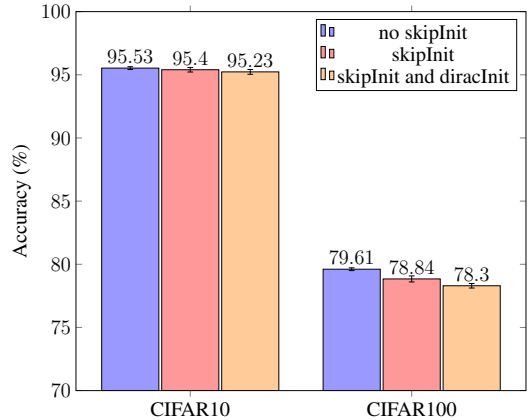
\begin{figure}[t]
\centering
\resizebox{0.5\columnwidth}{!}{
\begin{tikzpicture}
    \begin{axis}[
        style={font=\Large},
        symbolic x coords={CIFAR10,CIFAR100},
        xtick=data,
        ybar,
        bar width=40pt,
        nodes near coords,
        enlarge x limits={abs=80pt},
        ymin=70,
        ymax=100,
        ylabel={Accuracy (\%)},
        width = 360pt
    ]
    \addplot[style={fill=blue!40, draw=black}, error bars/.cd, y dir=both, y explicit] coordinates 
        {(CIFAR10,95.53) += (0, 0.11) -= (0, 0.11)
        (CIFAR100,79.61) += (0, 0.11) -= (0, 0.11)};
    \addplot[style={fill=red!40, draw=black}, error bars/.cd, y dir=both, y explicit] coordinates 
    {(CIFAR10,95.40) += (0, 0.18) -= (0, 0.18)
     (CIFAR100,78.84) += (0, 0.24) -= (0, 0.24)};
    \addplot[style={fill=orange!40, draw=black}, error bars/.cd, y dir=both, y explicit] coordinates 
    {(CIFAR10,95.23) += (0, 0.19) -= (0, 0.19)
     (CIFAR100,78.30)  += (0, 0.18) -= (0, 0.18)};
    \legend{no skipInit, skipInit, skipInit and diracInit}
    \end{axis}
\end{tikzpicture}
}
\caption{Effect of skipInit and diracInit on accuracy in superposition mode $N$=$2$, 200 epochs. We report the average accuracy and the standard deviation (error bars) over five runs with different seeds.}
\label{experiment6and7_accuracy}
\end{figure}

\begin{figure}[t]
 \centering
 \begin{subfigure}{0.49\linewidth}
 \centering
 \resizebox{\columnwidth}{!}{
\includegraphics{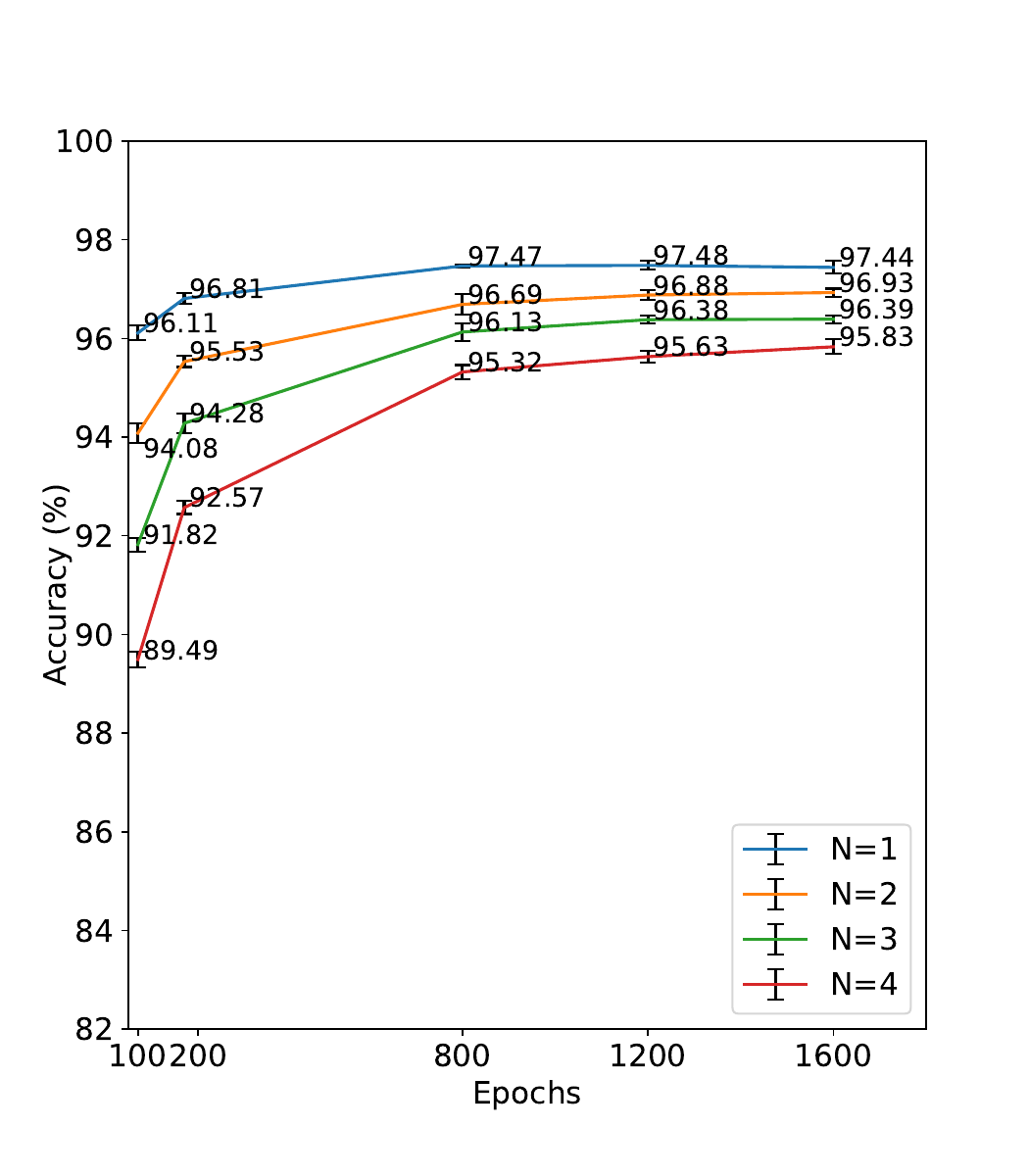}
}
 \caption{CIFAR10}
 \end{subfigure}
 \begin{subfigure}{0.49\linewidth}
 \centering
 \resizebox{\columnwidth}{!}{
\includegraphics{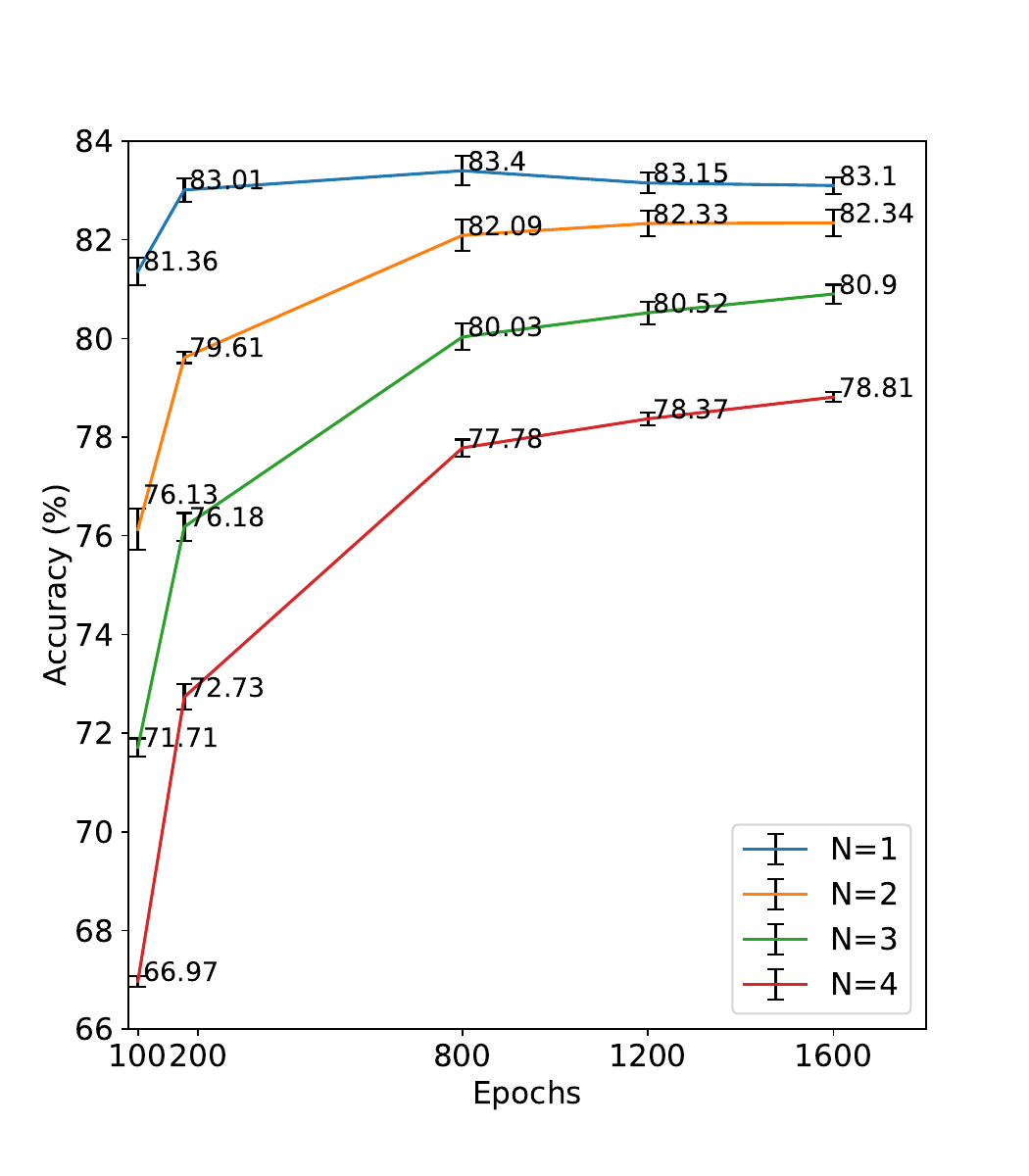}
}
 \caption{CIFAR100}
 \end{subfigure}
 \caption{Accuracy of \MIMOConv when trained with a variable number of epochs. We report the average accuracy and the standard deviation (error bars) over five runs with different seeds.}
\label{experiment4}
 \end{figure}

\paragraph{SkipInit and DiracInit.}\label{results:skipInit_diracInit}
In~\cite{qi2020deep}, the benefit of skip initialization \cite{de2020batch} via inducing maximum isometry at initialization was discussed. Furthermore, an initialization scheme for the convolutional layers as an identity, called \emph{diracInit}, was promoted. In our experiments, both additions worsen the performance on CIFAR10 and CIFAR100, as seen in Figure~\ref{experiment6and7_accuracy}.
\paragraph{Number of training epochs.}
Training a neural network model to simultaneously handle multiple images while providing high accuracy is a more difficult task than without superposition present.
Figure~\ref{experiment4} shows that the performance gap between single-image mode and multiple-image mode narrows when training for more epochs. 
For each epoch, the training set is passed through the network as many times as superposition channels were used: this corrects for the larger batch size used in superposition modes and to have roughly equal training time for each epoch.  

\paragraph{Number of feature maps in the first layer.}\label{results:initial_width}
Compared to a standard ResNet, the WideResNet architecture~\cite{BMVC2016_87} increases the number of feature maps of every layer by a \textit{width factor}. 
The only exception is the first layer, which has 16 feature maps in WideResNet-28-10. 
However, binding takes place after the first layer in \MIMOConv. In order to enter the regime of high dimensionality and to benefit from the Blessing of Dimensionality, we experiment with an additional parameter termed \textit{initial width factor}, which increases the number of feature maps of the output of the first convolutional layer. 
For example, an initial width factor of $4$ yields $4\cdot16$=$64$ feature maps after the first convolutional layer. 
The initial width factor can be configured independently from the general width factor. 

\begin{figure}[t]
\centering
\resizebox{0.5\columnwidth}{!}{
\begin{tikzpicture}
    \begin{axis}[
        style={font=\Large},
        symbolic x coords={CIFAR10,CIFAR100},
        xtick=data,
        ybar,
        bar width=40pt,
        nodes near coords,
        enlarge x limits={abs=100pt},
        ymin=0,
        ymax=100,
        ylabel={Accuracy (\%)},
        legend pos=south west,
        width = 440pt
    ]
    \addplot[style={fill=blue!40, draw=black}, error bars/.cd, y dir=both, y explicit] coordinates 
    {(CIFAR10,95.57) += (0, 0.17) -= (0, 0.17)
     (CIFAR100,63.59) += (0, 34.99) -= (0, 34.99)};
    \addplot[style={fill=red!40, draw=black}, error bars/.cd, y dir=both, y explicit] coordinates 
    {(CIFAR10,95.49) += (0, 0.15) -= (0, 0.15)
     (CIFAR100,79.63) += (0, 0.29) -= (0, 0.29)};
    \addplot[style={fill=orange!40, draw=black}, error bars/.cd, y dir=both, y explicit] coordinates 
    {(CIFAR10,95.53) += (0, 0.11) -= (0, 0.11)
     (CIFAR100,79.61) += (0, 0.11) -= (0, 0.11)};
    \addplot[style={fill=teal!40, draw=black}, error bars/.cd, y dir=both, y explicit] coordinates 
    {(CIFAR10,95.73) += (0, 0.15) -= (0, 0.15)
     (CIFAR100,42.17) += (0, 38.00) -= (0, 38.00)};
    \legend{width factor 10/1, width factor 10/2, width factor 10/4, width factor 10/8}
    \end{axis}
\end{tikzpicture}
}
\caption{Initial width majorly affects stability in \MIMOConv ($N$=2, 200 epochs). Configurations are labeled with \textit{width factor}/\textit{initial width factor}. Standard initial width factor of vanilla WideResNet would be $1$. We report the average accuracy and the standard deviation (error bars) over five runs with different seeds.}
\label{experiment8}
\end{figure}
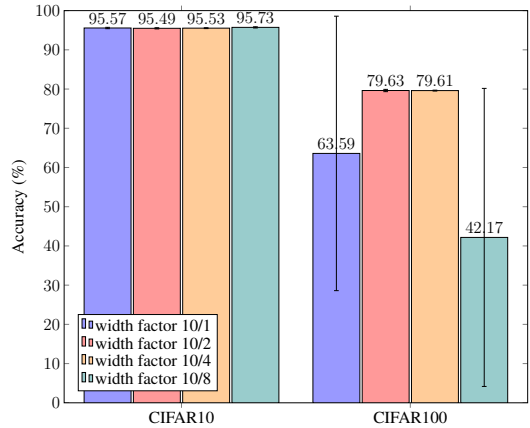

Figure~\ref{experiment8} shows the performance against variable initial width factors, while the general width factor is fixed to $10$.
Initial width factors $2$ and $4$ give satisfactory results, while factors $1$ and $8$ yield very unstable training (as indicated by the large variance). 
We attribute this to the large step either from $3$ feature maps to $16\cdot 8$ in the first layer (initial width factor $8$), or from $1\cdot16$ feature maps to $16\cdot 10$ in the second layer (initial factor $1$). 
An initial width factor of $4$ strikes a good balance and provides stability; hence we will use this value for all other experiments.

\begin{figure}[t]
\centering
\resizebox{0.5\columnwidth}{!}{
\begin{tikzpicture}
    \begin{axis}[
        style={font=\Large},
        symbolic x coords={CIFAR10,CIFAR100},
        xtick=data,
        ybar,
        bar width=40pt,
        nodes near coords,
        enlarge x limits={abs=100pt},
        ymin=70,
        ymax=100,
        ylabel={Accuracy (\%)},
        width = 440pt,
        legend cell align={left},
    ]
    \addplot[style={fill=blue!40, draw=black}, error bars/.cd, y dir=both, y explicit] coordinates 
    {(CIFAR10,94.36) += (0, 0.2) -= (0, 0.2)
     (CIFAR100,75.33) += (0, 0.19) -= (0, 0.19)};
    \addplot[style={fill=red!40, draw=black}, error bars/.cd, y dir=both, y explicit] coordinates 
    {(CIFAR10,95.41) += (0, 0.12) -= (0, 0.12)
     (CIFAR100,78.89) += (0,0.27) -= (0, 0.27)};
    \addplot[style={fill=orange!40, draw=black}, error bars/.cd, y dir=both, y explicit] coordinates 
    {(CIFAR10,95.53) += (0, 0.11) -= (0, 0.11)
     (CIFAR100,79.61) += (0, 0.11) -= (0, 0.11)};
    \addplot[style={fill=teal!40, draw=black}, error bars/.cd, y dir=both, y explicit] coordinates 
    {(CIFAR10,95.65) += (0, 0.09) -= (0, 0.09)
     (CIFAR100,80.09) += (0, 0.36) -= (0, 0.36)};
    \legend{width factor  4/2, width factor  8/4, width factor 10/4, width factor 12/4}
    \end{axis}
\end{tikzpicture}
}
\caption{Performance effect of \MIMOConv width factor ($N$=2, 200 epochs). Configurations are labeled with \textit{width factor}/\textit{inital width factor}. We report the average accuracy and the standard deviation (error bars) over five runs with different seeds.}
\label{experiment9}
\end{figure}
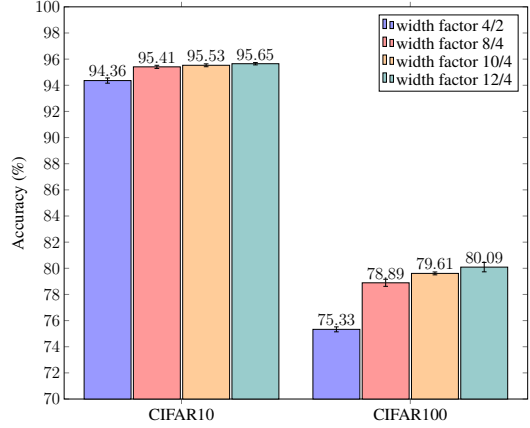

\paragraph{\MIMOConv width factor.}\label{results:total_width}
WideResNet architectures identify the number of feature maps as the most crucial factor in determining the capacity of a model.
At the same time, a higher number of feature maps means less interference between bound vectors.
In Figure~\ref{experiment9}, the benefits of large width factors can be observed. Notice that width enters both the number of parameters and the computational complexity quadratically unless grouped convolutions are used, where the number of groups increases with the channel width. Not exploring grouped convolutions, a trade-off between performance and accuracy has to be struck. We shall go with a width factor of $10$ for all other experiments.

\begin{figure}[t]
\centering
\begin{subfigure}{0.49\linewidth}
\resizebox{\linewidth}{!}{\begin{tikzpicture}
    \begin{axis}[
        style={font=\LARGE},
        symbolic x coords={$N$=$1$, $N$=$2$, $N$=$3$, $N$=$4$, $N$=$5$},
        xtick=data,
        ybar,
        bar width=40pt,
        nodes near coords,
        enlarge x limits={abs=60pt},
        ymin=65,
        ymax=100,
        ylabel={Accuracy (\%)},
        width = 550pt,
        legend cell align={left},
    ]
    \addplot[style={fill=blue!40, draw=black}, error bars/.cd, y dir=both, y explicit] coordinates 
                                         {($N$=$1$,96.81) += (0, 0.11) -= (0, 0.11)
                                          ($N$=$2$,95.53) += (0, 0.11) -= (0, 0.11)
                                          ($N$=$3$,94.28) += (0, 0.20) -= (0, 0.20)
                                          ($N$=$4$,92.57) += (0, 0.13) -= (0, 0.13)
                                          ($N$=$5$,91.15) += (0, 0.16) -= (0, 0.16)};
    \addplot[style={fill=red!40, draw=black}, error bars/.cd, y dir=both, y explicit] coordinates 
                                         {($N$=$1$,96.95) += (0, 0.09) -= (0, 0.09)
                                          ($N$=$2$,95.58) += (0, 0.14) -= (0, 0.14)
                                          ($N$=$3$,94.18) += (0, 0.29) -= (0, 0.29)
                                          ($N$=$4$,92.61) += (0, 0.09) -= (0, 0.09)
                                          ($N$=$5$,91.13) += (0, 0.2) -= (0, 0.2)};
    \legend{trainable keys, fixed keys}
    \end{axis}
\end{tikzpicture}}
\caption{CIFAR10}
\label{experiment12_cifar10}
\end{subfigure}
\begin{subfigure}{0.49\linewidth}
\resizebox{\linewidth}{!}{\begin{tikzpicture}
    \begin{axis}[
        style={font=\LARGE},
        symbolic x coords={$N$=$1$, $N$=$2$, $N$=$3$, $N$=$4$, $N$=$5$},
        xtick=data,
        ybar,
        bar width=40pt,
        nodes near coords,
        enlarge x limits={abs=60pt},
        ymin=65,
        ymax=100,
        ylabel={Accuracy (\%)},
        width = 550pt,
        legend cell align={left},
    ]
    \addplot[style={fill=orange!40, draw=black}, error bars/.cd, y dir=both, y explicit] coordinates
                                         {($N$=$1$,83.01) += (0, 0.24) -= (0, 0.24)
                                          ($N$=$2$,79.61) += (0, 0.11) -= (0, 0.11)
                                          ($N$=$3$,76.18) += (0, 0.28) -= (0, 0.28)
                                          ($N$=$4$,72.73) += (0, 0.26) -= (0, 0.26)
                                          ($N$=$5$,69.64) += (0, 0.19) -= (0, 0.19)};
    \addplot[style={fill=teal!40, draw=black}, error bars/.cd, y dir=both, y explicit] coordinates 
                                         {($N$=$1$,82.86) += (0, 0.32) -= (0, 0.32)
                                          ($N$=$2$,79.70) += (0, 0.23) -= (0, 0.23)
                                          ($N$=$3$,76.48) += (0, 0.15) -= (0, 0.15)
                                          ($N$=$4$,72.72) += (0, 0.25) -= (0, 0.25)
                                          ($N$=$5$,69.54) += (0, 0.17) -= (0, 0.17)};

    \legend{trainable keys, fixed keys}
    \end{axis}
\end{tikzpicture}}
\caption{CIFAR100}
\label{experiment12_cifar100}
\end{subfigure}

\caption{\MIMOConv with trainable or frozen binding keys for different numbers of superposition channels ($N$). Models are trained for 200 epochs. We report the average accuracy and the standard deviation (error bars) over five runs with different seeds.}
\label{experiment12}
\end{figure}
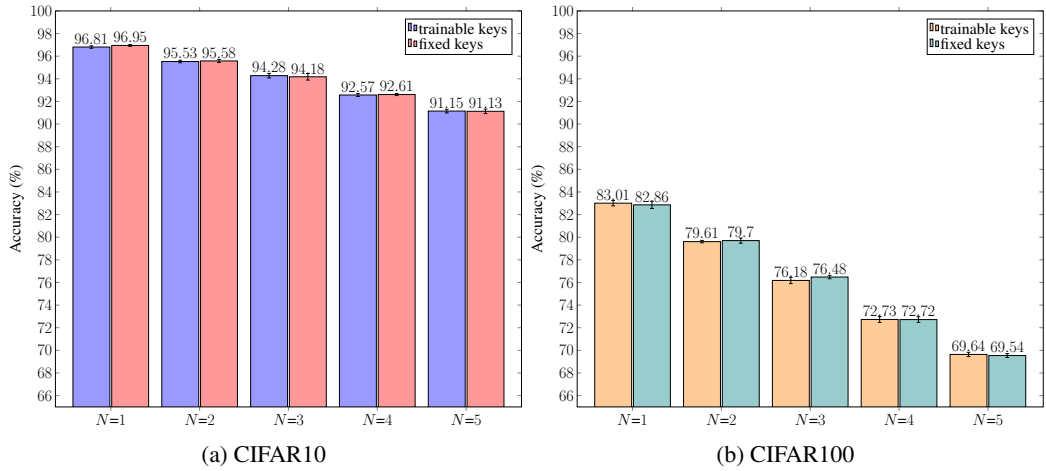

\paragraph{Freezing of binding keys.}\label{results:trainable_keys}
We investigate, if the binding keys can be frozen during training. As can be observed in Figure~\ref{experiment12}, keys do not need to be trainable while still maintaining a high accuracy across a wide range of superposition channel counts. We note that unbinding keys are always left trainable.

\newpage

\section{Experimental Setup and Evaluations on MIMOFormer}\label{sec:results_transformer}

\subsection{Experimental setup}

\subsubsection*{Datasets} 
\paragraph{LRA.}The long range arena (LRA) benchmark~\cite{tay2021long} is a suite of tasks consisting of sequences ranging from 1K to 16K tokens, covering a wide range of data
types and modalities. 
Below, we list the LRA tasks used in this work: 
\begin{itemize}
    \item \textbf{ListOps}: This dataset~\cite{nangia2018listops} tests the capability of modeling hierarchically structured data comprised of long sequences with operators (max, mean, median, modulo sum). The sequence length is up to 2K, and the task is to perform a ten-way classification. ListOps is released under an MIT license. 
    \item \textbf{Text}: The IMDb reviews dataset~\cite{maas2011learning} is a document classification benchmark. The benchmark uses byte-level sequences of up to 4K, entailing a binary classification task. 
    \item \textbf{Retrieval}: This task measures the model's capability of storing text document information into a compressed representation and matching or retrieving it against other documents. 
    The ACL Anthology Network~\cite{radev2013acl} is a byte/character level dataset with sequence lengths of up to 4K. It is a binary classification task. The ACL anthology corpus is released under the CC BY-NC 4.0 license.
    \item \textbf{Image}: In this task, RGB images of resolution $32\times 32$ from the CIFAR10~\cite{krizhevsky2009learning} dataset are flattened and classified by the sequence classification model. As a result, this is a ten-way classification task with sequences of length 1024. 
    \item \textbf{Pathfinder}: This task~\cite{linsley2018learning} requires a model to decide if two points (circles) are connected by a path of dashes on a black and white 2D image of dimension $32\times32$. The 2D images are flattened to sequences of length 1024, and classified by the sequence model as a binary classification. Pathfinder is released under MIT license. 
\end{itemize}
LRA further contains a more fine-grained version of Pathfinder, called Pathfinder-X, with $128\times 128$ image resolution leading to a sequence length of 16K. However, all Transformer variants considered in~\cite{tay2021long} (including the Performer) failed to learn this task. 
Since this work demonstrates the MIMO-capability of self-attention models rather than increasing their sequence length, we do not consider Pathfinder-X. 

\paragraph{Synthetic sequence benchmarks.}
We use two synthetic benchmarks~\cite{fu2022h3} which measure the basic reasoning capability of neural sequence models.
\begin{itemize}
    \item \textbf{Associative recall}: The task is to remember associations between pairs of tokens. For example, given a sequence of tokens \emph{a 2 c 4 b 3 d 1}, if the model is prompted with \emph{a}, the expected output is \emph{2}, the token following \emph{a} in the input sequence. If it were prompted with \emph{b}, the correct output would be \emph{3}, etc.
    Each sequence contains 40 characters, whereby a dictionary with 20 different characters is used. 
    \item \textbf{Induction head}: The task is to recall content after a special token (e.g., $\vdash$). For example, the string \emph{a d b $\vdash$ g f ... h c $\vdash$} would expect the response \emph{g}. 
    Each sequence contains 30 characters, whereby a dictionary with 20 different characters is used. 
\end{itemize}
Both tasks provide a training set (5000 examples) and a test set (500 examples).

\subsubsection*{Training setup}
As with \MIMOConv, the experiments are performed on an NVIDIA A100 Tensor Core GPU with 80\,GB memory and 8 CPU cores.
All experiments are repeated five times with a different random seed. We report the mean and standard deviations of accuracy to account for variability in training.
Overall, the training and evaluation of all \MIMOFormer models, including the ablation of the number of training steps, consumed 3112\,GPU hours. 
\paragraph{LRA.} Table~\ref{tab:mimoformer_setup} lists the deep \MIMOFormer architecture and the training setup for each task in the LRA benchmark~\cite{tay2021long}. 
Both wide and deep models use the same training setup, but wide models shrink to a single layer $L=1$ with inverse scaling in the number of heads $N_{head}$.
\MIMOFormer uses the same base architecture (number of heads, layers, dimensions, etc.) as proposed in the initial evaluation on LRA~\cite{tay2021long}. 
In addition, Table~\ref{tab:mimoformer_setup} also summarizes the model configuration and the settings used for the training of DataMux~\cite{murahari2022datamux}, our main competitor, on the Listops dataset. We base our DataMux model on the \emph{Roberta} architecture as specified in~\cite{murahari2022datamux}. We adjusted the number of heads, the number of layers, the embedding dimension, and the hidden dimension to approximately match the \MIMOFormer's number of parameters. Before training on ListOps, the scaled-down DataMux model is first pre-trained on the "retrieval warm-up task" as outlined in~\cite{murahari2022datamux}.

The training setup and the evaluation setup for \MIMOFormer is based on code provided by~\cite{xiong2021nystromformer}. 
Training uses an Adam optimizer ($\beta_1$=$0.9$ and $\beta_2$=$0.99$) with a OneCycleLR policy~\cite{smith2019clr} and additional warmup. 
All \MIMOFormer configurations use the same number of training steps per task; we note however that configurations with many superposition channels converge more slowly and consequently might benefit from additional training steps. 
Dropout after the attention block is applied. 
Finally, the output tokens are fed through average pooling and classified with a task-specific readout mechanism. 

\MIMOFormer faces training issues when the number of superposition channels is high (e.g., $N$=$4$). 
To this end, we propose a curriculum learning strategy where the number of superposition is reduced to $N$'=$N/2$ at the beginning of the training.
This warmup period is set to $1/6$th of the total number of training epochs.
The overall training setup, including the learning rate scheduling, remains the same. 

Table~\ref{tab:lra_training} shows the training time for the reported models. 
Since all \MIMOFormer configurations use the same number of training steps, we observe a reduced training time using a large number of superposition channels. Contrary to \MIMOConv, we do not repeatedly send batches through to ensure equal training time. 

\begin{table}[]
\centering
\caption{Architecture and training setup on LRA. $L$=number of layers; $N_\text{head}$=number of heads; $D_\text{head}$=head dimension; $E$=embedding dimension; $D_\text{hidden}$=hidden dimension in MLP; Bs=batch size; Lr=learning rate.}
\label{tab:mimoformer_setup}
\resizebox{0.9\linewidth}{!}{
\begin{tabular}{lccrrrrcrrc}
\toprule
           & \multicolumn{5}{c}{Model}                                                                         & \multicolumn{5}{c}{Training}                                                                                                                                                                                                                                    \\
           \cmidrule(r){2-6}\cmidrule(r){7-11}
           & $L$ & $N_\text{head}$ & \multicolumn{1}{l}{$D_\text{head}$} & \multicolumn{1}{l}{$E$} & \multicolumn{1}{l}{$D_\text{hidden}$} & \multicolumn{1}{l}{Bs} & Lr   & \multicolumn{1}{l}{\begin{tabular}[c]{@{}c@{}}Warmup\\ steps\end{tabular}} & \multicolumn{1}{c}{\begin{tabular}[c]{@{}c@{}}Train \\ steps\end{tabular}} & Dropout  \\
           \cmidrule(r){1-1}\cmidrule(r){2-6}\cmidrule(r){7-11}
\textbf{\MIMOFormer}\\
Listops    & 6 & 8       & 64                          & 512                   & 2048                          & 64                     & 1e-4 & 1000                                                                       & 20,000                                                                      & 0.1                                                           \\
Text       & 6 & 8       & 64                          & 512                   & 2048                          & 32                     & 1e-4 & 8000                                                                       & 40,000                                                                      & 0.1                                                             \\
Retrieval  & 4 & 4       & 32                          & 128                   & 512                           & 32                     & 1e-4 & 800                                                                        & 60,000                                                                      & 0.1                                                              \\
Image      & 3 & 4       & 64                          & 64                    & 128                           & 256                    & 1e-4 & 175                                                                        & 70,000                                                                      & 0.1                                                             \\
Pathfinder & 4 & 8       & 128                         & 128                   & 128                           & 256                    & 1e-4 & 312                                                                        & 124,800                                                                     & 0.1                   \\
\cmidrule(r){1-1}\cmidrule(r){2-6}\cmidrule(r){7-11}
\textbf{DataMUX}\\
Listops    & 12 & 12       & 120                          & 120                   & 3072                          & 48                     & 2e-5 & 0                                                                       & 20,000                                                                      & 0.1                                                           \\
\bottomrule
\end{tabular}
}
\end{table}

\begin{table}[t]
\centering
\caption{Training time (hours) on the long range arena (LRA) using an NVIDIA A100 GPU.}
\label{tab:lra_training}
\resizebox{\linewidth}{!}{
\begin{tabular}{lllllll}
\toprule
                        & \multicolumn{1}{c}{ListOps} & \multicolumn{1}{c}{Text}  & \multicolumn{1}{c}{Retrieval} & \multicolumn{1}{c}{Image} & \multicolumn{1}{c}{Pathfinder} & \multicolumn{1}{c}{Total}  \\
\cmidrule(r){1-1}\cmidrule(r){2-7}
\textbf{Deep models} & $L$=$6$, $H$=$8$ & $L$=$6$, $H$=$8$ & $L$=$4$, $H$=$4$ & $L$=$3$, $H$=$4$ & $L$=$4$, $H$=$8$ &\\
\cmidrule(r){1-1}\cmidrule(r){2-7}
Performer (reproduced)        & $3.45^{\pm0.00}$	& $5.89^{\pm0.00}$	& $5.93^{\pm0.01}$	& $3.55^{\pm0.02}$	& $26.66^{\pm0.02}$	& $45.47^{\pm0.04}$ \\
MIMOFormer (N=2, att.) & $3.96^{\pm0.03	}$ & 	$7.69^{\pm0.01}$ & 	$5.99^{\pm0.02}$ & 	$3.51^{\pm0.01}$ & 	$25.42^{\pm0.17}$ & 	$46.57^{\pm0.20}$  \\
MIMOFormer (N=2, att.+MLP) &  $3.42^{\pm0.03}$ & 	$6.71^{\pm0.26}$ & 	$5.38^{\pm0.01}$ & 	$3.14^{\pm0.01}$ & 	$22.50^{\pm0.18}$ & 	$41.15^{\pm0.31}$ \\
MIMOFormer (N=4, att.) & $3.09^{\pm0.01}$ &	$5.89^{\pm0.01}$ &	$4.35^{\pm0.02}$ &	$2.73^{\pm0.29}$ &	$21.26^{\pm0.21}$ &	$37.32^{\pm0.27}$  \\
MIMOFormer (N=4, att.+MLP) & $2.42^{\pm0.26	}$ &	$4.48^{\pm0.26}$ &	$3.44^{\pm0.01}$ &	$2.10^{\pm0.01}$ &	$17.01^{\pm0.37}$ &	$29.44^{\pm0.31}$ 	 \\
\cmidrule(r){1-1}\cmidrule(r){2-7}
\textbf{Wide models} & $L$=$1$, $H$=$48$ & $L$=$1$, $H$=$48$ & $L$=$1$, $H$=$16$ & $L$=$1$, $H$=$12$ & $L$=$1$, $H$=$32$ &\\
\cmidrule(r){1-1}\cmidrule(r){2-7}
Performer (reproduced)    & $2.46^{\pm0.05}$ & 	$5.23^{\pm0.01}$ & 	$5.26^{\pm0.01}$ & 	$3.21^{\pm0.39}$ & 	$29.52^{\pm0.03}$ & 	$45.68^{\pm0.38}$ \\
MIMOFormer (N=2, att.) & $2.54^{\pm0.00}$ &	$4.79^{\pm0.01}$ &	$4.46^{\pm0.01}$ &	$2.93^{\pm0.01}$ & 	$23.45^{\pm0.01}$ & 	$38.17^{\pm0.02}$ \\
MIMOFormer (N=2, att.+MLP) &  $2.31^{\pm0.02}$ &	$4.30^{\pm0.00}$ &	$4.16^{\pm0.00}$ &	$2.65^{\pm0.00}$ &	$20.69^{\pm0.06}$ &	$34.12^{\pm0.07}$ \\
MIMOFormer (N=4, att.) & $1.95^{\pm0.08	}$ &	$4.04^{\pm0.15}$ &	$3.17^{\pm0.14}$ &	$2.25^{\pm0.10}$ &	$19.43^{\pm0.37}$ &	$30.84^{\pm0.66}$\\
MIMOFormer (N=4, att.+MLP) & $1.57^{\pm0.08	}$ &	$3.39^{\pm0.00}$ &	$2.71^{\pm0.12}$ &	$1.87^{\pm0.08}$ &	$15.51^{\pm0.04}$ &	$25.06^{\pm0.16}$  \\
\bottomrule
\end{tabular}
}
\end{table}

\paragraph{Synthetic sequence benchmarks.} We use a light-weight \MIMOFormer with two layers, one head, an embedding dimension of 32, and a hidden dimension of 128. 
The model is trained with SGD for 400 epochs using a learning rate of 5e-4, batch size 32, and a weight decay of 0.1. 

To configure DataMux for associative recall, we use the \emph{SimpleLM} language model with 30.5K trainable parameters specified in the Safari repository \footnote{https://github.com/HazyResearch/safari} and insert multiplexing and demultiplexing layers at the input and the output of the model as specified in~\cite{murahari2022datamux}. Before experimenting with N=2 channels, we first tested the setup with a single channel where DataMux reached 99\% accuracy.

\subsection{Number of training steps}
In our standard training setup, we train the Performer and \MIMOFormer models for a large number of training steps ($\approx 2 \times$ of what was described in~\cite{xiong2021nystromformer}). 
Here, we show the benefit of a longer training procedure. 
Table~\ref{tab:lra_steps} compares the performance of the models when trained with $0.5\times$ or $1\times$ as many steps as the training setup reported in Table~\ref{tab:mimoformer_setup}.
%
Note that the test accuracies reported in Table~2 of the main text also used the standard training setup ($1\times$).
The longer training procedure improves the Performer's accuracy marginally ($0.23$--$0.48$\% gain). 
 Conversely, MIMOFormer notably benefits from the longer training in both deep ($1.57$--$3.06$\% gain) and wide models ($0.99$--$2.21$\% gain).

\begin{table}[t]
\centering
\caption{Training for more steps improves \MIMOFormer accuracy. We report the average test accuracy (\%) on LRA over five runs with different seeds when training the model for 0.5$\times$ / 1$\times$ the training steps described in Table~\ref{tab:mimoformer_setup}. \MIMOFormer uses an equal number of query superpositions ($N$) and value-key tensor product superpositions ($M$), i.e., $N$=$M$. Computation in superposition is performed either in attention only (att.) or in both attention and MLP (att.+MLP). $L$~is the number of layers, $H$ the number of heads.}
\label{tab:lra_steps}
\resizebox{\linewidth}{!}{
\begin{tabular}{lllllll}
\toprule
                        & \multicolumn{1}{c}{ListOps} & \multicolumn{1}{c}{Text}  & \multicolumn{1}{c}{Retrieval} & \multicolumn{1}{c}{Image} & \multicolumn{1}{c}{Pathfinder} & \multicolumn{1}{c}{Avg.}  \\
\cmidrule(r){1-1}\cmidrule(r){2-7}
\textbf{Deep models} & $L$=$6$, $H$=$8$ & $L$=$6$, $H$=$8$ & $L$=$4$, $H$=$4$ & $L$=$3$, $H$=$4$ & $L$=$4$, $H$=$8$ &\\
\cmidrule(r){1-1}\cmidrule(r){2-7}
Transformer~\cite{vaswani2017attention}             & $36.37$   & $64.27$ & $57.46$     & $42.44$ & $71.40$      & $53.39$ \\
Performer~\cite{choromanski2020performer}               & $18.01$   & $65.40$ & $53.82$     & $42.77$ & $77.05$      & $51.41$ \\
Performer (reproduced)        & $37.93$\mimotwo{38.94} &	$65.45$\mimotwo{65.70} & $81.37$\mimotwo{81.58} &	$40.04$\mimotwo{40.14} &	$73.01$\mimotwo{73.82}	& $59.56$\mimotwo{60.04} \\
MIMOFormer (N=2, att.) & $38.07$\mimotwo{38.08}&	$64.47$\mimotwo{65.00}&	$77.16$\mimotwo{79.37}&	$37.33$\mimotwo{38.21}&	$68.19$\mimotwo{72.36}& $57.04$\mimotwo{58.61}\\
MIMOFormer (N=2, att.+MLP) &  $37.28$\mimotwo{37.65}&  	$64.30$\mimotwo{64.39}&	$73.33$\mimotwo{76.02}&	$31.62$\mimotwo{33.85}& $56.31$\mimotwo{67.98}& $52.57$\mimotwo{55.98}\\
MIMOFormer (N=4, att.) & $31.39$\mimotwo{37.22}&	$64.73$\mimotwo{64.59}&		$57.67$\mimotwo{60.99}&		$27.48$\mimotwo{28.16}&		$49.86$\mimotwo{55.50}&		$46.23$\mimotwo{49.29}\\
MIMOFormer (N=4, att.+MLP) & $17.91$\mimotwo{17.74}	& $53.97$\mimotwo{60.71}&	$66.24$\mimotwo{72.20}&	$23.30$\mimotwo{24.01}&	$50.26$\mimotwo{50.33}&	$42.33$\mimotwo{45.00}\\
\cmidrule(r){1-1}\cmidrule(r){2-7}
\textbf{Wide models} & $L$=$1$, $H$=$48$ & $L$=$1$, $H$=$48$ & $L$=$1$, $H$=$16$ & $L$=$1$, $H$=$12$ & $L$=$1$, $H$=$32$ &\\
\cmidrule(r){1-1}\cmidrule(r){2-7}
Performer (reproduced)    & $39.13$\mimotwo{39.40}& 	$65.73$\mimotwo{65.73}& $83.20$\mimotwo{83.67}& $41.53$\mimotwo{41.67}& $73.88$\mimotwo{74.11}& $60.70$\mimotwo{60.93}\\
MIMOFormer (N=2, att.) & $38.31$\mimotwo{38.90}	& $65.40$\mimotwo{65.39}&	$78.71$\mimotwo{81.27}& $39.98$\mimotwo{40.25}&	$71.97$\mimotwo{73.51}& $58.87$\mimotwo{59.86}\\
MIMOFormer (N=2, att.+MLP) & $37.76$\mimotwo{37.59}&	$64.73$\mimotwo{64.64}	& $75.26$\mimotwo{78.30}&	$35.14$\mimotwo{36.69}& $67.60$\mimotwo{68.22}& 	$56.10$\mimotwo{57.09} \\
MIMOFormer (N=4, att.) & $36.97$\mimotwo{37.71}	&	$64.61$\mimotwo{64.22}	&	$71.50$\mimotwo{74.99}	&	$31.13$\mimotwo{35.43}	&	$67.56$\mimotwo{69.52}	&	$54.35$\mimotwo{56.37}\\
MIMOFormer (N=4, att.+MLP) & $17.41$\mimotwo{18.52}&	$64.24$\mimotwo{63.53}&	$68.91$\mimotwo{74.30}&	$24.21$\mimotwo{26.54}&	$53.36$\mimotwo{56.33}& $45.63$\mimotwo{47.84}\\
\bottomrule
\end{tabular}
}
\end{table}

\subsection{Computational complexity}

\begin{table}[t]
\caption{Billions of multiply-accumulate (GMAC) operations per sample on Text (subtask of LRA). Model configuration reads \(L[\text{ayers}] = 6, N_{\text{head}} = 8, D_{\text{head}} = 64, E[\text{mbedding}] = 512, D_{\text{hidden}} = 2048, \text{ and } R = 256\) where \(R\) determines the fidelity of the FAVOR+ attention approximation. Number in parenthesis shows the relative share of the overall complexity.}
\label{tab:mac_mimoformer}
\centering
\resizebox{\linewidth}{!}{
\begin{tabular}{lrrrrrr}
\toprule
                 & 
                 \multicolumn{1}{l}{\begin{tabular}[c]{@{}c@{}}K/Q/V \\ Projections  \end{tabular}}
                   & \multicolumn{1}{l}{Attention}
                   & \multicolumn{1}{l}{\begin{tabular}[c]{@{}c@{}}Binding \& \\ Unbinding\end{tabular}}& \multicolumn{1}{l}{MLPs}        & \multicolumn{1}{l}{\begin{tabular}[c]{@{}c@{}}Readout \\ Layer\end{tabular}} & \textbf{Total}     \\
\cmidrule(r){1-1}\cmidrule(r){2-7}
Transformer & 19.34 (14.9\%)        & 58.80 (45.3\%)          & \multicolumn{1}{c}{n.a.}   & 51.62 (39.8\%)         & 0.001 (0.001\%)  & \textbf{129.8}   \\
Performer & 19.34 (21.4\%)         & 19.58 (21.6\%)          & \multicolumn{1}{c}{n.a.}  & 51.62 (57.0\%)        & 0.001 (0.001\%)  & \textbf{90.5}  \\
\cmidrule(r){1-1}\cmidrule(r){2-7}
MIMOFormer (N=2, att.)   & 19.34 (23.0\%)           & 13.05 (15.5\%)  & 0.050 (0.06\%)  & 51.62 (61.4\%)  & 0.001 (0.001\%)  & \textbf{84.1}   \\
MIMOFormer (N=2, att.+MLP)   & 19.34 (35.2\%)          & 9.80 (17.8\%)  & 0.050 (0.09\%)  & 25.81 (46.9\%)  & 0.001 (0.002\%)  & \textbf{55.0}   \\
MIMOFormer (N=4, att.)   & 19.34 (23.9\%)       & 9.78 (12.1\%)   & 0.050 (0.06\%)  & 51.62 (63.9\%)   & 0.001 (0.001\%)  & \textbf{80.8}  \\
MIMOFormer (N=4, att.+MLP)   & 19.34 (52.0\%)          & 4.90 (13.2\%)   & 0.050 (0.14\%)  & 12.90 (34.7\%)   & 0.001 (0.003\%)  & \textbf{37.2}  \\
\bottomrule
\end{tabular}
}
\end{table}

As can be deduced from Table~\ref{tab:mac_mimoformer}, the integration of variable binding mechanisms via binding and unbinding operations is inconsequential. 
It amounts to only between 0.06\% and 0.14\% of the computational complexity for MIMOFormer despite being performed at each attention layer. 
The K/Q/V projections make up a considerable part of the overall computational complexity; hence, computing them in superposition would further reduce the number of computes per input.


\subsection{The importance of faithful attention scores}
DataMux~\cite{murahari2022datamux}, claims to retain high performance for subsets of the GLUE~\cite{wang2019glue} and CoNLL-2003~\cite{sang2003introduction} benchmarks, despite using up to $40$ inputs in superposition. However, as discussed in~\cite{hassid2022does}, none of the tasks reported on require attention layers at all. Indeed, DataMUX does not redesign its attention algorithm, but keeps a single scalar attention score $A_{i,j}$ for each pair of token positions, which effectively multiplies the (unnormalized) attention score of each (protected) superposition channel: 
\begin{equation}
    A_{i,j} = \exp\left(\langle \sum\limits_{w=1}^N k_j^{(w)},\ \sum\limits_{t=1}^N q_i^{(t)}\rangle / \sqrt{D}\right) \approx \exp\left(\sum\limits_{w=1}^N \langle k_j^{(w)},\ q_i^{(w)}\rangle / \sqrt{D}\right) = \prod\limits_{t=1}^N A_{i,j}^{(w)}
\end{equation}
As our experiments confirm (see Section~5.2), on more nuanced tasks in NLP such as “associative recall” and “induction head”, which require faithful attention, their method drops to 20.04\% and 6.06\% for N=2, while ours, at a score of  96.52\% and 99.40\% respectively, succeeds. Despite investing significant efforts in the training of DataMUX, it cannot perform on these synthetic tasks. This is in line with the findings of~\cite{fu2022h3} which identifies the lack of attention as the reason that the Structured State Space Sequence (S4) model~\cite{gu2022efficiently} is able to completely outperform state of the art in LRA~\cite{tay2021long}, but is not competetive for large language models. In contrast to DataMUX, our work approximates true attention and our theoretical derivations show convergence to actual dot-product attention as the hidden dimension increases, giving us an even stronger case for applicability to large language models (for instance, GPT-3 uses embedding dimension 12,888, far exceeding the maximum of 512 we report on). 

\newpage

\section{Supporting Theorems}\label{app:theorems}
The theorems presented in this section are of general nature and stated for completeness.
\begin{thm}
    Any inner-product preserving map $T:X \to Y$ between two inner-product spaces $X,Y$ is linear\label{thm:inner_product_preserving_map_is_linear}
\end{thm}
\begin{proof}
    Let $u,v \in X$ and $\lambda \in \mathbb{C}$. Then
    \begin{align}
     &\ \norm{T(\lambda u + v) - \lambda T u - T v}^2 = \langle T(\lambda u + v) - \lambda T u - T v , T(\lambda u + v) - \lambda T u - T v \rangle\\
    = &\ \langle T(\lambda u + v), T(\lambda u + v) \rangle - 2 \lambda \langle T(\lambda u + v), T u \rangle - 2 \langle T(\lambda u + v), T v \rangle \nonumber \\
    &+ \lambda^2 \langle T u, T u \rangle + 2 \lambda \langle T u, T v \rangle + \langle T v, T v \rangle\\
    = &\ \langle \lambda u + v, \lambda u + v \rangle - 2 \lambda \langle \lambda u + v, u \rangle - 2 \langle \lambda u + v, v \rangle
    + \lambda^2 \langle u, u \rangle + 2 \lambda \langle u, v \rangle + \langle v, v \rangle\\
    = &\ 2 \langle \lambda u + v, \lambda u + v \rangle - 2 \lambda \langle \lambda u + v, u \rangle - 2 \langle \lambda u + v, v \rangle\\
    = &\ 2 \langle \lambda u + v, \lambda u + v \rangle - 2 \langle \lambda u + v, \lambda u + v \rangle\\
    = &\ 0
\end{align}
which implies
\begin{equation}
    T(\lambda u + v) = \lambda T u + T v
\end{equation}

\end{proof}

\begin{thm}[Hoeffding's Inequality]
    Let $X_1, \ldots, X_n$ be independent bound random variables satisfying $\abs{X_i} \leq a_i$ and $\mathbb{E}[X_i] = 0$. Then,
    \begin{equation}
        \mathbb{P}\left\{ \abs{\sum\limits_{i=1}^n X_i} > t\right\} \leq 2 \exp \left( - \tfrac{t^2}{2\sum_{i=1}^n a_i^2} \right)
    \end{equation}
    \label{thm:hoeffding's inequality}
\end{thm}
\begin{proof}
Following \cite{bandeira2020}, we shall prove that
\begin{align}
    \mathbb{P}\left\{\sum\limits_{i=1}^n X_i > t\right\} \leq \exp \left( - \tfrac{t^2}{2\sum_{i=1}^n a_i^2} \right)
\end{align}
from which by symmetry and union bound the statement follows.
By Markov's inequality and for $\lambda > 0$
\begin{align}
    \mathbb{P}\left\{\sum\limits_{i=1}^n X_i > t\right\} & = \mathbb{P}\left\{\lambda \sum\limits_{i=1}^n X_i > \lambda t\right\} = \mathbb{P}\left\{\exp(\lambda \sum\limits_{i=1}^n X_i) > \exp(\lambda t)\right\}\\
    & \leq \mathbb{E}\left[\exp(\lambda \sum\limits_{i=1}^n X_i)\right] \bigg/ \exp(\lambda t) = \exp(-\lambda t) \prod\limits_{i=1}^n \mathbb{E}\left[\exp(\lambda X_i)\right] 
\end{align}
where the last equality follows from independence of $\{X_i\}_{i=1}^n$. Because the function $x \mapsto \exp(\lambda x)$ is convex it holds
\begin{equation}
    \exp(\lambda x) \leq \tfrac{a_i+x}{2a_i}\exp(\lambda a_i) + \tfrac{a_i-x}{2a_i}\exp(-\lambda a_i)
\end{equation}
for all $x \in [-a_i, a_i]$. Thus, since $\abs{X_i} \leq a_i$ we may use the above and that $\mathbb{E}[X_i] = 0$ to bound $\mathbb{E}\left[\exp(\lambda X_i)\right]$
\begin{align}
    \mathbb{E}[\exp(\lambda X_i)] & \leq \mathbb{E}\left[  \tfrac{a_i+x}{2a_i}\exp(\lambda a_i) + \tfrac{a_i-x}{2a_i}\exp(-\lambda a_i)\right] = \tfrac{1}{2}(e^{\lambda a_i} + e^{- \lambda a_i}) = \cosh(\lambda a_i)\\
    & = \sum\limits_{n=0}^\infty \tfrac{(\lambda a_i)^{2n}}{(2n)!} \leq \sum\limits_{n=0}^\infty \tfrac{(\lambda a_i)^{2n}}{2^n n!} = \exp((\lambda a_i)^2/2)
\end{align}
where the Taylor expansion of $\cosh(\cdot)$ and $\exp((\cdot)^2/2)$ were used and the penultimate step is given by $(2n)! \geq 2^n n!$. Hence, we get for any $\lambda > 0$
\begin{equation}
    \mathbb{P}\left\{\sum\limits_{i=1}^n X_i > t\right\} \leq \exp(-\lambda t) \prod\limits_{i=1}^n \exp((\lambda a_i)^2/2) = \exp(-\lambda t +  \tfrac{\lambda^2}{2}\sum\limits_{i=1}^n a_i^2)
\end{equation}
and in particular for $\lambda = t/(\sum_{i=1}^n a_i^2) > 0$ one gets
\begin{equation}
    \mathbb{P}\left\{\sum\limits_{i=1}^n X_i > t\right\} \leq \exp(-\tfrac{t^2}{2 \sum_{i=1}^n a_i^2})
\end{equation}
\end{proof}

\begin{thm}[On the Norm of Hadamard Products]\label{thm:norm_hadamard_product}
    Let $X \in S^{D-1}$ follow an arbitrary distribution and let $Y \in S^{D-1}$ be uniformly distributed and independent from $X$. Then
    \begin{equation}
        \mathbb{E}\left[ \norm{X \odot Y}_2^2 \right] = \tfrac{1}{D}
    \end{equation}
    and 
    \begin{equation}
        \mathbb{P}\left\{ \norm{X \odot Y}_2^2 \leq  \tfrac{1 + \beta}{D} \right\} \geq \tfrac{1}{1+1/\beta}
    \end{equation}
\end{thm}
\begin{proof}
    Since by definition $\norm{Y}_2^2 = 1$, it follows by rotational symmetry and linearity of expectation that
    \begin{equation}
        \mathbb{E}\left[Y_q^2\right] = \sum\limits_{p=1}^D \mathbb{E}\left[Y_p^2\right]\bigg/ D  =  \mathbb{E}\left[ \sum\limits_{p=1}^D Y_p^2\right]\bigg/ D = \tfrac{1}{D}
    \end{equation}
    Also, by linearity of expectation and independence
    \begin{align}
        \mathbb{E}\left[ \norm{X \odot Y}_2^2 \right] = \mathbb{E}\left[ \sum\limits_{p=1}^D X_p^2 Y_p^2 \right] = \sum\limits_{p=1}^D \mathbb{E}\left[X_p^2\right] \mathbb{E}\left[Y_p^2\right] = \tfrac{1}{D}\sum\limits_{p=1}^D \mathbb{E}\left[X_p^2\right] = \tfrac{1}{D} \mathbb{E}\left[ \sum\limits_{p=1}^D X_p^2\right] = \tfrac{1}{D}
    \end{align}
    Hence, we can apply Markov to get
    \begin{equation}
        \mathbb{P}\left\{ \norm{X \odot Y}_2^2 \leq  \tfrac{1 + \beta}{D} \right\} = 1 - \mathbb{P}\left\{ \norm{X \odot Y}_2^2 \geq  \tfrac{1 + \beta}{D} \right\} \geq 1 - \tfrac{D \cdot \mathbb{E}\left[\norm{X \odot Y}_2^2\right]}{1 + \beta} = 1 - \tfrac{1}{1 + \beta} = \tfrac{1}{1+1/\beta}
    \end{equation}
\end{proof}

\newpage

\section{Limitations}
MIMONets exploit the Blessing of Dimensionality, that with high probability exponentially many (in dimension D) vectors are almost orthogonal. Although the components of MIMONet are made near isometric through regularization, a certain number of (hidden) dimensions is still necessary. This naturally limits MIMONets to large (oftentimes over-parametrized) models or models employing low-rank decompositions.

The number of inputs that can be superposed without incurring heavy losses in accuracy is limited given a fixed neural network due to increasingly strong interference between the superposition channels. 

The proposed superposition capable attention mechanism converges to faithful attention (without interference between channels) as the embedding dimension increases, but at the price of only a speedup of $N$ when using $N^2$ superposition channels. Being built on linearized attention such as FAVOR+, it further inherits all their benefits (linear scaling) and drawbacks (limited parallelization and increased memory accesses for autoregressive training (see Section 3.1 in\cite{hua2022transformer}). On the other hand, trivial superposition would yield a speedup of $N^2$ instead, but at the cost of blurring the attention scores with each token-token score summarizing attention in all superposition channels at once. Such models employing blurry attention are limited to application where imprecise “summarizing” information suffices.


\newpage

\bibliographystyle{IEEEtran}
\bibliography{references}